\documentclass[10pt]{article} % For LaTeX2e

\usepackage[accepted]{rlj} % Should be uncommented for the camera-ready
\usepackage{amssymb}            % Defines common symbols like \mathbb R
\usepackage{mathtools}          % Extends amsmath, providing common math tools
\usepackage{mathrsfs}           % Enables \mathscr, which can work in cases that \mathcal does not
\usepackage{graphicx}           % For including images
\usepackage{subcaption}         % Allows for the use of subfigures and subcaptions
\usepackage[space]{grffile}     % For spaces in image names
\usepackage{url}                % For displaying URLs
\usepackage{lipsum}             % For placeholder text
\usepackage[utf8]{inputenc} % allow utf-8 input
\usepackage[T1]{fontenc}    % use 8-bit T1 fonts
\usepackage{hyperref}       % hyperlinks
\usepackage{url}            % simple URL typesetting
\usepackage{booktabs}       % professional-quality tables
\usepackage{amsfonts}       % blackboard math symbols
\usepackage{nicefrac}       % compact symbols for 1/2, etc.
\usepackage{microtype}      % microtypography
\usepackage{xcolor}         % colors
\usepackage{natbib}
\usepackage{graphicx}

\usepackage{amsmath}
\usepackage{amssymb}
\usepackage{mathtools}
\usepackage{amsthm}

\usepackage{bm}
\usepackage{appendix}
\usepackage{caption}

\usepackage{mathrsfs}

\usepackage{dsfont}
\usepackage{extarrows}
\usepackage{enumitem}
\usepackage{xcolor}
\usepackage{wrapfig}
\usepackage{algorithm}
\usepackage{algpseudocode}

\newcommand{\cS}{\mathcal{S}}
\newcommand{\cA}{\mathcal{A}}
\let\hat\widehat
\let\bar\overline
\theoremstyle{plain}
\newtheorem{theorem}{Theorem}[section]
\newtheorem{proposition}[theorem]{Proposition}
\newtheorem{lemma}[theorem]{Lemma}
\newtheorem{corollary}[theorem]{Corollary}
\theoremstyle{definition}

\newtheorem{assumption}[theorem]{Assumption}
\theoremstyle{remark}

\title{Statistical Inference for Policy Evaluation with Temporal Difference Learning}

\setrunningtitle{Statistical Inference for Policy Evaluation with Temporal Difference Learning}

\author{Weichen Wu\textsuperscript{1}, Gen Li\textsuperscript{2}, Yuting Wei\textsuperscript{3}, Alessandro Rinaldo\textsuperscript{4}}

\emails{weichen.wu.1996@gmail.com, \ genli@cuhk.edu.hk, \ ytwei@wharton.upenn.edu, \ alessandro.rinaldo@austin.utexas.edu}

\affiliations{
$^{1}$\textbf{The Voleon Group, New York, NY, USA}\\
$^{2}$\textbf{Department of Statistics, The Chinese University of Hong Kong, Hong Kong SAR.}\\
$^{3}$\textbf{Department of Statistics and Data Science, The Wharton School, University of Pennsylvania, Philadelphia, PA, USA.}\\
$^{4}$\textbf{Department of Statistics and Data Sciences, University of Texas, Austin TX, USA.}
}

\contribution{
    We derive a new high-dimensional Berry-Esseen bound for the estimation error over the class of convex sets with an online variance estimator, achieving an $O(T^{-\frac{1}{3}})$ rate with $T$ the number of iterations, which is the fastest in the literature.
    }
    {
    The Gaussian approximation results we obtained deliver the sharpest rate in the literature, improving on recent contributions.
    }

\contribution{
    We develop a computationally efficient online algorithm to construct an estimator of the asymptotic variance of the TD estimator. We also show that the total variation distance between the computable estimated Gaussian distribution and the asymptotic Gaussian distribution vanishes with high probability at a rate $O(T^{-\frac{1}{3}})$, matching the order of our Berry-Esseen rate. 
    }
    {
    Existing estimators of the asymptotic variance suffer from a significantly higher time and space complexity  than the online procedure we propose. Thus, our methodology is practically relevant.
    }

\contribution{
    We construct confidence regions and simultaneous confidence intervals for the linear parameters of the value function approximation, with guaranteed finite-sample coverage. 
    }
    {
   To the best of our knowledge, ours is the first procedure for formal statistical inference on the model parameters with a finite sample coverage guarantee.
    }

\contribution{
    We derive the first high-probability convergence guarantee for the TD estimator with Polyak-Ruppert averaging that matches the asymptotic variance up to logarithm factors, without requiring a projection step to control the iterative trajectory or stepsizes that depend on the number of iterations.
    }
    {
    Prior work established similar convergence guarantees with larger variances and under stronger assumptions on the stepsizes.
    }

\keywords{Temporal Difference Learning, Statistical Inference, Berry-Esseen bound, Central Limit Theorem} % Your keywords

\summary{We investigate the statistical properties of Temporal Difference (TD) learning with Polyak-Ruppert averaging, arguably one of the most widely used algorithms in reinforcement learning, for the task of estimating the parameters of the optimal linear approximation to the value function. Assuming independent samples, we make three theoretical contributions that improve upon the current state-of-the-art results: (i) we establish refined high-dimensional Berry-Esseen bounds over the class of convex sets, achieving faster rates than the best known results, and (ii) we propose and analyze a novel, computationally efficient online plug-in estimator of the asymptotic covariance matrix; (iii) we derive sharper high probability convergence guarantees that depend explicitly on the asymptotic variance and hold under weaker conditions than those adopted in the literature.
These results enable the construction of confidence regions and simultaneous confidence intervals for the linear parameters of the value function approximation, with guaranteed finite-sample coverage. 
We demonstrate the applicability of our theoretical findings through numerical experiments. 
}

\begin{document}

\makeCover  % Create the cover page
\maketitle  % Make the title section

\begin{abstract}
We investigate the statistical properties of Temporal Difference (TD) learning with Polyak-Ruppert averaging, arguably one of the most widely used algorithms in reinforcement learning, for the task of estimating the parameters of the optimal linear approximation to the value function. Assuming independent samples, we make three theoretical contributions that improve upon the current state-of-the-art results: (i) we establish refined high-dimensional Berry-Esseen bounds over the class of convex sets, achieving faster rates than the best known results, and (ii) we propose and analyze a novel, computationally efficient online plug-in estimator of the asymptotic covariance matrix; (iii) we derive sharper high probability convergence guarantees that depend explicitly on the asymptotic variance and hold under weaker conditions than those adopted in the literature.
These results enable the construction of confidence regions and simultaneous confidence intervals for the linear parameters of the value function approximation, with guaranteed finite-sample coverage. 
We demonstrate the applicability of our theoretical findings through numerical experiments. 
\end{abstract}

\section{Introduction}

The feasibility and accuracy of statistical inference tasks, such as hypothesis testing or constructing confidence regions or simultaneous confidence intervals, are traditionally predicated on using estimators that are not only consistent but also satisfy some type of distributional approximation. In most cases, this is accomplished via Central Limit Theorems (CLTs), which ensure that the procedures in use admit a Gaussian approximation. Due to the high dimensionality of modern scientific and engineering applications, there has been a surge of interest in establishing new {\it high-dimensional Berry-Esseen bounds,} CLTs that hold in high dimensions and have a finite-sample guarantee. While efforts in this area have focused on Berry-Esseen bounds for normalized sums, few results are available to handle more complicated scenarios, where estimators are computed in a nonlinear, iterative manner. In this paper, we investigate one such scenario and study in detail the consistency, Gaussian approximations, and statistical inference for the output of the Temporal Difference (TD) learning algorithm for policy evaluation with linear function approximation in Reinforcement Learning (RL). %, a prominent area that underpins many recent machine learning and AI breakthroughs. 

A critical task in RL is to evaluate with confidence the quality of a given policy, measured by its value function.  
As RL models are based on Markov Decision Processes (MDPs), statistical inference must accommodate the online nature of the sampling mechanism. 
TD learning  \citep[see, e.g.,][]{sutton1988learning}, an iterative procedure that can be framed as an instance of stochastic approximation (SA), is arguably the most widely used and effective algorithm for policy evaluation in RL. %TD learning, which can be regarded as , approximates the value function of a given policy in an iterative manner.
The last several years have witnessed a flurry of activities within the statistical machine learning community to derive finite sample consistency rates of TD iterates, leading to deep results and a much improved understanding of the performance and limitations of TD learning. See, e.g.,  \cite{khamaru2020temporal,li2024q,bhandari2018finite,dalal2018finite,srikant2019finite,lakshminarayanan2018linear,mou2020linear,durmus2022finite,li2023sharp,samsonov2023finitesample,Prashanth2013ConcentrationBF,Borkar25} for a partial and growing list of relevant references.  
Though for the tabular case (finite state and action spaces) the TD algorithm is known to be minimax rate optimal \citep{li2024q}, its properties -- and, in particular, its sample complexity -- in more complex and high-dimensional settings have not been fully understood.

In a more specialized line of work, concerned with establishing distributional approximations of the TD iterates, it has been shown that, using Polyak-Ruppert averaging \citep{polyak1992acceleration,ruppert1988efficient}, the TD estimation error converges to a Gaussian distribution in the large sample limit; see, e.g., \cite{fort2015central,mou2020linear,li2023online}. While these results provide a formal justification for the validity of statistical inference, the guarantees from those contributions are non-quantitative, in the sense that they are of asymptotic nature and hold only in fixed dimensions. As a result, they do not adequately capture the dependence on the sample size and other problem-related quantities, most notably the dimensionality, all of which affect the accuracy of statistical inference.  
In this paper, we partially address this gap. We focus on TD learning with Polyak-Ruppert averaging for estimating the coefficients of the best linear approximation of the value function, assuming i.i.d. samples. We make the following theoretical and methodological contributions.
\begin{enumerate}

\item \emph{Sharper Berry-Esseen bounds.} We derive a new high-dimensional Berry-Esseen bound for the estimation error over the class of convex sets and using an estimated variance, achieving an $O(T^{-\frac{1}{3}})$ rate as a function of the number of iterations, which appears to be the fastest in the literature.  While completing this work, we became aware of the recent related paper by \citet{samsonov2024gaussian}, which, under the same settings, derived a high-dimensional Berry-Esseen bound for linear SA with Polyak-Ruppert averaging, obtaining the slower rate $O(T^{-\frac{1}{4}})$. See Section~\ref{sec:berry-esseen} for a discussion and detailed comparisons.
%.of es theWe illustrate that the difference between the finite-sample and asymptotic distributions of the estimation error is bounded by $O(T^{-\frac{1}{3}})$ when measured on \emph{any convex set}, with an appropriate choice of stepsize-decaying speed. This result utilized the tools applied by \cite{samsonov2024gaussian} but improved upon their $O(T^{-\frac{1}{4}})$ rate by polishing the details in the proof.

\item \emph{Improved high-probability guarantees.} We show that, for an appropriate choice of the initial stepsize, the Euclidean norm of the estimation error for the TD learning algorithm is of order
\begin{align*}
O\left(\sqrt{\frac{\mathsf{Tr}(\bm{\Lambda}^\star)}{T} \log \frac{1}{\delta}}\right),
\end{align*}
with probability at least $ 1- \delta$, where $T$ denotes the number of iterations, $\bm{\Lambda}^\star$ represents the asymptotic variance matrix of the estimation error, and $\mathsf{Tr}(\bm{\Lambda}^\star)$  its trace. To the best of our knowledge, this is the first high-probability convergence guarantee for this problem that matches the asymptotic variance, up to logarithm factors, without requiring a projection step to control the iterative trajectory. Furthermore, our assumptions are weaker than those used in the literature. Table~\ref{table:sample-complexity} compares our high-probability convergence guarantees with the latest available results.

\item \emph{Novel, computationally efficient variance estimator.} We develop a computationally efficient online algorithm to construct an estimator $\hat{\bm{\Lambda}}_T$ of the asymptotic variance $\bm{\Lambda}^\star$. We also show that the total variation distance between the (computable) distribution $\mathcal{N}(\bm{0},\hat{\bm{\Lambda}}_T)$ and the asymptotic distribution $\mathcal{N}(\bm{0},\bm{\Lambda}^\star)$ vanishes with high probability at a rate $O(T^{-\frac{1}{3}})$, matching the order of our Berry-Esseen rate. As a result, estimating the variance, a crucial step for constructing confidence sets, does not lead to worse rates.

\item \emph{Statistical inference.} Based on the Berry-Esseen bound and the variance estimator, we design an online, efficient algorithm to construct confidence regions and simultaneous confidence intervals for the linear parameters of the value function approximation. We prove that the finite-sample coverage rate of proposed confidence region converges to the target level at a rate of $O(T^{-\frac{1}{3}})$.

\end{enumerate}

\begin{figure*}[t]
\begin{center}
\begin{tabular}{cc}
\includegraphics[width =\textwidth]{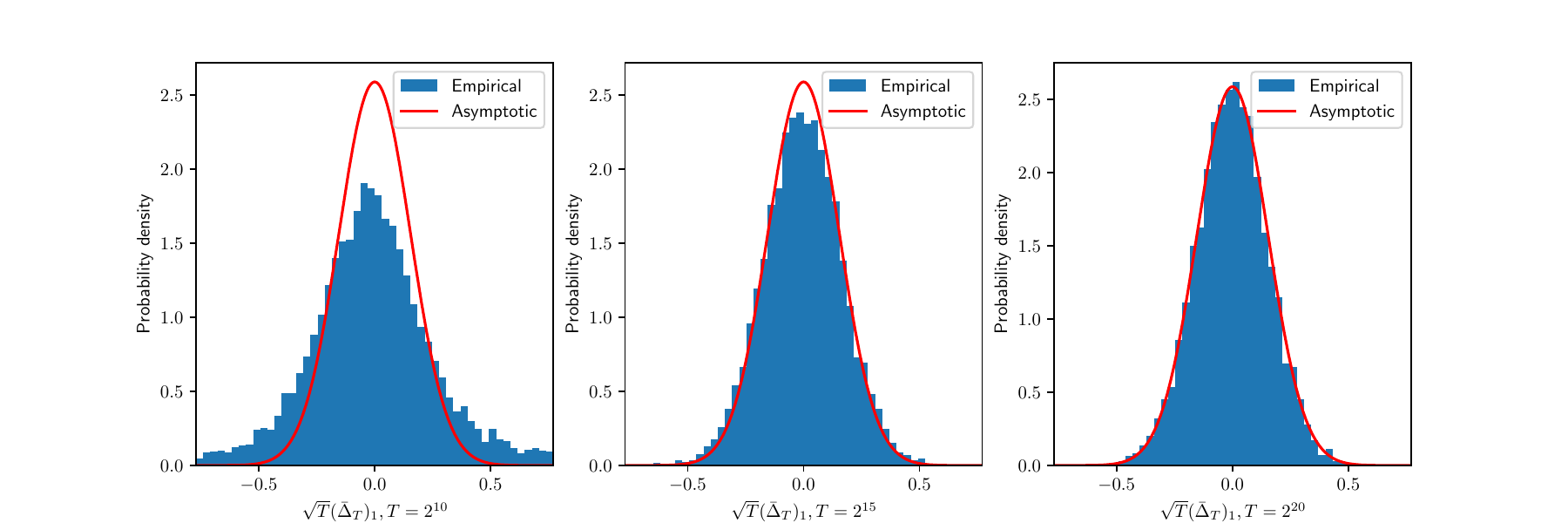}\\
\end{tabular}
\end{center}
\caption{Emergence of the first entry of TD estimation error on an MDP with a 3-dimensional feature map.}
% \yuting{if we don't have space, we can include 3 of them instead 4}
\label{fig:iid-d3-alpha067-deltaT1-converge}
\end{figure*}

%\yuting{correct footnote issue in the table}

%\begin{table*}[t]
%\centering
%\renewcommand{\arraystretch}{2.3}
%\begin{tabular}{c|c|c} 
%\toprule
%paper   &  convergence guarantee & choice of $\alpha$\\ \toprule
%	\cite{samsonov2024gaussian}  & $d_{\mathsf{C}}(\sqrt{T}\bar{\bm{\Delta}}_T,\bm{\Lambda}^{\star \frac{1}{2}}\bm{z}) \leq O(T^{-\frac{1}{4}})$ & $\alpha = \frac{1}{2}$\\ \hline 
%	\textbf{This work, Theorems \ref{thm:Berry-Esseen}, \ref{thm:asymptotic}}  & $d_{\mathsf{C}}(\sqrt{T}\bar{\bm{\Delta}}_T,\bm{\Lambda}^{\star \frac{1}{2}}\bm{z}) \leq O(T^{-\frac{1}{3}})$ & $\alpha = \frac{2}{3}$ \\ \toprule
%\end{tabular}
%\caption{Comparisons between our Berry-Esseen bounds for TD estimation error with prior work.}

%\label{table:Berry-Esseen}
%\end{table*}

We verify our theoretical findings with numerical simulations, discussed in Section \ref{sec:simulation} and the appendix. As an illustration, Figure \ref{fig:iid-d3-alpha067-deltaT1-converge} shows how the empirical distribution of TD estimation error in $10,000$ independent trials (as shown in blue histograms), scaled by the square-root of the number of iterations $T$, converges to its asymptotic Gaussian distribution (shown by the red curves). %When $T = 100$, the empirical distribution is biased towards the left; as $T$ increases to $10^3$, $10^4$ and finally $10^5$, the distribution not only moves toward but also concentrates around $0$. When $T = 10^5$, the empirical distribution align well with the asymptotic distribution.

%\subsection{Notation}
%\paragraph{Numbers.} For any positive integer $n$, we use $[n]$ to denote the set of all positive integers no larger than $n$: $[n]:= \{1,2,...,n\}$. Throughout the paper, we use $C,C',C_1,C_2...$ to denote universal constants that are independent of the number of iteration $T$, target probability $\delta$, the stepsizes, and all other MDP-related parameters; notice that their exact values may vary from line to line. Furthermore, we use $\widetilde{C},\widetilde{C}',...$ to denote parameters that are independent of the number of iteration $T$ and tolerance level $\delta$ but may depend on other parameters related to the MDP or the choice of stepsizes; their exact values, as well as their dependency on these parameters, may also vary from line to line. 

{\bf The i.i.d. condition.} Our results are derived under the standard assumption of i.i.d. data, which has been widely used in the recent literature on statistical analyses of TD learning, SA and stochastic gradient descent \citep[see, e.g.][]{sheshukova2025gaussianapproximationmultiplierbootstrap}. While unrealistic, the theoretical investigation of this simplified scenario has proved challenging; the present work addresses some of the gap still existing in the literature and provide state-of-the art rates. We mention that analogous results, obtained using different techniques and yielding slower rates, have been recently obtained for the more challenging case of Markovian data: see \cite{srikant2024rates} and \cite{weichen.markov.25}.

{\bf Notation.} %Vectors and matrices.} 
We use boldface small letters to denote vectors and capital letters to denote matrices. For a vector $\bm{x}$,  $\|\bm{x}\|_2$ denotes its Euclidean norm; for a matrix $\bm{M}$, we use $\mathsf{Tr}(\bm{M})$, $\mathsf{det}(\bm{M})$, $\|\bm{M}\|_{\mathsf{F}}$ and $\|\bm{M}\|$ to denote its trace, determinant, Frobenius and spectral norm, respectively.
% (i.e., the largest singular value), and  to denote its , i.e., $\|\bm{M}\|_{\mathsf{F}}=\sqrt{\mathsf{Tr}(\bm{M}^\top \bm{M})}$. 
For a symmetric matrix $\bm{M} \in \mathbb{R}^{d \times d}$, we use $\lambda_{\max}(\bm{M}) = \lambda_1(\bm{M}) \geq \ldots \lambda_d(\bm{M}) = \lambda_{\min}(\bm{M})$ to denote its eigenvalues and $\mathrm{cond}(\bm{M})$ to indicate its condition number. For a vector $\bm{x} \in \mathbb{R}^d$ and a symmetric matrix $\bm{M} \in \mathbb{R}^{d \times d}$, we let $\|\bm{x}\|_{\bm{M}} = \sqrt{\bm{x}^\top \bm{Mx}}$. We will use the notation $\tilde{C}$ to express a quantity independent of $T$, the number of iterations/sample size, but possibly dependent on other problem-related parameters.  Similarly, for positive sequences $\{f_t\}_{t \in \mathbb{N}}$ and $\{g_t\}_{t \in \mathbb{N}}$, we write $f_t \lesssim g_t$, or $f_t = O(g_t)$, to signify that $f_t \leq \tilde{C} g_t$, for some $C > 0$ independent of $t$. 
%\paragraph{The Big $O$ and small $o$.} \textcolor{violet}{ALE: come back to this. Also mention that$\lesssim$ means inequality up to a positive universal constant} 
%Given two sequences $\{f_t\}_{t \in \mathbb{N}}$ and $\{g_t\}_{t \in \mathbb{N}}$, we use the expression $f_t \lesssim g_t$, or $f_t = O(g_t)$, to represent the fact that there exists a quantity $\widetilde{C} > 0$ independent of $t$, such that $f_t \leq \widetilde{C} g_t$ for all $t \in \mathbb{N}$. \yuting{what about other problem-related parameters? what about $d$? modify theorem statements accordingly}\weichen{$\tilde{C}$ can depend on other parameters. Theorems modified.}
%If there exist universal constants $C_1>0$ and $C_2 > 0$ such that $C_1 g_t \leq f_t \leq C_2 g_t$ holds uniformly for all natural number $t$, we write $f_t \asymp g_t$, or $f_t = \Theta(g_t)$. Meanwhile, if $\lim_{t \to \infty} f(t)/g(t) = 0$, we write $f_t = o(g_t)$. 
%\paragraph{Distance between probability distributions.} 
The {\it convex distance} between two probability distributions on $\mathbb{R}^d$, say $P$ and $Q$, is defined as 
\begin{align*}
d_{\mathsf{C}}(P,Q) = \sup_{\mathcal{A} \in \mathscr{C}_d} |P(\mathcal{A}) - Q(\mathcal{A})|,
\end{align*}
where $\mathscr{C}_d$ is the class of all convex subsets of $\mathbb{R}^d$.
Also, we use $d_{\mathsf{TV}}(P,Q)$ to denote the total variation (TV) distance between $P$ and $Q$. Notice that $d_{\mathsf{C}}(P,Q) \leq d_{\mathsf{TV}}(P,Q)$.

%\section{PROBLEM FORMULATION}
\section{Background}
%\yuting{why capital letters?}
\label{sec:models}

%\subsection{Model and settings}

%\yuting{rewrite; can't be the same with your previous paper}

\paragraph{Markov decision process.}

An infinite-horizon discounted Markov Decision Process (MDP) \citep{bertsekas2017dynamic,Sutton1998} is the tuple $\mathcal{M} = (\cS, \cA, P, r, \gamma)$ where $\cS$  and $\cA$ represent the {\it state space} of the environment and the {\it action space} available to an agent, respectively, both assumed for simplicity to be finite (though our results hold for countable spaces). The function $P:\cS\times\cA\mapsto\Delta(\cS)$ is a transition kernel that associates to any state-action pair $(s,a) \in \cS \times \cA$ the probability distribution $P(\cdot \mid s,a)$ of the next state of the environment when the agent executes action $a$ while the environment is in state $s$. The {\it reward function} $r: \cS \times \cA \mapsto [0,1]$ measures the immediate reward to the agent after taking action $a$ at state $s$; %Throughout the paper, we assume without loss of generality that $r(s,a) \in [0,1]$ for any state $s \in \cS$ and any action $a \in \cA$. 
finally, $\gamma \in [0,1)$ is the {\it discount factor}, used to assess the impact of future rewards.% in quantifying the value function as would be discussed shortly after.

%\footnotetext{$\lambda$ can take any positive value smaller than $\min_{1 \leq i \leq d}\{\mathsf{Re}(\lambda_i(\bm{A}))\}$.}

In choosing their actions, the agent follows a randomized action selection rule,  or \emph{policy},  represented as a (time-invariant) conditional probability distribution  $\pi: \cS \mapsto \Delta(\cA)$ over the action space given the state of the environment. For any policy $\pi$, its associated reward function $r:\cS \mapsto [0,1]$ is the expected reward of the next action selected according to the policy $\pi$, i.e.
$
r(s):=\mathbb{E}_{a \sim \pi(\cdot \mid s)}[r(s,a)],$ for $s \in \cS.
$
%\yuting{did we use $r^{\pi}$? or simply $r$?}
%For readability, we will omit the superscript $\pi$ and use $r(\cdot)$ to denote $r(\cdot)$. 
Any policy $\pi$ induces a transition kernel $P$, representing the probability distribution of the next state given the current state $s$ and the policy $\pi$, given by
$
P(s' \mid s) = \mathbb{E}_{a \sim \pi(\cdot \mid s)} P(s'\mid s,a)$ for $s,s' \in \cS.
$
We assume throughout that there exists a unique stationary distribution $\mu$ for the transition matrix ${P}$;  that is,  %\mathbb{P}_{s \sim \mu^{\pi}}(s = i) 
 $\mu(i) = \mathbb{P}_{s \sim \mu, s' \sim P(\cdot \mid s)}(s' = i)$, $i \in \mathcal{S}$.
%\begin{align*}
%0 < %\mathbb{P}_{s \sim \mu^{\pi}}(s = i) 
% \mu(i) = \mathbb{P}_{s \sim \mu, s' \sim P(\cdot \mid s)}(s' = i), \quad \forall i \in \mathcal{S}.
%\end{align*}

\paragraph{The value function and the Bellman equation.} 
The worth of a policy $\pi$ is measured by its {\it value function}  $s \in \cS \mapsto V(s):=\mathbb{E}\left[\sum_{t=0}^{\infty}\gamma^{t}r(s_{t},a_{t})\,\big|\,s_{0}=s\right]$ returning, for each initial state,   the expected value of all cumulative future rewards, appropriately discounted. %Formally, 
%\begin{align}
%\label{eqn:value-func}
%$V(s):=\mathbb{E}\left[\sum_{t=0}^{\infty}\gamma^{t}r(s_{t},a_{t})\,\big|\,s_{0}=s\right],$ for $s\in \cS$.
%\quad s\in \cS.
%\end{align}
A fundamental, defining feature of the value function is that it arises as the {\it unique} solution to the {\it Bellman equation}, a functional fixed-point equation of the form 
\begin{align}\label{eq:Bellman}
	V(s) = r(s) + \gamma \mathbb{E}_{s' \sim P(\cdot \mid s)}V(s')=: \mathcal{T}V(s). 
\end{align}
where $\mathcal{T}$ is known as the \emph{Bellman's operator}. %Indeed, the value function is the {\it unique} solution to the Bellman equation. 
%With a slight abuse of notation, we shall omit the superscript $\pi$ in $V^\pi$. % and abbreviate it as $V$  as the target policy is considered fixed. 
%An important result in the theory of MDPs  turns out that $V^{\pi}$ satisfies the renown Bellman equation by taking a one-step look ahead:

% Applying the vector and matrix representation, \eqref{eq:Bellman} can be expressed as
% \begin{align}\label{eq:Bellman-matrix}
% \bm{V} = \bm{r} + \gamma \bm{P}\bm{V}=: \mathcal{T} \bm{V},
% \end{align}
% where we define $\mathcal{T}:\mathbb{R}^{|\cS|} \mapsto \mathbb{R}^{|\cS|}$ as the \textbf{Bellman's operator}. 

\paragraph{Linear policy approximation and the projected Bellman equation.}
%In words, the value function corresponds to the expected discounted cumulative reward by executing policy $\pi$ under MDP $\mathcal{M}$ when starting from the state $s_{0} = s$.
% where the expectation is taken with respect to $a_t \sim \pi(s_t)$ and $s_{t+1} \sim P(\cdot \mid s_t,a_t)$ for every $t \in [T]$. 
In most practical scenarios, the value function is very complex and impossible to compute due to the prohibitively large size of the state space $\mathcal{S}$.
% As a result, it is difficult to collect sufficient samples which can support a meaningful and reliable estimation of the original value function $V$. 
% A natural alternative is to approximate the value function by some lower-dimensional representation. 
A common way to overcome this issue is to consider a parametric approximation of $V$ through a linear function, which is both computationally tractable and amenable to statistical estimation. %A large body of literature has investigates Therefore, it is common practice to approximate the value function with some lower-dimensional space,with linear space being the most trackable and popular choice as investigated by a strand of earlier works\cite{}. \yuting{TBA} 
% This paper is concerned with the most simplified version of such representation: a linear combination of a feature map. 
In detail, for a given feature map $\bm{\phi}:\mathcal{S} \mapsto \mathbb{R}^d$, mapping each state in  $\cS$ to a centered feature vector in $\mathbb{R}^d$, where $d$ is typically smaller than $|\cS|$, a linear value function approximation is given by
\begin{align}\label{eq:linear-approximation}
s \in \cS \mapsto	 V_{\bm\theta}(s) := {\bm\phi}(s)^\top {\bm\theta},
\end{align}
where ${\bm\theta} \in \mathbb{R}^d$ is a vector of linear coefficients. Throughout this paper, we assume that the Euclidean norm of the features is uniformly bounded, i.e., $\|\bm{\phi}(s)\|_2 \leq 1$ for all $s \in \mathcal{S}$. Throughout, we let %$\bm{\Sigma}$ 
\begin{align}\label{eq:defn-Sigma}
	\bm{\Sigma} := \mathbb{E}_{s \sim \mu } [\bm{\phi}(s)\bm{\phi}^\top(s)]
\end{align}
be the feature covariance matrix and use $\lambda_0$, $\lambda_{\Sigma}$ to denote its smallest and largest eigenvalues, respectively. We assume $\lambda_0 > 0$.

The best approximating linear function is the one parametrized by the  coefficient $\bm{\theta}^\star$  solving the \emph{projected Bellman equation} \citep{tsitsiklis1997analysis}
%\begin{align}
	%\label{eq:projected-Bellman}
	%%%%%%%V_{\bm{\theta}^\star} := \underset{V_{\bm{\theta}}:\bm{\theta} \in \mathbb{R}^d}{\text{argmin }} \mathbb{E}_{s \sim \mu} (V_{\bm{\theta}}(s) - \mathcal{T}V_{\bm{\theta}}(s))^2.
	$V_{\bm{\theta}^\star} = \Pi \mathcal{T}V_{\bm{\theta}^\star},$
%\end{align}
with $\Pi$ the orthogonal projection onto the subspace of functions that can be represented by linear combinations of the features. See, e.g., \cite{duan2021optimal}.
% With the vector and matrix notation, \eqref{eq:projected-Bellman} can be represented as
% \begin{align}
% \bm{\Phi}\bm{\theta}=\Pi_{\bm{D}_{\mu}}\mathcal{T}^{\pi}\bm{\Phi}\bm{\theta}\label{eq:projected-Bellman-eqn},
% \end{align}
% where $\Pi_{\bm{D}_{\mu}}$ represents the projection onto the column space of $\bm{{\Phi}}$:
% \begin{align*}
% \Pi_{\bm{D}_{\mu}} \bm{z} = \underset{y = \bm{\Phi} x:\bm{x} \in \mathbb{R}^d}{\text{argmin}} \|\bm{z} - \bm{y}\|_{\bm{D}_{\bm{\mu}}}^2, \quad \text{for all } \bm{z} \in \mathbb{R}^{|\cS|}.
% \end{align*} 
In turn, this (unique!) solution is given by \footnote{notice that the matrix $\bm{A}$ is invertible due to our assumption that $\lambda_0 > 0$; please refer to Lemma \ref{lemma:A} for details.}
\begin{align}
\label{eq:defn-theta-star}
\bm{\theta}^{\star}:=\bm{A}^{-1}\bm{b}, 
\end{align}
where %the matrix $\bm{A} \in \mathbb{R}^{d \times d}$ and the vector $\bm{b} \in \mathbb{R}^d$ are defined respectively as
%\begin{subequations} 
%\begin{align}
%	\bm{A} & :=\mathop{\mathbb{E}}\limits _{s\sim\mu,s'\sim P(\cdot\mid s)}\left[\bm{\phi}(s)\left(\bm{\phi}(s)-\gamma\bm{\phi}(s')\right)^{\top}\right]\in\mathbb{R}^{d\times d},\label{eq:defn-At-mean}\\
%	\text{ and }\bm{b} & :=\mathop{\mathbb{E}}\limits _{s\sim\mu}\left[\bm{\phi}(s)r(s)\right]\in\mathbb{R}^{d}.
%	\label{eq:defn-bt-mean}
%\end{align}
%\end{subequations}
\begin{align}\label{eq:defn-At-mean}
&\bm{A} :=\mathop{\mathbb{E}}\limits _{s\sim\mu,s'\sim P(\cdot\mid s)}\left[\bm{\phi}(s)\left(\bm{\phi}(s)-\gamma\bm{\phi}(s')\right)^{\top}\right]\in\mathbb{R}^{d\times d} \\ 
&\text{and} \quad \bm{b}  :=\mathop{\mathbb{E}}\limits _{s\sim\mu}\left[\bm{\phi}(s)r(s)\right]\in\mathbb{R}^{d}.
\end{align}
%Here, we use $\mu$ to denote the stationary distribution of $P^{\pi}.$
%As a result, to find the best linear function approximation of the value function, it is sufficient to find the solution to the linear equation~\eqref{eq:defn-theta-star}. 

\paragraph{The Temporal Difference (TD) learning with Polyak-Ruppert averaging.} Although in most real-world RL scenarios the state-action-reward tuples follow a Markov decision process, many of the most recent theoretical contributions -- including those delivering the sharpest guaranties listed in Table~\ref{table:sample-complexity} and \cite{samsonov2024gaussian} -- have made the simplifying but technically very convenient assumption of $i.i.d.$ data. %there are applications where $i.i.d.$ samples become available, such as in game playing and autonomous driving. Furthermore, it is illuminating to first understand the statistical properties of this simpler setting before considering the more complicated setting of Markov samples. In this paper, we will make this simplifying assumption.
%Specifically, we will use the observed data consisting 
Here we follow the same convention and analyze the properties of the TD learning algorithm applied to a sequence of $T$ $i.i.d.$ state-action-reward tuples of the form
% This work follows this practice and begin by analyzing the TD algorithm with $i.i.d.$ samples
\begin{align}
	\label{eq:iid-samples}
	 \{(s_t,a_t,s_t',r_t)\}_{t=1}^T,
\end{align}
where $(s_1,\ldots, s_T) \stackrel{i.i.d.}{\sim} \mu$, the stationary distribution and, for each $t$,  $a_t \sim \pi(s_t), r_t = r(s_t,a_t)$ and $s_t' \sim P(\cdot \mid s_t,a_t)$.

The TD learning algorithm is a stochastic approximation method to find, in an iterative manner, the solution \eqref{eq:defn-theta-star}. % of the linear system $\bm{A} \bm{\theta} = \bm{b}$. 
In detail, for a sequence of pre-selected step sizes $\{\eta_t\}_{t\geq 1},$ TD proceeds using the update rule 
%\begin{subequations}
%	\label{eq:TD-update-all} 
%	\begin{align}
%	\bm{\theta}_{t} & =\bm{\theta}_{t-1}-\eta_{t}(\bm{A}_{t}\bm{\theta}_{t-1}-\bm{b}_{t}),
%	\label{eq:TD-update-rule}
%\end{align}
%\end{subequations}
\begin{equation}
	\label{eq:TD-update-all}
	\bm{\theta}_{t}  =\bm{\theta}_{t-1}-\eta_{t}(\bm{A}_{t}\bm{\theta}_{t-1}-\bm{b}_{t}),
\end{equation}
where 
\begin{equation}
\label{eq:defn-At}
\bm{A}_t := \bm{\phi}(s_{t})\left(\bm{\phi}(s_{t})-\gamma\bm{\phi}(s_{t}')\right)^{\top} \quad 
\text{and }\quad \bm{b}_{t}  :=r_t\bm{\phi}(s_t).%\label{eq:defn-bt}
\end{equation}
%In vanilla TD, $\bm{\theta}_T$, the final iterate, is the  estimator for $\bm{\theta}^\star$.
%It is important to note that if the initial state $s_0$ is drawn from the stationary distribution $\mu$, then for every $t$, $\mathbb{E}[\bm{A}_t] = \bm{A}$ and  $\mathbb{E}[\bm{b}_t] = \bm{b}$. Though the time series of sample points are naturally drawn an MDP, in this paper, we make the simplifying assumption that the $T$ data points are i.i.d., where  $s_t \sim \mu$, as opposed to  $s_{t+1} \sim P(\cdot \mid s_t,a_t)$, for each $t$. While restrictive, this is a standard assumption in the theoretical literature on the analysis of TD learning. 
% Here, $\bm{A}_{t}$ and $\bm{b}_{t}$ are defined in the expressions~\eqref{eq:defn-At} and \eqref{eq:defn-bt}, respectively, and $\eta_t>0$ denotes the learning rate or stepsize at the iteration $t$. 
 
Rather than the final TD iterate $\bm{\theta}_T$, to estimate $\bm{\theta}^\star$ it is often preferred to use instead  the average of all the iterates
\begin{align}
	\label{eq:TD-averaging}
	\bar{\bm{\theta}}_T & =\frac{1}{T}\sum_{t=1}^{T}\bm{\theta}_{t}. 
\end{align}
 This estimator is an instance of the Polyak-Ruppert averaging scheme \citep{polyak1992acceleration,ruppert1988efficient}, a stochastic approximation methodology that provably reduces the variability and accelerates convergence. Furthermore, with a proper choice of the stepsizes, the averaged iterates converge to a Gaussian distribution, thus enabling statistical inference.
To date, the best Berry-Esseen bounds  rates for stochastic approximation methods and TD learning with Polyak-Ruppert averaging are due to \cite{ samsonov2024gaussian}, which in turn are based on general non-linear Berry-Esseen bounds by \cite{shao2021berry}.  %\cite{} \yuting{TBA}, 
In this paper, we improve on these results by deriving novel and sharper high-probability and Berry-Esseen bounds for the estimation error of the TD algorithm 
\begin{align}
	\label{eq:defn-bar-deltat}
	\bar{\bm{\Delta}}_T = \bar{\bm{\theta}}_T - \bm{\theta}^\star.
\end{align}
We conclude by mentioning that the choice of learning rates $\{\eta_t\}_{t \geq 0}$ in the updates \eqref{eq:TD-update-all}  plays a critical role in determining the statistical properties of (both the original and)  the averaged TD estimator \eqref{eq:TD-averaging}. 
Here we assume a \emph{polynomial-decaying} stepsize schedule $\eta_t = \eta_0 t^{-\alpha}$ for $\alpha \in (1/2,1)$.

\section{Results}

% While in most real-world RL scenarios the state-action-reward tuples are generated following a Markov decision process, there are applications where $i.i.d.$ samples become available, such as in game playing and autonomous driving. Furthermore, it is illuminating to first understand the statistical properties of this simpler setting before considering the more complicated setting of Markov samples. In this paper, we will make this simplifying assumption.
% Specifically, we will analyze the performance of the TD learning algorithm with Polyak-Ruppert averaging
% % each $s_t$ is independently drawn from the stationary distribution $\mu$ and 
% assuming a sample consisting of $T$ $i.i.d.$ state-action-reward tuples of the form
% % This work follows this practice and begin by analyzing the TD algorithm with $i.i.d.$ samples
% \begin{align}
% 	\label{eq:iid-samples}
% 	 \{(s_t,a_t,s_t',r_t)\}_{t=1}^T,
% \end{align}
% where $(s_1,\ldots, s_T) \stackrel{i.i.d.}{\sim} \mu$ and, for eact $t$,  $a_t \sim \pi(s_t), r_t = r(s_t,a_t)$ and $s_t' \sim P(\cdot \mid s_t,a_t)$. Correspondingly, $\bm{A}_t$ and $\bm{b}_t$ are defined as
% \begin{align*}
% \bm{A}_t = \bm{\phi}(s_{t})\left(\bm{\phi}(s_{t})-\gamma\bm{\phi}(s_{t}')\right)^{\top},\quad
% \text{and} \quad \bm{b}_t &= r_t\bm{\phi}(s_t).
% \end{align*}
% This section provides the main results regarding the non-asymptotic analysis of averaged TD learning 
It is known \citep{polyak1992acceleration,fort2015central} that, in fixed-dimensional settings, the TD estimation error of \eqref{eq:defn-bar-deltat} satisfies 
\begin{align}
\label{eqn:classic-clt}
\sqrt{T} \, \bar{\bm{\Delta}}_T  \xrightarrow{d} \mathcal{N}(\bm{0},\bm{\Lambda}^\star),
\end{align}
where the asymptotic covariance matrix $\bm{\Lambda}^\star$ is given by 
\begin{align}
&\bm{\Lambda}^\star = \bm{A}^{-1} \bm{\Gamma} \bm{A}^{-\top}, \quad \text{with}   \\ 
&\bm{\Gamma} = \mathbb{E}[(\bm{A}_t \bm{\theta}^\star - \bm{b}_t)(\bm{A}_t \bm{\theta}^\star - \bm{b}_t)^\top].  \label{eq:defn-Lambdastar}
\end{align}
% \yuting{$\bm{\Gamma}$ not defined yet}
Thus, the above result implies the familiar parametric rate
\begin{align}
\|\overline{\bm{\Delta}}_T\|_2 = O_P \left(\sqrt{\frac{\mathsf{Tr}(\bm{\Lambda}^\star)}{T}} \right), \label{eq:rate.asymp}
\end{align}
which may serve as a benchmark.
Below, we will extend these guarantees to a high-dimensional scenario by demonstrating an explicit dependence on all model-related parameters including, in particular, the dimension of the feature space. Our main goal is to show how the TD algorithm can be used to construct non-asymptotic confidence regions and simultaneous confidence intervals for $\bm{\theta}^\star$ in high dimensions.
To that effect, we will establish a high-probability convergence guarantee for $\bar{\bm{\Delta}}_T$, develop high-dimensional Berry-Esseen bounds over the class of convex sets, and finally build confidence regions for $\bm{\theta}^\star$ using $\bar{\bm{\theta}}_T$ and a novel, online estimator of the asymptotic variance. %These topics will be addressed in this order.

% \textcolor{red}{TODO: address the necessity of the condition on $\eta_0$ and $\delta$.}

%The statistical properties, especially the sample complexity of both the original and the averaged TD learning algorithms have been extensively investigated. Recently, there have been various attempts to establish high-probability guarantees for the convergence rate of estimating the best linear parameter with different choices of stepsizes. 

% \paragraph{The selection of the intial stepsize $\eta_0$.} An important fact to note about 
% \yuting{also compare with results in \cite{li2024q}}
% The bound of Theorem~\ref{thm:bar-delta-t} can be made dependent on the asymptotic variance $\bm{\Lambda}^\star$ without affecting the rate. Indeed, as 
% %\paragraph{Tightness of Theorem \ref{thm:bar-delta-t}.} 
% we will see below, the difference between $\bar{\bm{\Lambda}}_T$ and $\bm{\Lambda}^\star$ converges at the rate of $O(T^{\alpha-1})$. Thus, the dominant terms in Theorem \ref{thm:bar-delta-t} match the convergence rate  \eqref{eq:rate.asymp} implied by the asymptotic distribution up to a logarithm term. 
% \yuting{not sure I understand this statement...there is no convergence rate in the asymptotic result?} \textcolor{violet}{I Weichen: I changed the text to clarify Yuting's concern - is this what you meant?}

\subsection{High-dimensional Berry-Esseen bounds}\label{sec:berry-esseen}

We start with a novel high-dimensional Berry-Esseen bounds for the TD learning estimator, certifying that, uniformly over the class of $d$-dimensional convex sets, $\sqrt{T} \bar{\bm{\Delta}}_T$ is close in distribution to a centered Gaussian random vector with the optimal asymptotic covariance \eqref{eq:defn-Lambdastar}, thus closely mimicking the classical, fixed-dimensional CLT guarantee in \eqref{eqn:classic-clt}. We reiterate that, to the best of our knowledge, the resulting rates are the sharpest to date.

\begin{theorem}\label{thm:Berry-Esseen-combined}
Consider the TD algorithm with Polyak-Ruppert averaging~\eqref{eq:TD-update-all}, independent samples and decaying stepsizes $\eta_t = \eta_0 t^{-\alpha}$ for $\alpha \in (\frac{1}{2},1)$ and $\eta_0< \frac{1}{4}$. Then,
%\ifarxiv
% \begin{align*}
% &d_{\mathsf{C}}(\sqrt{T}\bar{\bm{\Delta}}_T,\mathcal{N}(\bm{0},\bm{\Lambda}^\star)) \lesssim \sqrt{\frac{\eta_0 }{(1-\alpha)(1-\gamma) \lambda_0}} \mathsf{Tr}(\bm{\Gamma})\|\bm{\Gamma}^{-1}\| T^{-\frac{\alpha}{2}}+ \frac{\sqrt{d\mathsf{cond}(\bm{\Gamma})}}{(1-\gamma)\lambda_0\eta_0} T^{\alpha-1}
% + \tilde{C} \; T^{-\frac{1}{2}} + \tilde{C}' \; T^{2\alpha-2}
% \end{align*} 
% \else
\begin{align*}
&d_{\mathsf{C}}(\sqrt{T}\bar{\bm{\Delta}}_T,\mathcal{N}(\bm{0},\bm{\Lambda}^\star)) \lesssim \sqrt{\frac{\eta_0 }{(1-\alpha)(1-\gamma) \lambda_0}} \mathsf{Tr}(\bm{\Gamma})\|\bm{\Gamma}^{-1}\| T^{-\frac{\alpha}{2}}\\ 
&+ \frac{\sqrt{d\mathsf{cond}(\bm{\Gamma})}}{(1-\gamma)\lambda_0\eta_0} T^{\alpha-1}
+ \tilde{C} \; T^{-\frac{1}{2}} + \tilde{C}' \; T^{2\alpha-2}
\end{align*} 
% \fi
where $\tilde{C},\tilde{C}'$ are problem-related quantities independent of $T$.
\end{theorem}

{\bf On the final rate.} By choosing $\alpha=2/3$, we obtain a rate in $T$ of order $O(T^{-1/3})$. In fact, in Theorem~\ref{thm:Berry-Esseen} in the Appendix we establish the more refined result that the Berry-Esseen bound (in the convex distance) between the distribution of $\sqrt{T}\bar{\bm{\Delta}}_T$ and a Gaussian distribution with (approximately) matching first two moments is of order $\sqrt{\frac{\eta_0 }{(1-\alpha)(1-\gamma) \lambda_0}} \mathsf{Tr}(\bm{\Gamma})\|\bm{\Gamma}^{-1}\| T^{-\frac{\alpha}{2}}$.  As $\alpha \to 1$, this rate approaches the optimal rate of $1/\sqrt{T}$. The additional term $ \frac{\sqrt{d\mathsf{cond}(\bm{\Gamma})}}{(1-\gamma)\lambda_0\eta_0} T^{\alpha-1}$ in the above bound, which is ultimately the reason for the final rate $O(T^{-1/3})$, originates from replacing an approximation to the finite sample variance of $\sqrt{T}\bar{\bm{\Delta}}_T$ with its asymptotic variance $\bm{\Lambda}^\star$. 

{\bf Comparison with \cite{samsonov2024gaussian}.} 
Prior to this work, the best Berry-Esseen bounds for stochastic approximation with polynomial-decaying stepsizes and Polyak-Ruppert averaging are those obtained by \cite{samsonov2024gaussian}, thus covering the settings considered here.
%While the problem of deriving non-asymptotic Gaussian approximation for stochastic approximation was investigated in a sequence of recent literature \cite{} \yuting{TBA}, the most advanced result featuring the Berry-Esseen bound under condition \eqref{eq:eta0-alpha-condition} was recently achieved in \cite{samsonov2024gaussian}.
Specifically, the authors show in their Theorem 3 that when $\eta_0 \leq \min\{\lambda_0(1-\gamma),\frac{1-\gamma}{(1+\gamma)^2}\}$ and $\alpha \in [\frac{1}{2},1)$,%the convex distance between $\sqrt{T}\bar{\bm{\Delta}}_T$ and the Gaussian distribution with variance matching the asymptotic variance $\bm{\Lambda}$ is
\begin{align*}
	d_{\mathsf{C}}(\sqrt{T}\bar{\bm{\Delta}}_T,\mathcal{N}(\bm{0},\bm{\Lambda}^\star)) \lesssim O(T^{-\frac{\alpha}{2}} + T^{\frac{\alpha-1}{2}}).
\end{align*}
Choosing  $\alpha = \frac{1}{2}$ leads to a convergence rate in $T$ of order $O(T^{-\frac{1}{4}})$. In contrast, Theorem~\ref{thm:Berry-Esseen-combined} shows that, by selecting  $\alpha = \frac{2}{3}$, we obtain the faster rate $O(T^{-\frac{1}{3}})$.
% \begin{align*}
% d_{\mathsf{C}}(\sqrt{T}\bar{\bm{\Delta}}_T,\mathcal{N}(\bm{0},\bm{\Lambda}^\star))\leq O(T^{-\frac{1}{3}}).
% \end{align*}
% The proof of Theorem \ref{thm:Berry-Esseen} relies on Theorem 3.1 of \citet{shao2021berry}, a high-dimensional Berry-Esseen bound for non-linear statistics,  which is also used by \citet{samsonov2024gaussian} to establish their main result about Gaussian approximations for the TD error, Theorem 3.4 therein. 
As demonstrated in Appendix \ref{sec:tightness.thm:asymptotic}, this rate is tight for $\alpha \in (\frac{2}{3},1)$, since the $O(T^{\alpha-1})$ term is inevitable.  See also Figure~\ref{fig:Berry-Esseen} below, showing the output of a numerical experiment indicating that, in fact, the choice of $\alpha = \frac{2}{3}$ leads to a smaller approximation error. %\footnote{The tightness for $\alpha \in (\frac{1}{2},\frac{2}{3}]$ remains an open question.}. 
Though worked out independently, the proofs of our Theorem~\ref{thm:Berry-Esseen-combined} and of Theorem 3 of \cite{samsonov2024gaussian} bear similarities, as they both rely on the general Berry-Esseen bounds for non-linear statistics developed by \cite{shao2021berry}. We obtain sharper rates by a more targeted analysis of the TD settings, while, in comparison, the results of \citet{samsonov2024gaussian} apply more broadly to SA problems. Specifically we (i) adopt a different decomposition of the TD estimation error, (ii)  handle in a more refined way certain terms appearing in the bound of Theorem 3.1 of \cite{shao2021berry}\footnote{Specifically, the term $I_3$ in Appendix \ref{sec:proof-thm-Berry-Esseen}.} and (iii) carry out a more specialized analysis of the properties of the matrices $\bm{A}_t$ and $\bm{A}$ (see, e.g., Lemma~\ref{lemma:A} in the Appendix) originating from TD learning, which has allowed us to tighten problem-related coefficient, as well as to work with a substantially larger initial stepsize $\eta_0$. This non-trivial sharpening may be potentially generalized to analyze stochastic gradient algorithms applied to non-linear functions, as, e.g., in Theorem 3.4 of \citet{shao2021berry}. %We leave this extension to future work. It is also noteworthy that through our analysis of the specific structure of matrices $\bm{A}_t$ and $\bm{A}$ generated from TD learning mechanism, we are able to tighten problem-related coefficient in the bound, as well as allow for a substantially larger intial stepsize $\eta_0$, which could fasten the convergence in practice. 

After the appearance of our preprint, \cite{butyrin2025improved} obtained analogous Berry–Esseen rates for linear stochastic approximation using similar techniques.

\subsection{High-probability bounds}

Our next result presents a high-dimensional, high-probability bound on the size of the TD estimation error. 
% The following theorems provide high-probability guarantees for the convergence rate of the TD estimation error.

%\begin{align}
%& \overline{\bm{\Lambda}}_T=\frac{1}{T}\sum_{t=1}^T \bm{Q}_t \bm{\Gamma}\bm{Q}_t^\top, \quad \text{where} \label{eq:defn-barLambdaT}\\
%& \bm{Q}_t = \eta_t \sum_{j=t}^{T}\prod_{k=t+1}^{j} (\bm{I}-\eta_k \bm{A})\label{eq:defn-Qt}.
%&\end{align}

%. In our bound, as well through the paper, we w result with its proof deferred to Section~\ref{app:proofs}. \yuting{add ...}\weichen{checked.}

\begin{theorem}
\label{thm:bar-delta-t}
Consider the TD algorithm with Polyak-Ruppert averaging~\eqref{eq:TD-update-all}, independent samples and decaying stepsizes $\eta_t = \eta_0 t^{-\alpha}$ for $\alpha \in (\frac{1}{2},1)$.
For any $\delta \in (0,1)$, there exists $\eta_0 = \eta_0(\delta)> 0$, such that %the estimation error $\bar{\bm{\Delta}}_T$ (cf.~\eqref{eq:defn-bar-deltat}) satisfies 
\begin{align*}
	\|\overline{\bm{\Delta}}_T\|_2 &\lesssim \sqrt{\frac{\mathsf{Tr}(\bm{\Lambda}^\star)}{T}\log \frac{1}{\delta}} + \tilde{C} (T^{-\frac{\alpha+1}{2}}+T^{\frac{\alpha}{2}-1})\log\frac{T}{\delta}
\end{align*}
with probability at least $1-\delta$, where $\tilde{C}$ is a problem-related quantity independent of $T$ and $\delta$. 
\end{theorem}

The proof of this result can be found in Appendix~\ref{app:proof-whp}, where the value of $\tilde{C}$ is given in equation \eqref{eq:bar-delta-t-tilde-C}. Throughout the paper, for the sake of readability,  we have opted to express unwieldy problem-dependent quantities that do not directly depend on $T$ in implicit form; their values can be tracked in the proofs. Accordingly, when writing our finite-sample bounds, we will first present the leading terms in $T$.   % is in Appendix \ref{app:proof-bar-delta-t-iid}.

%\yuting{maybe call it TD with Polyak-Ruppert averaging instead of the averaged TD algorithm}

% \yuting{I am slightly torn on whether we should state our main result on Berry-Essen bound first and unpack the other results later, but I guess both order can work}
In a nutshell, Theorem~\ref{thm:bar-delta-t} states that $\bar{\bm{\theta}}_T$ converges to $\bm{\theta}^*$ (in Euclidean distance) at the rate of order $O(\sqrt{{\mathsf{Tr}(\bm{\Lambda}^\star)}/{T}})$, up to a log factor. It is worth comparing the high-probability sample complexity bound of Theorem~\ref{thm:bar-delta-t} with those in the literature. To the best of our knowledge, the tightest results were attained by \cite{dalal2018finite} for inverse-decaying stepsizes and \cite{li2023sharp,samsonov2023finitesample} for constant stepsize as summarized in Table~\ref{table:sample-complexity}. 
Theorem \ref{thm:bar-delta-t} is the first result to achieve tight high-probability convergence guarantees up to log factors using polynomial-decaying stepsizes. Another subtle but consequential difference concerns the restriction on the step size. For our bound to hold, the choice of the initial stepsize $\eta_0$ depends on $\delta$ -- specifically, $\eta_0$ is required to be sufficiently small such that $\eta_0 \sqrt{ \log {1}/(\delta\eta_0}) \leq \tilde{C}$, where $\tilde{C}$ is independent of $T$ and $\delta$ but depends on the other parameters of interest (its specific form is shown in \eqref{eq:defn-tstar-alpha} and \eqref{eq:delta-t-condition} in the Appendix). Though not ideal, this condition is necessary for a Chernoff-type convergence guarantee; we refer to Appendix \ref{app:delta-t-condition} for a detailed discussion. Nonetheless, our stepsize requirements are considerably weaker than those imposed by \cite{samsonov2023finitesample} and \cite{li2023sharp}, which depend on both $\delta$ and $T$; see Table~\ref{table:sample-complexity} in Appendix~\ref{sec:morecomparions}.

As it turns out, the dependence of $\eta_0$ on the target probability coverage $\delta$ can be removed in the special case of {\it tabular settings} in which the state space is finite. See Appendix \ref{sec:morecomparions}, where we also discuss related (but not applicable to our problem!) high probability bounds for TD-learning in tabular settings by \cite{chen2024concentration} and \cite{khodadadian2025general}, in which the stepsize is also independent of the choice of $\delta$.

\subsection{Online estimation of the covariance}\label{sec:online}

In the last section, we obtained finite sample bounds on the convex distance between $\sqrt{T}\bar{\bm{\Delta}}_T$ and its Gaussian approximation, which features an {\it unknown} variance. 
To construct confidence regions for $\bm{\theta}^\star$, it is necessary to estimate the covariance matrix $\bm{\Lambda}^\star$. 
Two types of approaches have recently been proposed:  \emph{plug-in estimation} \citep{chen2020statistical} and \emph{bootstrapping} \citep{white2010interval, hanna2017bootstrapping,hao2021bootstrapping,ramprasad2022online,Zhu02012023,samsonov2024gaussian}. %nontheless, the statistical and computational efficiency of these estimations can both be further enhanced.

% Theorems \ref{thm:Berry-Esseen} and \ref{thm:asymptotic} measures the non-asymptotic rate of convergence for the variance-covariance matrix of the averaged TD estimation error $\bar{\bm{\Delta}}_T$. 
Here we propose to use a computationally feasible online plug-in estimator for $\bm{\Lambda}^\star$ with superior computational efficiency to existing methodologies and establish its non-asymptotic rate of convergence. Specifically, we consider the following 
\vspace{-0.3cm}
\begin{align}
&\overline{\bm{A}}_T = \frac{1}{T}\sum_{t=1}^{T} \bm{A}_t, \quad \text{and} \\ 
&\widehat{\bm{\Gamma}}_T = \sum_{t=1}^T w_{t,T}(\bm{A}_t \overline{\bm{\theta}}_t - \bm{b}_t)(\bm{A}_t \overline{\bm{\theta}}_t - \bm{b}_t)^\top \label{eq:defn-hat-Gamma}
\end{align}
as estimators for $\bm{A}$ and $\bm{\Gamma}$ respectively, where the \emph{weights} $\{w_{t,T}\}_{1 \leq t \leq T}$ in $\widehat{\bm{\Gamma}}_T$ are chosen as 
\vspace{-0.1cm}
\begin{align*}
w_{t,T} := \frac{t}{\sum_{t=1}^T t} = \frac{2t}{T(T+1)},
\end{align*}
following the intuition that as $t$ grows larger, $\bar{\bm{\theta}}_{t}$ is a more accurate estimator of $\bm{\theta}^\star$ and hence the corresponding term should be assigned higher weights. We then consider
\begin{align}\label{eq:defn-hat-Lambda}
\widehat{\bm{\Lambda}}_T = \overline{\bm{A}}_T^{-1}\widehat{\bm{\Gamma}}_T \overline{\bm{A}}_T^{-\top} 
\end{align}
as the final estimator of $\bm{\Lambda}^\star$. 

%\yuting{use consistent notation for $\widehat{\bm{\Lambda}}_T$ and $\hat{\bm{\Lambda}}_T$}

%\paragraph{Online calculation of the estimator $\hat{\bm{\Lambda}}_{T}$.} 

Crucially, the estimator $\hat{\bm{\Lambda}}_T$ can be evaluated in an online fashion. 
See Algorithm~\ref{alg:online} in Appendix~\ref{app:online-alg} for a detailed description of the procedure, which does \emph{not} require a priori knowledge of the total round of iterations $T$, and has time and space complexities of order $O(Td^2+d^3)$ and $O(d^2)$, respectively. 
The estimator $\hat{\bm{\Lambda}}_T$ also enjoys nice statistical properties. This is made precise by the following Gaussian comparison result, showing that the unknown asymptotic variance $\bm{\Lambda}^\star$ in the Berry-Esseen bound of Theorem~\ref{thm:asymptotic} can be replaced by $\hat{\bm{\Lambda}}_T$ without compromising the accuracy of the approximation (i.e., without affecting the rate in $T$).
The proof of this result can be found in Appendix~\ref{app:proof-plug-in-iid}. 

\begin{theorem}\label{thm:plug-in-iid}
Consider the TD algorithm \eqref{eq:TD-update-all} with independent samples and decaying stepsizes $\eta_t = \eta_0 t^{-\alpha}$ for all $t > 0$, any $\alpha \in (1/2,1)$ and $\eta_0 < \frac{1}{4}$ , and let $\widehat{\bm{\Lambda}}_T$, $\bm{\Lambda}^\star$ be defined as in \eqref{eq:defn-hat-Lambda}, \eqref{eq:defn-Lambdastar} respectively. Then, 
%\ifarxiv
\begin{align*}
&d_{\mathsf{TV}}(\mathcal{N}(\bm{0},\hat{\bm{\Lambda}}_T),\mathcal{N}(\bm{0},\bm{\Lambda}^\star)) \lesssim \frac{\sqrt{\mathsf{Tr}(\bm{\Gamma})}}{\lambda_0(1-\gamma)}(\|\bm{\theta}^\star\|_2+1)\|\bm{\Gamma}^{-1}\| T^{-\frac{1}{3}} + o(T^{-\frac{1}{3}}).
\end{align*}
% \else
% \begin{align*}
% &d_{\mathsf{TV}}(\mathcal{N}(\bm{0},\hat{\bm{\Lambda}}_T),\mathcal{N}(\bm{0},\bm{\Lambda}^\star)) \\ 
% &\lesssim \frac{\sqrt{\mathsf{Tr}(\bm{\Gamma})}}{\lambda_0(1-\gamma)}(\|\bm{\theta}^\star\|_2+1)\|\bm{\Gamma}^{-1}\| T^{-\frac{1}{3}} + o(T^{-\frac{1}{3}}).
% \end{align*}
% \fi
with probability at least $1-3T^{-\frac{1}{3}}-\frac{\sqrt{\mathsf{Tr}(\bm{\Gamma})}}{\lambda_0(1-\gamma) \sqrt{T}}$.
% {\color{red}With the new estimator $\check{\bm{\Gamma}}_T$, there will be an additional $\log^{1/3} T$ term in the upper bound.}
% where $\tilde{f}$ depends on problem-related quantities and converges by the rate of $o(T^{-\frac{1}{3}})$ as a function of $T$.
\end{theorem}
%\yuting{change $\tilde{f}$}\weichen{checked.}
Above, the term $o(T^{-\frac{1}{3}})$ depends on all the parameters of the problem (and, in particular, on $\alpha$) but, as a function of $T$  alone, it is of  order  $O(T^{-\frac{1}{3}})$. See Appendix~\ref{app:proof-plug-in-iid} for the proof and details.
%\noindent\emph{Proof:} See Appendix \ref{app:proof-plug-in-iid}.\qed

{\bf Comparison with \citet{chen2020statistical}.} 
In different settings, \citet{chen2020statistical} considered the problem of carrying out statistical inference for the true model parameters using stochastic gradient descent and assuming a strongly convex and smooth population loss function. 
In their work, they consider the slightly different plug-in estimator $\hat{\bm{\Lambda}}_T'=\bar{\bm{A}}_T^{-1}\hat{\bm{\Gamma}}_T' \bar{\bm{A}}_T^{-\top}$, where $\hat{\bm{\Gamma}}_T' := \frac{1}{T} \sum_{t=1}^T (\bm{A}_t {\bm{\theta}}_t - \bm{b}_t)(\bm{A}_t {\bm{\theta}}_t - \bm{b}_t)^\top$.
%\begin{align}
%	\label{eq:defn-Chen-hat-Gamma}
%	\hat{\bm{\Gamma}}_T' := \frac{1}{T} \sum_{t=1}^T (\bm{A}_t {\bm{\theta}}_t - \bm{b}_t)(\bm{A}_t {\bm{\theta}}_t - \bm{b}_t)^\top,
%\end{align}
%is proposed, which differs slightly from ours. 
The authors proved that the estimation error of $\hat{\bm{\Lambda}}_T'$ is bounded in expectation as
\begin{align}\label{eq:Chen-plugin}
\mathbb{E}\|\hat{\bm{\Lambda}}_T' - \bm{\Lambda}^\star\| \leq \|\bm{\Gamma}\| \cdot (\widetilde{C}_1 T^{-\frac{\alpha}{2}} + \widetilde{C}_2 T^{-\alpha}),
\end{align}
for problem-related constants $\widetilde{C}_1$ and $\widetilde{C}_2$. 
It is possible to modify the proof of Theorem \ref{thm:plug-in-iid} to obtain a guarantee in expectation for our problem as well, albeit in the TV metric. Specifically, we can show that 
\begin{align}\label{eq:E-dTV-plug-in-iid}
\mathbb{E}\left[d_{\mathsf{TV}}(\mathcal{N}(\bm{0},\hat{\bm{\Lambda}}_T),\mathcal{N}(\bm{0},\bm{\Lambda}^\star))\right]&\lesssim \widetilde{C}\|\bm{\Gamma}\|T^{-\frac{1}{2}}
\end{align}
where $\widetilde{C}$ depends on $\lambda_0,\gamma,\eta_0,\alpha$ and $\bm{\theta}^\star$.
Comparing  \eqref{eq:Chen-plugin} with \eqref{eq:E-dTV-plug-in-iid}, we have sharpened the rate from $O\left(T^{-\frac{\alpha}{2}}\right)$ to $O\left(T^{-\frac{1}{2}}\right)$, {\it independent of} the choice of $\alpha$. 
%This is a significant improvement, at least for our problem, not only because of the different metric but also because a root-$T$ rate based on \eqref{eq:Chen-plugin} would require $\alpha = 1$, which is not permissible in our settings.

%the resulting rate happens to match the one obtained by choosing $\alpha = 2/3$ in Theorems~\ref{thm:Berry-Esseen}. and \ref{thm:asymptotic}. and ~ref, at least in our present problem, because it dispenses us from choosi. Thus to be $2/3$ not only because $\alpha$ is strictly less than 1 in our setting, but also because taking $\alpha=1$ would affect the Berry-Esseen bound shown in Theorem \ref{thm:Berry-Esseen}.
%\yuting{Setting $\alpha = 1$ recovers the $-1/2$ rate no? not sure if we should focus on the distributions bound v.s. tv bound. rewrite.}\weichen{Removed the point of TV distance.}
% It is noteworthy that the online Algorithm \ref{alg:online} takes advantage of the linearity of the problem; in general, potentially non-linear settings, replacing $\bm{\theta}_t$ in $\hat{\bm{\Gamma}}_T' $ by $\bar{\bm{\theta}}_T$ in $\hat{\bm{\Gamma}}_T$ would require storing all the samples throughout the iterative trajectory, resulting in a larger space complexity.
\paragraph{Comparison with \citet{samsonov2024gaussian}.} Leveraging the multiplier-bootstrap technique proposed by \cite{Zhu02012023}, \citet{samsonov2024gaussian} construct a randomized estimator $\bar{\bm{\theta}}_T^{(b)}$ based on the samples $\mathcal{S}_T:=\{s_t,s_t'\}_{t=1}^T$. Their Theorems 4 and 5 yield that  %imply that, in our notation, %that \textcolor{violet}{*ALE*: just wanted to make sure this is correct. Their paper does not contain such result} \textcolor{blue}{weichen: Both Theorems appear in their appendix. Also, I addded a $o(T^{-1/4})$ term.}
%\ifarxiv 
\begin{align}
&d_{\mathsf{C}}(\sqrt{T}(\bar{\bm{\theta}}_T^{(b)}-\bm{\theta}^\star)| \mathcal{S}_T, \mathcal{N}(\bm{0},\bm{\Lambda}^\star))\lesssim \frac{\|\bm{\Gamma}^{-1}\|}{[\lambda_0(1-\gamma)]^{\frac{3}{2}}}(2\|\bm{\theta}^\star\|_2+1)^2 \frac{\log T}{T^{\frac{1}{4}}} + o(T^{-\frac{1}{4}})\label{eq:bootstrap-error}
\end{align}
% \else
% \begin{align}
% &d_{\mathsf{C}}(\sqrt{T}(\bar{\bm{\theta}}_T^{(b)}-\bm{\theta}^\star)| \mathcal{S}_T, \mathcal{N}(\bm{0},\bm{\Lambda}^\star))\nonumber \\ 
% &\lesssim \frac{\|\bm{\Gamma}^{-1}\|}{[\lambda_0(1-\gamma)]^{\frac{3}{2}}}(2\|\bm{\theta}^\star\|_2+1)^2 \frac{\log T}{T^{\frac{1}{4}}} + o(T^{-\frac{1}{4}})\label{eq:bootstrap-error}
% \end{align}
% \fi
%\yuting{can we get rid of the term $ o\left(\frac{\log T}{T^{\frac{1}{4}}}\right)$ and make it $\lesssim$?}\weichen{Yes, changed the formula.}
Compared with this bootstrap-based procedure, our methodology is advantageous in the following ways: (i) the upper bound in Theorem \ref{thm:plug-in-iid} vanishes, as a function of $T$, at the rate of $O(T^{-\frac{1}{3}})$, which is faster than the $O(T^{-\frac{1}{4}}\log T)$  rate from the bound \eqref{eq:bootstrap-error}. (ii) The leading term in the upper bound in Theorem \ref{thm:plug-in-iid} entails a smaller problem-related coefficient than that of \eqref{eq:bootstrap-error}. (iii) Perhaps most importantly, in order to estimate the bootstrap distribution %$\sqrt{T}(\bar{\bm{\theta}}_T^{(b)}-\bm{\theta}^\star)|\mathcal{S}_T$ 
with an $O(T^{-\frac{1}{4}})$ error rate, it is necessary to store the sample trajectory $\mathcal{S}_T$ and generate $O(T^{\frac{1}{2}})$ perturbed trajectories. Consequently, the bootstrap procedure has a time complexity of order $O(T^{\frac{3}{2}}d^2)$ and a space complexity of order $O(T)$. This amounts to a much higher computational burden than our online Algorithm~\ref{alg:online}.
%Comparing the theoretical guarantees (i.e., the bound~\eqref{eq:bootstrap-error}) and the methodology of \cite{samsonov2024gaussian} with our Theorem \ref{thm:plug-in-iid}, our plug-in estimator $\hat{\bm{\Lambda}}_T$ enjoys at least three advantages: \yuting{two or three? :)} \weichen{yes, three.}
%\begin{enumerate}
%\item The upper bound in Theorem \ref{thm:plug-in-iid} vanishes, as a function of $T$, at the rate of $O(T^{-\frac{1}{3}})$, which is faster than the $O(T^{-\frac{1}{4}}\log T)$ convergence rate obtained from the bound \eqref{eq:bootstrap-error}. 
%\item The leading term in the upper bound in Theorem \ref{thm:plug-in-iid} entails a smaller problem-related coefficient than that of \eqref{eq:bootstrap-error}. 
%\item In order to estimate the bootstrap distribution $\sqrt{T}(\bar{\bm{\theta}}_T^{(b)}-\bm{\theta}^\star)|\mathcal{S}_T$ with an $O(T^{-\frac{1}{4}})$ error rate, one needs to store the sample trajectory $\mathcal{S}_T$ and generate $O(T^{\frac{1}{2}})$ perturbed trajectories. Consequently, the bootstrap procedure has a time complexity of order $O(T^{\frac{3}{2}}d^2)$ and a space complexity of order $O(T)$. When $d \ll T$, this results in a significant computational border compared to our online procedure of Algorithm~\ref{alg:online}.
%\end{enumerate}

\subsection{Statistical inference}\label{sec:inference}

Combining the results from Sections~\ref{sec:berry-esseen} and~\ref{sec:online}, we describe a general methodology for constructing convex confidence sets for $\bm{\theta}^\star$ with guaranteed finite sample coverage. Based on the TD estimator $\bar{\bm{\theta}}_T$ in \eqref{eq:TD-averaging} and the variance estimator $\hat{\bm{\Lambda}}_T$ \eqref{eq:defn-hat-Lambda}, and given any tolerance level $\delta \in (0,1)$, it is enough to pick any convex set $\mathcal{C}_{\delta}$ that has probability at least $1- \delta$ under the Gaussian distribution $\mathcal{N}(\bar{\bm{\theta}}_T,\frac{\hat{\bm{\Lambda}}_T}{T})$ with random mean and covariance; that is, satisfying the condition
\begin{align}
\mathbb{P}_{\bm{\theta} \sim \mathcal{N}(\bar{\bm{\theta}}_T,\frac{\hat{\bm{\Lambda}}_T}{T})} \left(\bm{\theta} \in \mathcal{C}_{\delta} \bigg| \bar{\bm{\theta}}_T,\hat{\bm{\Lambda}}_T \right) \geq 1-\delta. \label{eq:convex.Cdelta}
\end{align}
Natural choices of $\mathcal{C}_{\delta}$ are (i) hyper-rectangles, leading to \emph{simultaneous confidence intervals} of the form
\begin{align}\label{eq:multi-CI}
\mathcal{C}_{\delta} = \prod_{j=1}^d [(\bar{\bm{\theta}}_T)_j - \hat{R}/\sqrt{T}, (\bar{\bm{\theta}}_T)_j + \hat{R}/\sqrt{T}],
\end{align}
where $\hat{R}>0$ is the (random) quantile of the $\ell_{\infty}$ norm of a centered Gaussian random vectors with covariance $(\hat{\bm{\Lambda}}_T/{T})$, which can be calculated via simulations; and (ii) \emph{ellipsoids} or \emph{balls}, for example
\begin{align}\label{eq:ellipsoid-region}
\mathcal{C}_{\delta} = \left\{\bm{\theta}: \sqrt{T}\|\bm{\theta} - \bar{\bm{\theta}}_T\|_{\hat{\bm{\Lambda}}_T^{-1}}^2 \leq R\right\},
\end{align}
where $R$ is the $1-\delta$ quantile of the $\chi^2(d)$ distribution.

% \begin{enumerate}
% \item \emph{Confidence intervals for individual entries}. A confidence interval for the $i$-th entry of $\bm{\theta}^\star$ can be simply prescribed as
% \begin{align*}
% \text{CI}_i = \left[(\bar{\bm{\theta}}_T)_i - t\cdot \sqrt{(\hat{\bm{\Lambda}}_T)_{ii}/{T}}, (\bar{\bm{\theta}}_T)_i + t\cdot \sqrt{(\hat{\bm{\Lambda}}_T)_{ii}/{T}}\right]
% \end{align*}
% where $t$ is the $1-\delta/2$ quantile of the standard normal distribution. This is equivalent to setting \footnote{In this section, $\times$ and $\prod$ represent Cartesian product.}
% \begin{align}\label{eq:uni-CI}
% \mathcal{C}_{\delta} = \prod_{j=1}^{i-1} (-\infty,\infty) \times \text{CI}_i \times \prod_{j=i+1}^d (-\infty,\infty).
% \end{align}
% \item \emph{Simultaneous confidence intervals}. In order to obtain simultaneous confidence intervals for all the entries of $\bm{\theta}^\star$, one may consider the hyper-rectangle
% \begin{align}\label{eq:multi-CI}
% \mathcal{C}_{\delta} = \prod_{j=1}^d [(\bar{\bm{\theta}}_T)_j - t/\sqrt{T}, (\bar{\bm{\theta}}_T)_j + t/\sqrt{T}],
% \end{align}
% where $t>0$ is the quantile of the $\ell_{\infty}$ norm of a centered Gaussian random vectors with covariance $(\hat{\bm{\Lambda}}_T/{T})$, which can be calculated via simulations.
% \item \emph{Ellipsoid confidence regions}: For example,
% \begin{align}\label{eq:ellipsoid-region}
% \mathcal{C}_{\delta} = \left\{\bm{\theta}: \sqrt{T}\|\bm{\theta} - \bar{\bm{\theta}}_T\|_{\hat{\bm{\Lambda}}_T^{-1}}^2 \leq t\right\},
% \end{align}
% where $t$ is the $1-\delta$ quantile of the $\chi^2(d)$ distribution.
% \end{enumerate}

The next result, whose proof is in Appendix~\ref{app:proof-inference-iid}, shows that the coverage of $\mathcal{C}_{\delta}$ converges to the nominal level of $ 1- \delta$ at a rate $O(T^{-1/3})$, as a function of $T$ alone.

\begin{theorem}
\label{cor:confidence-region}
Consider TD with Polyak-Ruppert averaging~\eqref{eq:TD-update-all} with independent samples and decaying stepsizes $\eta_t = \eta_0 t^{-\alpha}$, for all $t > 0$ with $\alpha = 2/3$ and $\eta_0 < \frac{1}{4}$.%\in (\frac{1}{2},1)$.
 For a given significance level $\delta \in (0,1)$, it holds that
%\ifarxiv 
\begin{align*}
\mathbb{P}(\bm{\theta}^\star \in \mathcal{C}_{\delta}) &\geq 1-\delta - \frac{\sqrt{\mathsf{Tr}(\bm{\Gamma})}}{\lambda_0(1-\gamma)}(2\|\bm{\theta}^\star\|_2+1)\|\bm{\Gamma}^{-1}\| T^{-\frac{1}{3}} -\frac{\sqrt{d\mathsf{cond}(\bm{\Gamma})}}{2(1-\gamma)\lambda_0\eta_0} T^{-\frac{1}{3}}  - o(T^{-\frac{1}{3}}),
\end{align*}
% \else
% \begin{align*}
% \mathbb{P}(\bm{\theta}^\star \in \mathcal{C}_{\delta}) &\geq 1-\delta - \frac{\sqrt{\mathsf{Tr}(\bm{\Gamma})}}{\lambda_0(1-\gamma)}(2\|\bm{\theta}^\star\|_2+1)\|\bm{\Gamma}^{-1}\| T^{-\frac{1}{3}} \\ 
% &-\frac{\sqrt{d\mathsf{cond}(\bm{\Gamma})}}{2(1-\gamma)\lambda_0\eta_0} T^{-\frac{1}{3}}  - o(T^{-\frac{1}{3}}),
% \end{align*}
% \fi
uniformly over all convex sets  $\mathcal{C}_\delta \subset \mathbb{R}^d$ satisfying \eqref{eq:convex.Cdelta}.
\end{theorem}
% {\color{red}With the new estimator $\check{\bm{\Gamma}}_T$, there will be an additional $\log^{1/3} T$ term in the first term after $1-\delta$.}

%\yuting{in our final version, we can include a box plot to show coverage on the true parameter}\weichen{Yes, will include the plot for coverage rate.}

We highlight two important features of the above result. First, the coverage guarantee holds {\it uniformly} over all choices of the convex set $\mathcal{C}_\delta$ meeting condition \eqref{eq:convex.Cdelta}, such as those in \eqref{eq:multi-CI} and \eqref{eq:ellipsoid-region}. Secondly, the size (i.e., volume) of the confidence sets vanishes at the standard (and optimal) parametric rate of $O(T^{-1/2})$, keeping all the other variables constant. %This suggests that, in this regard, there is little or no room for improvements, at least in terms of rates.  
%In particular, this class includes all hyper-rectangles (cartesian products of univariate intervals), returning simultaneous confidence intervals that are especially useful due to their simplicity and interpretability. The construction of $\mathcal{C}_\delta$  requires computing appropriate quantiles of the $\mathcal{N}(\bar{\bm{\theta}}_T ,\hat{\bm{\Lambda}}_T) $ distribution, conditionally  on $\bar{\bm{\theta}}_T$ and $\hat{\bm{\Lambda}}_T$, depending on the choice of convex set. For the case of hyper-rectangles, this task is reduced to finding the quantile of the maximum of Gaussian  variables. %This can be done using simulations or approximations - for example, for thge case of hyper-rectangles, one can use e.g. Bonferroni or Sidak's) 

\section{Simulations}\label{sec:simulation}

We corroborate our theoretical findings with numerical experiments. Towards that goal, we consider a family of MDPs that are constructed to be difficult to distinguish from each other. The specification of this family of MDPs, along with detailed descriptions of our experiments, is provided in Appendix \ref{app:experiment-details}. 
Here we summarize some of our results. We ran $10,000$ independent trials, each for $T = 2^{20}$ iterations, and recorded data at iterations $2^1,2^2,2^3,...,2^{19},2^{20}$. We report the rate of convergence of $\|\bar{\bm{\Delta}}_T\|_2^2$, the rate of converge of $\bar{\bm{\Delta}}_T$ to its asymptotic Gaussian approximation, and the precision of our estimator for the covariance matrix, under different choices of the parameter $\alpha = \frac{1}{2},\frac{2}{3},\frac{3}{4},1$.
The results of these experiments confirm that the convergence speed is in accordance with our theoretical findings. In particular, for the choice of $\alpha= \frac{2}{3}$, the statistical inference procedure described in Section \ref{sec:inference} achieves the target coverage rate when $T$ is sufficiently large. %We refer the reader to Appendix \ref{app:experiment-details} for a full description of our experiments and of our results.

\begin{figure}[t]
\centering
%\ifarxiv
\includegraphics[width = 0.48\textwidth]{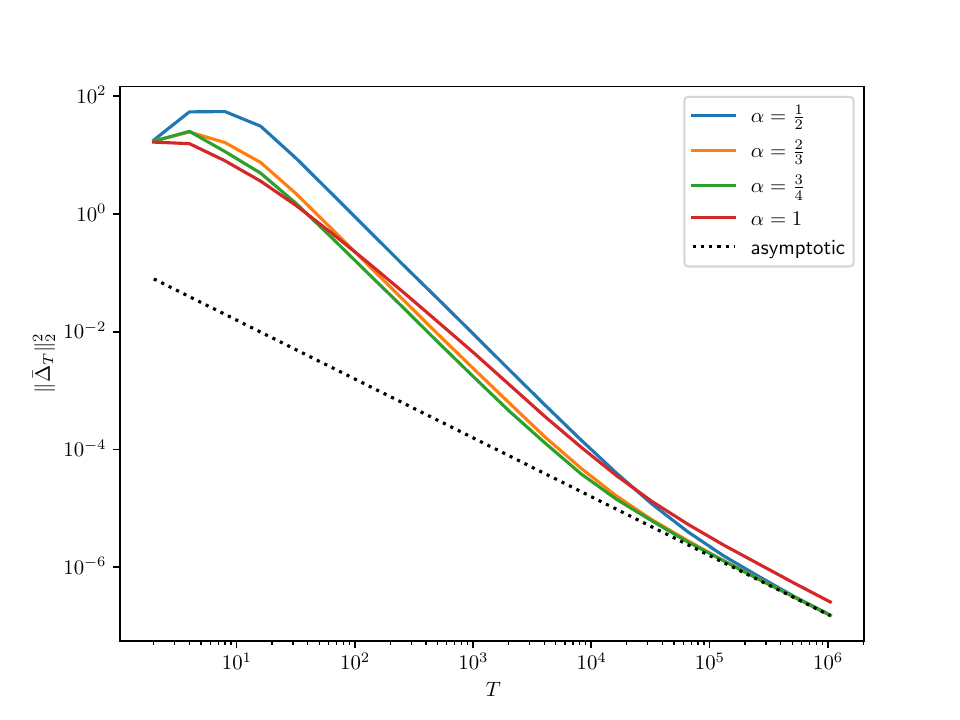} 
\includegraphics[width = 0.48\textwidth]{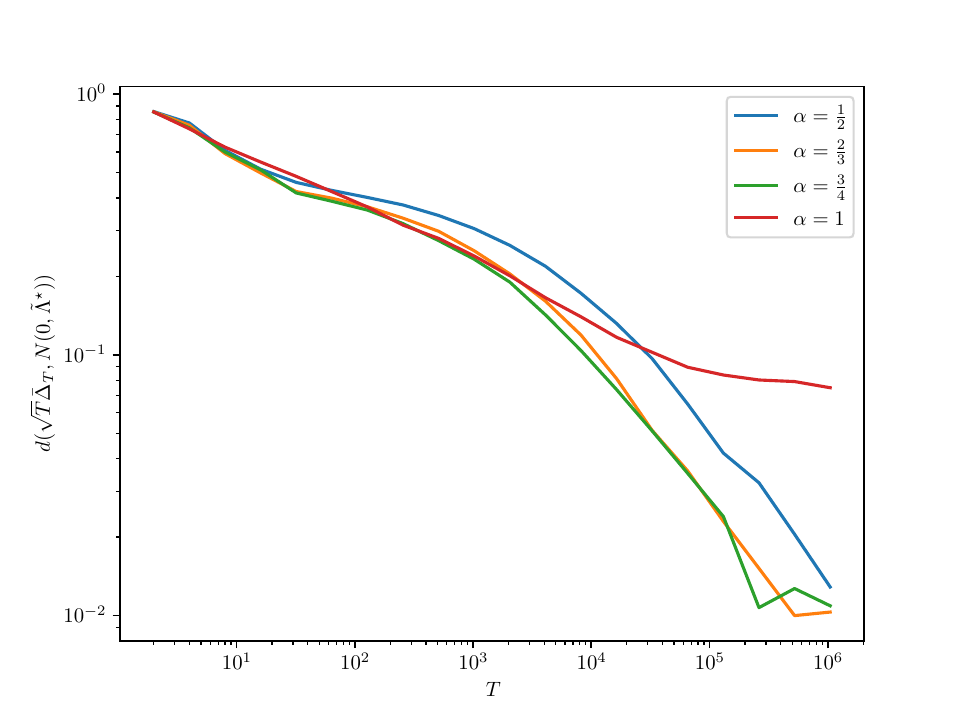} 
% \else
% \includegraphics[width = 0.4\textwidth]{imgs/figure_3a.pdf} \\
% \includegraphics[width = 0.4\textwidth]{imgs/figure_2.pdf} 
% \fi
\caption{(a) Empirical $95\%$ quantiles of $\|\bar{\bm{\Delta}}_T\|_2$, and (b) distance between the empirical and asymptotic distributions of $\bar{\bm{\Delta}}_T$.}
% See \ref{app:experiment-details} for details.}
\label{fig:Berry-Esseen}
\end{figure}

Figure~\ref{fig:Berry-Esseen}(a) illustrates the behavior of $\|\bar{\bm{\Delta}}_T\|_2^2$ under different choices of $\alpha$. As a reference, the black dotted line represents the $95\%$ quantile of $\{\bm{Z}_T\}$ where $\bm{Z}_T \sim \mathcal{N}(\bm{0},\bm{\Lambda}^\star/T)$. Our theory predicts that when $T$ grows large, this should be the asymptotic line for the empirical $95\%$ quantiles when $\alpha \in (\frac{1}{2},1)$. This is indeed what is shown for $\alpha=\frac{2}{3}$ and $\frac{3}{4}$. When $\alpha = 1$, however, the empirical quantiles do not appear to converge to the asymptotic quantiles. 
% This is consistent with our theory: since $\bar{\bm{\Lambda}}_T-\bm{\Lambda}^\star = \bm{O}(T^{\alpha-1})$, the variance of $\bar{\bm{\Delta}}_T$ will \emph{not} converge to $\bm{\Lambda}^\star$ when $\alpha=1$. 
This fact is also consistent with our theory, as reflected in the $O(T^{\frac{\alpha}{2}-1})$ term in the bound of Theorem \ref{thm:bar-delta-t}, which when $\alpha = 1$, converges at the $O(T^{-\frac{1}{2}})$ rate, just like the leading term. As for the case of $\alpha=\frac{1}{2}$, we observe that when $T$ is small,  $\|\bar{\bm{\Delta}}_T\|_2$ is relatively large and the empirical and asymptotic lines begin to overlap only when $T$ is very large (i.e., $T > 10^5$). This phenomenon is also predicted by the $O(T^{-\frac{\alpha+1}{2}})$ term in the bound of Theorem \ref{thm:bar-delta-t}, which vanishes at a slower rate when $\alpha = \frac{1}{2}$. Notice that for ease of computation, we use a different measure of the distributional difference than the convex distance, as specified in Eq.~\eqref{eqn:distance-measure-fig2} in Appendix \ref{app:experiment-details}.

Figure~\ref{fig:Berry-Esseen}(b) depicts the speed of convergence of the empirical distribution of the rescaled TD estimation error $\sqrt{T} \bar{\bm{\Delta}}_T$ to its asymptotic distribution $\mathcal{N}(\bm{0},\bm{\Lambda}^\star)$. 
% For ease of computation, instead of the convex distance we evaluated the distance%between the distributions 
% \begin{align}
% \label{eqn:distance-measure-fig2}
% %\notag &d(\sqrt{T}\bar{\bm\Delta}_T,\mathcal{N}(\bm{0},\bm{\Lambda}^\star)) \\ 
% %&:= 
% \max_{1 \leq j \leq d} \sup_{x \in \mathbb{R}} \left|\hat{\mathbb{P}}(\sqrt{T}(\bar{\bm\Delta}_T)_j \leq x) - \mathbb{P}_{\bm{z}}((\bm{\Lambda}^{\star \frac{1}{2}}\bm{z})_j \leq x)\right|,
% \end{align}
% where $\hat{\mathbb{P}}$ refers to the empirical distribution in our $10,000$ trials, and $\bm{z}$ is the $d$-dimensional standard Gaussian.
When $\alpha = 1$, this distance does \emph{not} converge to $0$ as $T$ grows large, attesting to our claim that the $O(T^{\alpha-1})$ term in the distributional difference is inevitable. On the other hand, choosing $\alpha = \frac{1}{2}$, as done in \citet{samsonov2024gaussian}, leads to  a slower convergence than with $\alpha = \frac{2}{3}$ (our choice) and $\alpha = \frac{3}{4}$.

% \textcolor{violet}{ALE: I think we can include more plots, or details/comments about the simulations to fill out out the 8 page limits. Weichen: do you have some plots or results that you find especially enlightening?}

%While we postpone most of the experimental results to Appendix \ref{app:experiment-details}, it is worthywhile to remark that, 

% \yuting{in our final version, we can also include a box plot to show coverage on the true parameter}

\section{Discussion}

We develop high-probability convergence guarantees and Berry-Esseen bounds for TD learning with linear function approximation and $i.i.d.$ samples. We also propose an efficient way to construct a covariance estimator and derive its consistency rates. These results lead to a practicable methodology for constructing convex confidence sets for the target parameter, with its size vanishing at a parametric rate $O(T^{-1/2})$. An important direction for future research is to investigate whether the order $O(T^{-1/3})$ for the convergence speed of coverage rate in Theorem \ref{cor:confidence-region} is tight. Another interesting extension would be to derive specialized Berry-Esseen bounds for the sub-class of hyper-rectangles, where one would expect to find a better dependence on the dimension $d$ of the feature space, possibly of poly-logarithmic, as opposed to polynomial, order. 

Our analysis and results can be generalized to cover other iterative algorithms, such as stochastic gradient descent and stochastic approximation methods. In Appendix \ref{sec:sa.extension}, we describe conditions under which our bounds and rates apply to the output of a general linear stochastic approximation algorithm for i.i.d. data, a setting that has received attention in the recent literature \citep[see, e.g.,][]{sheshukova2025gaussianapproximationmultiplierbootstrap,samsonov2024gaussian}.
%that has received settings that have been recently considered by  appropriate assumptions (detailed in ) our rates Recent analyses \citep[see, e.g.,][]{sheshukova2025gaussianapproximationmultiplierbootstrap,samsonov2024gaussian} have demonstrated Berry-Esseen bounds for the convex distance (with an estimated variance based on i.i.d. data), as well as valid bootstrap schemes. 

%%%%%%%%%%%%%%%%%%%%%%%%%%%%%%%%%%%%%%%%%%%%%%%%%%%%%%%%%%%%%%%%
%% NOTE: THIS MARKS THE END OF THE "MAIN TEXT"
%%%%%%%%%%%%%%%%%%%%%%%%%%%%%%%%%%%%%%%%%%%%%%%%%%%%%%%%%%%%%%%%

%%%%%%%%%%%%%%%%%%%%%%%%%%%%%%%%%%%%%%%%%%%%%%%%%%%%%%%%%%%%%%%%
%% Bibliography
%%%%%%%%%%%%%%%%%%%%%%%%%%%%%%%%%%%%%%%%%%%%%%%%%%%%%%%%%%%%%%%%
\bibliography{main}
\bibliographystyle{rlj}

%%%%%%%%%%%%%%%%%%%%%%%%%%%%%%%%%%%%%%%%%%%%%%%%%%%%%%%%%%%%%%%%
% AUTHOR: If your paper has no supplementary materials, you may 
%         comment out the line below, which creates the title for
%         the supplementary materials.
%%%%%%%%%%%%%%%%%%%%%%%%%%%%%%%%%%%%%%%%%%%%%%%%%%%%%%%%%%%%%%%%
\beginSupplementaryMaterials

\section{Preliminaries}
In this section, we summarize some preliminary facts that would be useful in the proof of the theoretical results in this paper.

\subsection{Useful algebraic facts}

\paragraph{Supplementary notation.} For any complex number $z \in \mathbb{C}$, we use $\mathsf{Re}(z)$ to denote its real part. If a matrix $\bm{M}$ is positive (semi-)definite, i.e. $\lambda_{\min}(\bm{M}) > (\geq) 0$, we write $\bm{M} \succ (\succeq) \bm{0}$; when two matrices $\bm{X}$ and $\bm{Y}$ satisfy $\bm{X} - \bm{Y} \succ (\succeq) \bm{0}$, we write $\bm{X} \succ(\succeq) \bm{Y}$ or equivalently $\bm{Y} \prec (\preceq) \bm{X}$. 
For a vector $\bm{x} \in \mathbb{R}^d$, we use $\{x_i\}_{1 \leq i \leq d}$ to denote its entries; for a matrix $\bm{X} \in \mathbb{R}^{m \times n}$, we use $\{X_{ij}\}_{1 \leq i \leq m,1\leq j \leq n}$ to denote its entries. The \emph{vectorization} of matrix $\bm{X} \in \mathbb{R}^{m \times n}$, denoted as $\textbf{vec}(\bm{X})$, is defined as a vector in $\mathbb{R}^{mn}$ with entries $[\textbf{vec}(\bm{X})]_{(i-1)m+j} = \bm{X}_{ij}$ for every $1 \leq i \leq m$ and $1 \leq j \leq n$. 
% For any matrices $\bm{X} \in \mathbb{R}^{m \times n}$, and $\bm{Y} \in \mathbb{R}^{p \times q}$, their \emph{Kronecker product}, denoted as $\bm{X} \otimes \bm{Y}$, is a matrix of the size of $mp \times nq$, and can be written in blocked form as
% \begin{align*}
% \bm{X} \otimes \bm{Y} = \begin{bmatrix}
% X_{11}\bm{Y} & X_{12}\bm{Y} & \ldots & X _{1n}\bm{Y} \\ 
% X_{21}\bm{Y} & X_{22}\bm{Y} & \ldots & X_{2n} \bm{Y} \\ 
% \ldots & \ldots & \ldots & \ldots \\ 
% X_{m1}\bm{Y} & X_{m2}\bm{Y} &\ldots &X_{mn} \bm{Y}
% \end{bmatrix}.
% \end{align*}
% With subscripts, it can be represented as $(\bm{X} \otimes \bm{Y})_{(i-1)m + j, (k-1)p + \ell} = X_{ik}Y_{j\ell}$.

For a matrix sequence $\{\bm{F}_t\}_{t \in \mathbb{N}}$, a number sequence $\{f_t\}_{t \in \mathbb{N}}$ and a fixed matrix $\bm{X}$, if there exists a constant $\widetilde{C}$ independent of $t$ and $\bm{X}$ such that
\begin{align}\label{eq:defn-O-matrix}
\|\bm{F}_t\| \leq \widetilde{C}f_t\|\bm{X}\|, \quad \text{and} \quad \mathsf{Tr}(\bm{F}_t) \leq \widetilde{C}f_t\mathsf{Tr}(\bm{X})
\end{align} 
hold simultaneously, we denote for simplicity $\bm{F}_t = O(f_t)\bm{X}$.
% \subsubsection{General facts in linear algebra}

The following lemma captures some basic properties in linear algebra regarding the relationship between Frobenius and operator norms of matrices.
\begin{lemma}\label{lemma:Frobenius}
The following properties hold true:
\begin{enumerate}
\item For any sequence of matrices $\bm{X}_1,\bm{X}_2,\ldots, \bm{X}_n \in \mathbb{R}^{d \times d}$:
\begin{align}\label{eq:Frobenius-sequence}
\left\|\prod_{i=1}^n \bm{X}_i \right\|_{\mathsf{F}} \leq \|\bm{X}_j\|_{\mathsf{F}} \prod_{\substack{1 \leq i \leq n \\ i \neq j}}\|\bm{X}_i\|.
\end{align}
\item For any matrix $\bm{X} \in \mathbb{R}^{d \times d}$,
\begin{align}\label{eq:Frobenius-operator}
\|\bm{X}\|_{\mathsf{F}} \leq \sqrt{d} \|\bm{X}\|.
\end{align}
\end{enumerate}
\end{lemma}

\begin{lemma}\label{lemma:trace}
For any matrix $\bm{X} \in \mathbb{R}^{d \times d}$ and a symmetric matrix $\bm{Y} \in \mathbb{S}^{d \times d}$, if $\bm{Y} \succeq \bm{0}$, then
\begin{align*}
\mathsf{Tr}(\bm{XY}) = \mathsf{Tr}(\bm{YX}) \leq \|\bm{X}\|\mathsf{Tr}(\bm{Y}).
\end{align*}
\end{lemma}

Lemma \ref{lemma:trace} directly imply that for any matrix sequence $\{\bm{F}_t\}_{t \in \mathbb{N}}$ on $\mathbb{R}^{d \times d}$ and a fixed matrix $\bm{X} \succeq \bm{0}$, it can be guaranteed that
\begin{align}\label{eq:lemma-O-matrix}
\bm{F}_t \bm{X} = O(\|\bm{F}_t\|)\bm{X}, \quad \text{and} \quad  \bm{X}\bm{F}_t = O(\|\bm{F}_t\|)\bm{X}.
\end{align}

The following lemma, due to \citet{chen2020statistical}[Lemma 4.1], bounds the difference between the inverse matrices and would be useful for our analysis of the plug-in variance estimator.
\begin{lemma}\label{lemma:delta-inv}
Let $\bm{X},\bm{Y}$ be two invertible matrices that satisfy $\|\bm{X}^{-1}(\bm{Y} - \bm{X})\| \leq \frac{1}{2}$. Then the difference between $\bm{Y}^{-1}$ and $\bm{X}^{-1}$ can be bounded by
\begin{align*}
\left\|\bm{Y}^{-1} - \bm{X}^{-1}\right\| \leq 2\|\bm{Y} - \bm{X}\| \|\bm{X}^{-2}\|.
\end{align*}
\end{lemma}

% \subsubsection{Properties of the matrices involved in the paper}
% This part reveals some useful properties of the matrices that are involved in the paper. 

The following lemma reveals several critical algebraic features of the matrix $\bm{A}$.
\begin{lemma}\label{lemma:A}
Let $\bm{A}_t,\bm{A}$ be defined as in \eqref{eq:defn-At} and \eqref{eq:defn-At-mean} respectively, and $\bm{\Sigma}$, $\lambda_0$ and $\lambda_{\Sigma}$ be defined as in \eqref{eq:defn-Sigma}.% and \eqref{eq:defn-lambda0}. 
Then the following features hold:
\begin{align}
&2(1-\gamma) \bm{\Sigma} \preceq \bm{A} + \bm{A}^\top \preceq 2(1+\gamma) \bm{\Sigma}; \label{eq:lemma-A-1} \\ 
&\min_{1 \leq i \leq d} \mathsf{Re}(\lambda_i(\bm{A})) \geq (1-\gamma) \lambda_0; \label{eq:lemma-A-2}\\ 
&\mathbb{E}[\bm{A}_t^\top \bm{A}_t] \preceq \bm{A} + \bm{A}^\top; \label{eq:lemma-A-3}\\ 
&\bm{A}^\top\bm{A} \preceq \lambda_{\Sigma}(\bm{A} + \bm{A}^\top); \label{eq:lemma-A-4}\\
&\left\|\bm{I}-\eta \bm{A}\right\| \leq 1-\frac{1-\gamma}{2}\lambda_0 \eta, \quad  \forall \eta \in \left(0,\frac{1}{2\lambda_{\Sigma}}\right); \label{eq:lemma-A-5}\\ 
&\|\bm{A}^{-1}\| \leq \frac{1}{\lambda_0(1-\gamma)}. \label{eq:lemma-A-6}
\end{align}
\end{lemma}
\noindent\emph{Proof}: We address these properties in order.
\paragraph{Proof of Equation \eqref{eq:lemma-A-1}.} By definition, $\bm{A} + \bm{A}^\top$ is featured by
\begin{align*}
\bm{A} + \bm{A}^\top &= \mathbb{E}_{s \sim \mu, s' \sim P(\cdot \mid s)} [\bm{\phi}(s)(\bm{\phi}(s)-\gamma \bm{\phi}(s'))^\top + (\bm{\phi}(s)-\gamma \bm{\phi}(s'))\bm{\phi}(s)^\top] \\ 
&= 2 \mathbb{E}_{s \sim \mu} [\bm{\phi}(s)\bm{\phi}(s)^\top] - \gamma \mathbb{E}_{s \sim \mu, s' \sim P(\cdot \mid s)}[\bm{\phi}(s)\bm{\phi}(s')^\top + \bm{\phi}(s')\bm{\phi}(s)^\top].
\end{align*}
By Cauchy-Schwartz inequality, the second term is bounded by
\begin{align*}
-\left(\bm{\phi}(s)\bm{\phi}(s)^\top + \bm{\phi}(s')\bm{\phi}(s')^\top\right) \preceq \bm{\phi}(s)\bm{\phi}(s')^\top + \bm{\phi}(s')\bm{\phi}(s)^\top \preceq \bm{\phi}(s)\bm{\phi}(s)^\top + \bm{\phi}(s')\bm{\phi}(s')^\top;
\end{align*}
Hence, $\bm{A} + \bm{A}^\top$ is bounded above by
\begin{align*}
\bm{A} + \bm{A}^\top &\preceq 2 \mathbb{E}_{s \sim \mu} [\bm{\phi}(s)\bm{\phi}(s)^\top]+\gamma \mathbb{E}_{s \sim \mu, s' \sim P(\cdot \mid s)} \left[\bm{\phi}(s)\bm{\phi}(s)^\top + \bm{\phi}(s')\bm{\phi}(s')^\top\right] \\ 
&= 2 \mathbb{E}_{s \sim \mu} [\bm{\phi}(s)\bm{\phi}(s)^\top]+\gamma \left(\mathbb{E}_{s \sim \mu} [\bm{\phi}(s)\bm{\phi}(s)^\top] + \mathbb{E}_{s' \sim \mu} [\bm{\phi}(s)\bm{\phi}(s)^\top]\right) \\ 
&= (2+2\gamma) \mathbb{E}_{s \sim \mu} [\bm{\phi}(s)\bm{\phi}(s)^\top] = 2(1+\gamma)\bm{\Sigma},
\end{align*}
and similarly below by 
\begin{align*}
\bm{A} + \bm{A}^\top \succeq 2(1-\gamma) \bm{\Sigma}.
\end{align*}
Notice we applied the fact that when $s \sim \mu$ and $s' \sim P(\cdot \mid s)$, the marginal distribution of $s$ is also $\mu$.
\paragraph{Proof of Equation \eqref{eq:lemma-A-2}.} This result is a direct corollary of \eqref{eq:lemma-A-1}. Specifically, let $z = \lambda+ i\mu  \in \mathbb{C}$ denote any eigenvalue of $\bm{A}$, where $\lambda, \mu \in \mathbb{R}$; and $\bm{x} + i\bm{y}$ denote the corresponding eigenvector in $\mathbb{C}^d$, where $\bm{x},\bm{y} \in \mathbb{R}^d$ satisfy $\|\bm{x}\|_2 = \|\bm{y}\|_2=1$. It then follows that
\begin{align*}
\bm{A}(\bm{x}+i\bm{y}) = (\lambda + i\mu) (\bm{x}+i\bm{y}) = (\lambda \bm{x} - \mu \bm{y}) +i(\mu \bm{x} + \lambda \bm{y}).
\end{align*}
Comparing the real and imaginary parts of the left-most and right-most terms, we obtain
\begin{align*}
\bm{A}\bm{x} = \lambda \bm{x} - \mu \bm{y}, \quad \text{and} \quad \bm{A}\bm{y} = \mu \bm{x} + \lambda \bm{y}.
\end{align*}
Direct calculation yields
\begin{align*}
&\bm{x}^\top \bm{A} \bm{x} = \lambda \|\bm{x}\|_2^2 - \mu \bm{x}^\top \bm{y} = \lambda - \mu \bm{x}^\top \bm{y}, \quad \text{and} \\ 
& \bm{y}^\top\bm{A}\bm{y} = \mu \bm{y}^\top \bm{x} + \lambda \|\bm{y}\|_2^2 = \mu \bm{y}^\top \bm{x} + \lambda.
\end{align*}
Hence, Equation \eqref{eq:lemma-A-1} implies 
\begin{align*}
\lambda &= \frac{1}{2}(\bm{x}^\top \bm{A} \bm{x} + \bm{y}^\top\bm{A}\bm{y}) \\ 
&= \frac{1}{4}\bm{x}^\top (\bm{A} + \bm{A}^\top) \bm{x} + \frac{1}{4} \bm{y}^\top (\bm{A} + \bm{A}^\top) \bm{y} \\ 
&\geq \frac{1}{4} \bm{x}^\top (1-\gamma) \bm{\Sigma} \bm{x} + \frac{1}{4} \bm{y}^\top (1-\gamma) \bm{\Sigma} \bm{y} \\ 
&\geq \frac{1}{4}(2(1-\gamma)\lambda_0 + 2(1-\gamma)\lambda_0) = (1-\gamma)\lambda_0.
\end{align*}
This completes the proof of \eqref{eq:lemma-A-2}.
\paragraph{Proof of Equation \eqref{eq:lemma-A-3}.} The left-hand-side of \eqref{eq:lemma-A-3} can be bounded by
\begin{align*}
\mathbb{E}[\bm{A}_t^\top \bm{A}_t] &= \mathbb{E}[ (\bm{\phi}(s_t) - \gamma \bm{\phi}(s_t'))\bm{\phi}(s_t)^\top\bm{\phi}(s_t) (\bm{\phi}(s_t) - \gamma \bm{\phi}(s_t'))^\top ]\\
&= \mathbb{E}\|\bm{\phi}(s_t)\|_2^2 [(\bm{\phi}(s_t) - \gamma \bm{\phi}(s_t'))(\bm{\phi}(s_t) - \gamma \bm{\phi}(s_t'))^\top]\\
&\preceq \max_{s \in \mathcal{S}}\|\bm{\phi}(s)\|_2^2 \mathbb{E}[(\bm{\phi}(s_t) - \gamma \bm{\phi}(s_t'))(\bm{\phi}(s_t) - \gamma \bm{\phi}(s_t'))^\top]\\
&\preceq \mathbb{E}[\bm{\phi}(s_t)\bm{\phi}(s_t)^\top] -\gamma \mathbb{E}[\bm{\phi}(s_t)\bm{\phi}(s_t')^\top + \bm{\phi}(s_t')\bm{\phi}(s_t)^\top] + \gamma^2 \mathbb{E}[\bm{\phi}(s_t')\bm{\phi}(s_t')^\top]\\
&\preceq 2\mathbb{E}[\bm{\phi}(s_t)\bm{\phi}(s_t)^\top]  - \gamma \mathbb{E}[\bm{\phi}(s_t)\bm{\phi}(s_t')^\top + \bm{\phi}(s_t')\bm{\phi}(s_t)^\top]\\
&= \bm{A}+\bm{A}^\top,
\end{align*}
Notice that the inequality on the fourth line follows from the assumption that $\max_{s \in \mathcal{S}}\|\bm{\phi}(s)\|_2 \leq 1$ and the inequality on the fifth line follows from the fact that since $s_t$ and $s_t'$ and both drawn from the stationary distribution, $\mathbb{E}[\bm{\phi}(s_t')\bm{\phi}(s_t')^\top] = \mathbb{E}[\bm{\phi}(s_t)\bm{\phi}(s_t)^\top]$.
\paragraph{Proof of Equation \eqref{eq:lemma-A-4}.} It suffices to show that for any $\bm{x} \in \mathbb{R}^d$, 
\begin{align*}
\bm{x}^\top \bm{A}^\top \bm{Ax} \leq \lambda_{\Sigma} \bm{x}^\top (\bm{A} + \bm{A}^\top) \bm{x}.
\end{align*}
By definition, the left hand side of this inequality is featured by
\begin{align*}
\bm{x}^\top \bm{A}^\top \bm{Ax} = \|\bm{Ax}\|_2^2 = \left\|\mathbb{E}[\bm{\phi}(s)(\bm{\phi}(s)-\gamma \bm{\phi}(s'))^\top \bm{x}]\right\|_2^2,
\end{align*}
where the expectation is taken over the distribution $s \sim \mu, s' \sim P(\cdot \mid s)$. For simplicity, denote
\begin{align*}
f(s) = \mathbb{E}_{s' \sim P(\cdot \mid s)}[(\bm{\phi}(s)-\gamma \bm{\phi}(s'))^\top \bm{x}];
\end{align*}
then the law of total expectations yields
\begin{align*}
\bm{x}^\top \bm{A}^\top \bm{Ax} = \left\|\mathbb{E}[f(s)\bm{\phi}(s)]\right\|_2^2.
\end{align*}
In what follows, we aim to show that
\begin{align}\label{eq:lemma-A4-Cauchy}
\left\|\mathbb{E}[f(s)\bm{\phi}(s)]\right\|_2^2\leq \lambda_{\Sigma}\mathbb{E}[f^2(s)].
\end{align}
Observe that for any vector $\bm{y} \in \mathbb{R}^d$, it can always be guaranteed that
\begin{align*}
\mathbb{E}[f(s) - \bm{\phi}^\top(s)\bm{y}]^2 \geq 0.
\end{align*}
A direct expansion yields
\begin{align*}
0 &\leq \mathbb{E}[f^2(s)] - 2\mathbb{E}[f(s)\bm{\phi}^\top(s)\bm{y}] + \bm{y}^\top \mathbb{E}[\bm{\phi}(s)\bm{\phi}(s)^\top] \bm{y} \\ 
&= \mathbb{E}[f^2(s)] - 2\mathbb{E}[f(s)\bm{\phi}^\top(s)\bm{y}] + \bm{y}^\top \bm{\Sigma} \bm{y}\\
&\leq \mathbb{E}[f^2(s)] - 2\mathbb{E}[f(s)\bm{\phi}^\top(s)\bm{y}] +\lambda_{\Sigma}\|\bm{y}\|_2^2 , \quad \forall \bm{y} \in \mathbb{R}^d.
\end{align*}
By taking
\begin{align*}
y = \frac{\mathbb{E}[f(s)\bm{\phi}(s)]}{\lambda_{\Sigma}},
\end{align*}
we obtain
\begin{align*}
0 \leq \mathbb{E}[f^2(s)]-\frac{\left\|\mathbb{E}[f(s)\bm{\phi}(s)]\right\|_2^2}{\lambda_{\Sigma}},
\end{align*}
and \eqref{eq:lemma-A4-Cauchy} follows immediately. In other words, it can be guaranteed for every $\bm{x} \in \mathbb{R}^d$ that
\begin{align*}
\bm{x}^\top \bm{A}^\top \bm{Ax} \leq \lambda_{\Sigma}\mathbb{E}[f^2(s)].
\end{align*}
and we now proceed to bounding $\mathbb{E}[f^2(s)]$. The Jensen's inequality implies
\begin{align*}
\mathbb{E}[f^2(s)] &= \mathbb{E}_{s \sim \mu} \left\{\bm{x}^\top \mathbb{E}_{s' \sim P(\cdot \mid s)}[\bm{\phi}(s)-\gamma\bm{\phi}(s')]\mathbb{E}_{s' \sim P(\cdot \mid s)}[\bm{\phi}(s)-\gamma\bm{\phi}(s')]^\top \bm{x}\right\}\\
&\leq  \mathbb{E}_{s \sim \mu} \left\{\bm{x}^\top \mathbb{E}_{s' \sim P(\cdot \mid s)}[\bm{\phi}(s)-\gamma\bm{\phi}(s')][\bm{\phi}(s)-\gamma\bm{\phi}(s')]^\top \bm{x}\right\} \\ 
&= \bm{x}^\top \mathbb{E}\left\{[\bm{\phi}(s)-\gamma\bm{\phi}(s')][\bm{\phi}(s)-\gamma\bm{\phi}(s')]^\top\right\}\bm{x} \\ 
&\leq \bm{x}^\top(\bm{A} + \bm{A}^\top)\bm{x},
\end{align*}
where the last inequality follows from the same logic as in the proof of Equation \eqref{eq:lemma-A-3}. Equation \eqref{eq:lemma-A-4} follows immediately. 
\paragraph{Proof of Equation \eqref{eq:lemma-A-5}.} We firstly notice that 
\begin{align*}
(\bm{I} - \eta \bm{A})^\top (\bm{I}-\eta \bm{A})= \bm{I} - \eta(\bm{A} + \bm{A}^\top) + \eta^2 \bm{A}^\top \bm{A}.
\end{align*}
By invoking \eqref{eq:lemma-A-4}, we obtain
\begin{align*}
(\bm{I} - \eta \bm{A})^\top (\bm{I}-\eta \bm{A}) \preceq \bm{I} - (\eta-\lambda_{\Sigma}\eta^2)(\bm{A} + \bm{A}^\top).
\end{align*}
For any $\eta \in (0,\frac{1}{2\lambda_{\Sigma}})$, this upper bound is controlled by
\begin{align*}
(\bm{I} - \eta \bm{A})^\top (\bm{I}-\eta \bm{A}) \preceq  \bm{I} - \frac{\eta}{2}(\bm{A} + \bm{A}^\top) \preceq (1 - (1-\gamma) \lambda_0 \eta) \bm{I}.
\end{align*}
Hence, the spectral norm of $\bm{I} -\eta \bm{A}$ is controlled by
\begin{align*}
\|\bm{I}-\eta \bm{A}\| \leq \sqrt{1-(1-\gamma)\lambda_0 \eta } \leq 1-\frac{1-\gamma}{2}\lambda_0 \eta.
\end{align*}
\paragraph{Proof of Equation \eqref{eq:lemma-A-6}.} For any unit vector $\bm{x} \in \mathbb{R}^d$, we observe
\begin{align*}
\bm{x}^\top \bm{\Sigma}^{-\frac{1}{2}}\bm{A}\bm{\Sigma}^{-\frac{1}{2}}\bm{x} &= \bm{x}^\top \bm{\Sigma}^{-\frac{1}{2}}\mathbb{E}[\bm{\phi}(s)\bm{\phi}^\top(s)]\bm{\Sigma}^{-\frac{1}{2}}\bm{x} - \gamma \bm{x}^\top \bm{\Sigma}^{-\frac{1}{2}}\mathbb{E}[\bm{\phi}(s)\bm{\phi}^\top(s')]\bm{\Sigma}^{-\frac{1}{2}}\bm{x} \\ 
&= \bm{x}^\top \bm{\Sigma}^{-\frac{1}{2}}\bm{\Sigma} \bm{\Sigma}^{-\frac{1}{2}}\bm{x}- \gamma \bm{x}^\top \bm{\Sigma}^{-\frac{1}{2}}\mathbb{E}[\bm{\phi}(s)\bm{\phi}^\top(s')]\bm{\Sigma}^{-\frac{1}{2}}\bm{x} \\ 
&= 1 - \gamma \mathbb{E}[(\bm{x}^\top \bm{\Sigma}^{-\frac{1}{2}}\bm{\phi}(s)) (\bm{x}^\top \bm{\Sigma}^{-\frac{1}{2}}\bm{\phi}(s'))] \\ 
&\leq 1 - \gamma \sqrt{\mathbb{E}[(\bm{x}^\top \bm{\Sigma}^{-\frac{1}{2}}\bm{\phi}(s))^2]\mathbb{E}[(\bm{x}^\top \bm{\Sigma}^{-\frac{1}{2}}\bm{\phi}(s'))^2]} \\ 
&= 1 - \gamma \sqrt{(\bm{x}^\top \bm{\Sigma}^{-\frac{1}{2}}\mathbb{E}[\bm{\phi}(s)\bm{\phi}^\top(s)]\bm{\Sigma}^{-\frac{1}{2}}\bm{x})\cdot (\bm{x}^\top \bm{\Sigma}^{-\frac{1}{2}}\mathbb{E}[\bm{\phi}(s')\bm{\phi}^\top(s')]\bm{\Sigma}^{-\frac{1}{2}}\bm{x})}\\ 
&= 1 - \gamma \bm{x}^\top \bm{\Sigma}^{-\frac{1}{2}}\bm{\Sigma} \bm{\Sigma}^{-\frac{1}{2}}\bm{x} = 1-\gamma,
\end{align*}
where the fourth line is an application of the Cauchy-Schwartz inequality. Consequently, for any unit vector $\bm{x} \in \mathbb{R}^d$, it can be guaranteed that
\begin{align*}
\|\bm{\Sigma}^{-\frac{1}{2}}\bm{A}\bm{\Sigma}^{-\frac{1}{2}}\bm{x}\|_2 &= \|\bm{x}\|_2 \|\bm{\Sigma}^{-\frac{1}{2}}\bm{A}\bm{\Sigma}^{-\frac{1}{2}}\bm{x}\|_2 \\ 
&\geq \bm{x}^\top \bm{\Sigma}^{-\frac{1}{2}}\bm{A}\bm{\Sigma}^{-\frac{1}{2}}\bm{x} \geq 1-\gamma,
\end{align*}
where the inequality, again, is an application of the Cauchy-Schwartz inequality. This essentially means that $\|\bm{\Sigma}^{\frac{1}{2}}\bm{A}^{-1}\bm{\Sigma}^{\frac{1}{2}}\| \leq \frac{1}{1-\gamma}$, and therefore
\begin{align*}
\|\bm{A}^{-1}\| \leq \frac{1}{1-\gamma}\frac{1}{\lambda_{\min}(\bm{\Sigma})} = \frac{1}{\lambda_0(1-\gamma)}.
\end{align*}
\qed

Lemma \ref{lemma:A} implies the following property regarding a Lyapunov equation.

\begin{lemma}\label{lemma:Lyapunov}
For any matrix $\bm{E} \succeq \bm{0}$, there exists a unique positive definite matrix $\bm{X} \succeq \bm{0}$, such that
\begin{align*}
\bm{AX} + \bm{XA}^\top  = \bm{E}.
\end{align*}
Furthermore, it can be guaranteed that
\begin{align*}
\|\bm{X}\| \leq \frac{1}{2(1-\gamma)\lambda_0} \|\bm{E}\|, \quad \text{and} \quad \mathsf{Tr}(\bm{X}) \leq \frac{1}{2(1-\gamma)\lambda_0} \mathsf{Tr}(\bm{E}).
\end{align*}
\end{lemma}
\emph{Proof:} The existence of $\bm{X}$ follows from the basic property of Lyapunov equation and \eqref{eq:lemma-A-2}; we focus on bounding the norm and trace of $\bm{X}$.
\paragraph{Bounding the operator norm of $\bm{X}$.} Let $\lambda = \lambda_{\max}(\bm{X})$ and $\bm{x}$ denote a corresponding eigen vector with $\|\bm{x}\|_2 = 1$. By definition, $\bm{Xx} = \lambda \bm{x}$, and therefore
\begin{align*}
\|\bm{E}\| \geq \bm{x}^\top \bm{Ex} =\bm{x}^\top (\bm{AX} + \bm{XA}^\top) \bm{x} = \bm{x}^\top (\bm{A}+\bm{A}^\top) \bm{x} \geq [2(1-\gamma) \lambda_0] \cdot \lambda,
\end{align*}
where the last inequality follows from \eqref{eq:lemma-A-1}. The bound on $\|\bm{X}\|$ follows naturally.
\paragraph{Bounding the norm of $\bm{X}$}. Since $\bm{X} \succeq \bm{0}$, direct calculation yields
\begin{align*}
\mathsf{Tr}(\bm{E}) &= \mathsf{Tr}(\bm{AX}+\bm{XA}^\top) \\ 
&= \mathsf{Tr}(\bm{X}^{\frac{1}{2}}\bm{AX}^{\frac{1}{2}} + \bm{X}^{\frac{1}{2}}\bm{A}^\top \bm{X}^{\frac{1}{2}}) \\ 
&= \mathsf{Tr}(\bm{X}^{\frac{1}{2}}(\bm{A} + \bm{A}^\top) \bm{X}^{\frac{1}{2}}) \\ 
&\geq \mathsf{Tr}(\bm{X}^{\frac{1}{2}} \cdot 2(1-\gamma)\lambda_0 \cdot \bm{X}^{\frac{1}{2}}) = 2(1-\gamma)\lambda_0\mathsf{Tr}(\bm{X}),
\end{align*}
where we invoked the cyclability of trace and \eqref{eq:lemma-A-1} through the process. 
\qed

The following lemma further features the compressive power of the matrix $\prod_{k=1}^t (\bm{I} - \eta_k \bm{A})$.
\begin{lemma}\label{lemma:R}
Assume that $\beta \in (0,1)$ and $\alpha \in (\frac{1}{2},1)$. Let $\eta_t = \eta_0 t^{-\alpha}$ for all positive integer $t$, then the following inequalities hold:
\begin{align}
&t^{\alpha} \prod_{k=1}^t \left(1-\beta k^{-\alpha} \right) \leq e \left(\frac{\alpha}{\beta}\right)^{\frac{\alpha}{1-\alpha}} ;\label{eq:lemma-R-1} \\ 
&\sum_{i=1}^t i^{-\nu} \prod_{k=i+1}^t \left(1-\beta k^{-\alpha} \right) \leq \frac{1}{\nu-1} \left(\frac{2(\nu-\alpha)}{\beta}\right)^{\frac{\nu-\alpha}{1-\alpha}} t^{\alpha-\nu}, \quad \forall \nu \in (1,\alpha+1]; \label{eq:lemma-R-2} \\ 
&\sum_{i=1}^t i^{-2\alpha} \prod_{k=i+1}^t \left(1-\beta k^{-\alpha} \right)  = \frac{1}{\beta}t^{-\alpha} + 5\left(\frac{2}{\beta}\right)^{\frac{2-\alpha}{1-\alpha}} t^{-1}; \label{eq:lemma-R-3} \\ 
&\max_{1 \leq i \leq t} i^{-\alpha} \prod_{k=i+1}^t \left(1-\beta k^{-\alpha} \right) \leq e \left(\frac{\alpha}{\beta}\right)^{\frac{\alpha}{1-\alpha}} t^{-\alpha}. \label{eq:lemma-R-4}
\end{align} 
\end{lemma}

\noindent\emph{Proof:} We prove these four inequalities in order.
\paragraph{Proof of Equation \eqref{eq:lemma-R-1}.} We firstly observe that
\begin{align*}
t^{\alpha} \prod_{k=1}^t \left(1-\beta k^{-\alpha} \right) < t^{\alpha} \exp \left(-\beta \sum_{k=1}^t k^{-\alpha}\right),
\end{align*}
in which
\begin{align*}
\sum_{k=1}^t k^{-\alpha} > \int_1^t x^{-\alpha} \mathrm{d}x = \frac{t^{1-\alpha}-1}{1-\alpha}.
\end{align*}
Hence, the left hand side of \eqref{eq:lemma-R-1} is upper bounded by
\begin{align*}
t^{\alpha} \prod_{k=1}^t \left(1-\beta k^{-\alpha} \right) &< t^{\alpha} \exp \left(-\beta \cdot \frac{t^{1-\alpha}-1}{1-\alpha}\right) \\ 
&= e^{\frac{\beta}{1-\alpha}} \cdot t^{\alpha} \exp\left(-\frac{\beta}{1-\alpha} t^{1-\alpha}\right).
\end{align*}
By letting $u = t^{1-\alpha}$, we obtain
\begin{align*}
t^{\alpha} \prod_{k=1}^t \left(1-\beta k^{-\alpha} \right) &< e^{\frac{\beta}{1-\alpha}} \cdot u^{\frac{\alpha}{1-\alpha}} \exp \left(-\frac{\beta}{1-\alpha} u\right),
\end{align*}
where the right hand side is a function of $u$. It is easy to verify that the maximum of this function is reached when $u = \alpha / \beta$, and hence
\begin{align*}
t^{\alpha} \prod_{k=1}^t \left(1-\beta k^{-\alpha} \right) &< e^{\frac{\beta}{1-\alpha}} \cdot \left(\frac{\alpha}{\beta}\right)^{\frac{\alpha}{1-\alpha}} \exp\left(-\frac{\alpha}{1-\alpha}\right).
\end{align*}
The upper bound \eqref{eq:lemma-R-1} follows naturally.

 \paragraph{Proof of Equation \eqref{eq:lemma-R-2}.} In order to derive an iterative relationship with respect to the left-hand-side of \eqref{eq:lemma-R-2}, we denote 
 \begin{align*}
X_t = \sum_{i=1}^{t}
i^{-\nu} \prod_{k=i+1}^t \left(1-\beta k^{-\alpha}\right)
 \end{align*}
 for any positive integer $t$. Then by definition, $X_1 = 1$ and the sequence $\{X_t\}_{t \geq 1}$ is featured by the iterative relation
 \begin{align}\label{eq:lemma-R2-iteration}
X_{t+1} = \left(1-\beta (t+1)^{-\alpha}\right)X_t + (t+1)^{-\nu}.
 \end{align} 
 For simplicity, we denote
 \begin{align*}
&C = \frac{1}{\nu-1} \left(\frac{2(\nu-\alpha)}{\beta}\right)^{\frac{\nu-\alpha}{1-\alpha}}.
 \end{align*}
 then the lemma is equivalent to the claim that
\begin{align}\label{eq:lemma-R2-induction}
t^{\nu-\alpha}X_t \leq C
\end{align}
holds for all positive integer $t$, and can hence be proved by an induction argument. On one hand, we observe that the iterative relation \eqref{eq:lemma-R2-iteration} implies
\begin{align*}
X_{t+1} < X_t + (t+1)^{-\nu},
\end{align*}
hence $X_t$ can be bounded by
\begin{align*}
X_t < \sum_{i=1}^t i^{-\nu} < \int_{1}^{\infty} x^{-\nu} \mathrm{d}x = \frac{1}{\nu-1}.
\end{align*}
Therefore, we can take 
\begin{align*}
t^\star = \left(\frac{2(\nu-\alpha)}{\beta}\right)^{\frac{1}{1-\alpha}},
\end{align*}
then for any $t \leq t^\star$, it can be guaranteed that
\begin{align*}
t^{\nu-\alpha} X_t < (t^\star)^{\nu-\alpha} \cdot \frac{1}{\nu-1}  = C.
\end{align*}

On the other hand, assume that \eqref{eq:lemma-R2-induction} holds for an integer $t = k \geq t^\star$, then for $t = k+1$, we observe 
\begin{align*}
&t^{\nu-\alpha} X_t \\ 
&= (k+1)^{\nu-\alpha} \left[(1-\beta (k+1)^{-\alpha}) X_k + (k+1)^{-\alpha}\right]\\ 
&\leq (k+1)^{\nu-\alpha} \cdot Ck^{\alpha-\nu} - \beta C k^{\alpha-\nu}(k+1)^{\nu-2\alpha} + (k+1)^{-\alpha}\\
&= C + C\left[\left(\frac{k+1}{k}\right)^{\nu-\alpha}-1\right]-\beta C (k+1)^{-\alpha}\left(\frac{k+1}{k}\right)^{\nu-\alpha} + (k+1)^{-\alpha} \\ 
&= C + C\left(\frac{k+1}{k}\right)^{\nu-\alpha}\left[1-\left(\frac{k}{k+1}\right)^{\nu-\alpha}\right]  \\ 
&- \beta C (k+1)^{-\alpha}\left(\frac{k+1}{k}\right)^{\nu-\alpha} + (k+1)^{-\alpha}\\
&\overset{(i)}{<} C + C\left(\frac{k+1}{k}\right)^{\nu-\alpha} \cdot \frac{\nu-\alpha}{k+1} -\beta C (k+1)^{-\alpha}\left(\frac{k+1}{k}\right)^{\nu-\alpha} + (k+1)^{-\alpha}\\
&\overset{(ii)}{<}C - \frac{\beta}{2}C(k+1)^{-\alpha}\left(\frac{k+1}{k}\right)^{\nu-\alpha} + (k+1)^{-\alpha} \\
& \overset{(iii)}{<} C - (k+1)^{-\alpha}\left(\frac{k+1}{k}\right)^{\nu-\alpha} + (k+1)^{-\alpha} < C.
\end{align*}
Throughout the process, we applied the following facts:
\begin{align*}
&(i): \left(\frac{k}{k+1}\right)^{\nu-\alpha} = \left(1-\frac{1}{k+1}\right)^{\nu-\alpha} \geq 1-\frac{\nu-\alpha}{k+1}, \quad \text{since }\nu-\alpha \in (1-\alpha,1];\\
&(ii): \frac{\nu-\alpha}{k+1} < \frac{\beta}{2}(k+1)^{-\alpha}, \quad \text{since } k > t^\star; \quad \text{and}\\ 
&(iii): \beta C = \frac{\beta}{\nu-1} \left(\frac{2(\nu-\alpha)}{\beta}\right)^{\frac{\nu-\alpha}{1-\alpha}} > \frac{\beta}{\nu-1}\frac{2(\nu-\alpha)}{\beta} > 2, \quad \text{since } \nu > 1 > \alpha. 
\end{align*}
An induction argument then validates \eqref{eq:lemma-R2-induction} for every positive integer $t$.

\paragraph{Proof of Equation \eqref{eq:lemma-R-3}.} In order to derive an iterative relationship with respect to the left-hand-side of \eqref{eq:lemma-R-3}, we denote 
 \begin{align*}
X_t = \sum_{i=1}^{t}
i^{-2\alpha} \prod_{k=i+1}^t \left(1-\beta k^{-\alpha}\right)
 \end{align*}
 for any positive integer $t$. Then by definition, $X_1 = 1$ and the sequence $\{X_t\}_{t \geq 1}$ is featured by the iterative relation
 \begin{align}\label{eq:lemma-R3-iteration}
X_{t} = \left(1-\beta t^{-\alpha}\right)X_{t-1} + t^{-2\alpha}, \quad \text{for every }t > 1. 
 \end{align} 
 Direct calculation yields
 \begin{align*}
t^{\alpha} X_t &= \left(\frac{t}{t-1}\right)^{\alpha} \left(1-\beta t^{-\alpha}\right) [(t-1)^{\alpha} X_{t-1}] + t^{-\alpha} \\ 
&= \left(\frac{t}{t-1}\right)^{\alpha} \left(1-\beta t^{-\alpha}\right) \cdot \frac{1}{\beta} + \left(\frac{t}{t-1}\right)^{\alpha} \left(1-\beta t^{-\alpha}\right)\left[(t-1)^{\alpha} X_{t-1}-\frac{1}{\beta}\right] + t^{-\alpha} \\ 
&= \frac{1}{\beta} + \left[\left(\frac{t}{t-1}\right)^{\alpha} -1\right]\left(\frac{1}{\beta}  + t^{-\alpha}\right) \\ 
&+ \left(\frac{t}{t-1}\right)^{\alpha} \left(1-\beta t^{-\alpha}\right)\left[(t-1)^{\alpha} X_{t-1}-\frac{1}{\beta}\right].
 \end{align*}
 Setting $Z_t = t^{\alpha} X_t - \frac{1}{\beta}$, we immediately obtain
 \begin{align*}
Z_t = \left(\frac{t}{t-1}\right)^{\alpha} \left(1-\beta t^{-\alpha}\right) Z_{t-1} + \left[\left(\frac{t}{t-1}\right)^{\alpha} -1\right]\left(\frac{1}{\beta} + t^{-\alpha}\right).
 \end{align*}
 Notice that for every $t > 1$ and $\alpha < 1$, 
 \begin{align*}
\left(\frac{t}{t-1}\right)^{\alpha} -1 < \frac{t}{t-1} - 1 = \frac{1}{t-1} < \frac{2}{t}.
 \end{align*} 
 Hence, the sequence $\{Z_t\}_{t \geq 1}$ is upper bounded iteratively by
 \begin{align*}
Z_t < \left(\frac{t}{t-1}\right)^{\alpha} \left(1-\beta t^{-\alpha}\right) Z_{t-1} + \frac{2}{t}\left(\frac{1}{\beta} + t^{-\alpha}\right).
 \end{align*}
 By induction, this directly implies
\begin{align*}
Z_t < \underset{I_1}{\underbrace{t^{\alpha} \prod_{k=1}^t \left(1-\beta k^{-\alpha}\right)Z_1}} + \underset{I_2}{\underbrace{2t^{\alpha} \sum_{i=1}^t \left(\frac{1}{\beta}i^{-\alpha-1}+i^{-2\alpha-1}\right) \prod_{k=i+1}^t (1-\beta k^{-\alpha})}},
\end{align*}
where $Z_1 = 1^{\alpha} X_1 - \frac{1}{\beta} = 1-\frac{1}{\beta}$. The proof now boils down to showing that both $I_1$ and $I_2$ converges to $0$ as $t \to \infty$. On one hand, $I_1$ is bounded by
\begin{align*}
t^{\alpha} \exp\left(-\beta \sum_{k=1}^t k^{-\alpha}\right) < \left(\frac{1}{\beta}\right)^{\frac{1}{1-\alpha}}t^{\alpha-1}.
\end{align*}
On the other hand, the convergence of $I_2$ is guaranteed by \eqref{eq:lemma-R-2}:
\begin{align*}
I_2 &\leq \frac{4t^{\alpha}}{\beta} \sum_{i=1}^t i^{-\alpha-1}\prod_{k=i+1}^t (1-\beta k^{-\alpha})  \\ 
&\leq \frac{2}{\alpha}\left(\frac{2}{\beta}\right)^{\frac{2-\alpha}{1-\alpha}} t^{\alpha-1}.
\end{align*}
 Combining these results, we obtain
\begin{align*}
t^{\alpha}X_t - \frac{1}{\beta} = Z_t < 5\left(\frac{2}{\beta}\right)^{\frac{2-\alpha}{1-\alpha}} t^{\alpha-1},
\end{align*}
which directly implies \eqref{eq:lemma-R-3}.

\paragraph{Proof of Equation \eqref{eq:lemma-R-4}.} For every positive integer $t$, we begin by identifying the index $i$ that maximizes
\begin{align*}
i^{-\alpha} \prod_{k=i+1}^t \left(1-\beta k^{-\alpha}\right):= X_{i}^t.
\end{align*}
Consider the ratio of two consecutive terms:
\begin{align*}
\frac{X_{i+1}}{X_i} &= \frac{(i+1)^{-\alpha}}{i^{-\alpha} (1-\beta (i+1)^{-\alpha})}\\
&= \frac{i^{\alpha}}{(i+1)^{\alpha} - \beta}, \quad \text{for any } 1 \leq i < t. 
\end{align*}
Equivalently, we observe
\begin{align*}
\frac{X_i}{X_{i+1}} = 1 + i^{-\alpha}\left[(i+1)^{\alpha}-i^{\alpha} - \beta \right].
\end{align*}
Since $(i+1)^{\alpha} - i^{\alpha} \searrow 0$ for any $\alpha \in (\frac{1}{2},1)$, we can conclude that when
\begin{align*}
2^{\alpha} - 1 - \beta < 0,
\end{align*}
the sequence ${X_i^t}_{1 \leq i \leq t}$ is monotonically increasing; otherwise, there exists a constant $i^\star$ such that $\{X_i^t\}_{1 \leq i \leq t}$ decreases when $i \leq i^\star$ and increases when $i > i^\star$. Either way, the maximum over $i$ is reached at either $i=1$ or $i=t$. That is,
\begin{align*}
\max_{1 \leq i \leq t} X_i^t = \max\left\{\prod_{k=2}^t\left(1-\beta k^{-\alpha}\right), t^{-\alpha} \right\}
\end{align*}
where the first term on the right hand side is bounded above by \eqref{eq:lemma-R-1}. The equation follows naturally. \qed

% In Scenario $\ref{case:nu}$, the compressive power of $(\bm{I}-\eta_k\bm{A})$ is featured by the following lemma.
% \begin{lemma}\label{lemma:R-nu}
% Consider any $\beta>1$ and $\nu>0$. Then for any integers $i,t$ satisfying $i < t$, it can be guaranteed that
% \begin{align*}
% \prod_{k=i+1}^t (1-\beta(i+\nu)^{-1}) \leq \left(\frac{i+\nu}{t+\nu}\right)^\beta.
% \end{align*}
% \end{lemma}
% \noindent\emph{Proof}: This lemma can be easily verified by an induction argument. \qed

\begin{lemma}\label{lemma:Q-uni}
For every $\beta \in (0,1)$, $\alpha \in (\frac{1}{2},1)$ and sufficiently large $T$, it can be guaranteed that
\begin{align}
&t^{-\alpha} \sum_{j=t}^T \prod_{k=t+1}^j (1-\beta k^{-\alpha}) < 3\left(\frac{2}{\beta}\right)^{\frac{1}{1-\alpha}}, \quad \text{and} \label{eq:Q-uni-1}\\ 
&t^{-\alpha} \sum_{j=t}^\infty \prod_{k=t+1}^j (1-\beta k^{-\alpha}) - \frac{1}{\beta} \lesssim  \left(\frac{1}{\beta}\right)^{\frac{1}{1-\alpha}}\Gamma\left(\frac{1}{1-\alpha}\right) t^{\alpha-1}. \label{eq:Q-uni-2}
\end{align}
\end{lemma}
\noindent\emph{Proof}: We address the two inequalities in order.
\paragraph{Proof of \eqref{eq:Q-uni-1}.}For simplicity, we denote
\begin{align*}
&t^\star = \left(\frac{2\alpha}{\beta}\right)^{\frac{1}{1-\alpha}},\\ 
&X_t= \sum_{j=t}^{T} \prod_{k=t+1}^{j} \left(1-\beta k^{-\alpha}\right), \quad \text{and} \\ 
&Y_t = t^{-\alpha} X_t.
\end{align*}
By definition, we observe that $X_T = 1$, $Y_T = T^{-\alpha}$ and that the sequences $\{X_t\}_{1 \leq t \leq T}$ and $\{Y_t\}_{1 \leq t \leq T}$ is featured by the iterative relation
\begin{align}
&X_t = 1 + \left[1-\beta(t+1)^{-\alpha}\right] X_{t+1}\label{eq:Q-iterate-Xt}\\
&Y_t = t^{-\alpha} + \left(\frac{t+1}{t}\right)^{\alpha} \left[1-\beta(t+1)^{-\alpha}\right]Y_{t+1}\label{eq:Q-iterate-Yt}.
\end{align}
In what follows, we consider the bound on $Y_t$ for both $t \geq t^\star$ and $1 \leq t < t^\star$.
\paragraph{Bounding $Y_t$ for $t \geq t^\star$.} Without loss of generality, we assume that $T > t^\star$; the bound on $Y_t$ for $t \geq t^\star$ is shown by an inverse induction from $t = T$ to $t = t^\star$. Namely, we observe that $Y_T = T^{-\alpha} < 1 < \frac{2}{\beta}$; if $Y_{t+1} < \frac{2}{\beta}$ holds for an integer $t \in (t^\star,T)$, then the iterative relation \eqref{eq:Q-iterate-Yt} implies
\begin{align*}
Y_t &< t^{-\alpha} + \left(\frac{t+1}{t}\right)^{\alpha} \left[1-\beta(t+1)^{-\alpha}\right] \cdot \frac{2}{\beta} \\ 
&< \frac{2}{\beta} + t^{-\alpha} \left\{\left[(t+1)^{\alpha} - t^{\alpha}\right] \frac{2}{\beta} - 1\right\}.
\end{align*}
Hence, it boils down to showing
\begin{align*}
(t+1)^{\alpha} - t^{\alpha} \leq \frac{\beta}{2},
\end{align*}
for which we observe
\begin{align*}
(t+1)^{\alpha} - t^{\alpha}  < \alpha t^{\alpha-1} \leq \alpha \left(t^{\star}\right)^{\alpha-1} = \frac{2}{\beta}.
\end{align*}
In this way, we prove that $Y_t \leq \frac{2}{\beta}$ for every $t \geq t^\star$.
\paragraph{Bounding $Y_t$ for $1 \leq t < t^\star$.} For every positive integer $t$ satisfying $1 \leq t < t^\star$, the iterative relation \eqref{eq:Q-iterate-Xt}
\begin{align*}
Y_t = t^{-\alpha} X_t 
    < X_t  
    < X_{t^\star} + (t^\star - t) 
    & < X_{t^\star} + t^\star  \\ 
    &= (t^\star)^{\alpha} Y_{t^\star} + t^\star \\ 
    &< \left(\frac{2\alpha}{\beta}\right)^{\frac{\alpha}{1-\alpha}} \frac{2}{\beta} + \left(\frac{2\alpha}{\beta}\right)^{\frac{1}{1-\alpha}} \\ 
    &= \left(\frac{2\alpha}{\beta}\right)^{\frac{1}{1-\alpha}}(\frac{1}{\alpha}+1).
\end{align*}
\paragraph{Proof of \eqref{eq:Q-uni-2}.} We firstly expand the left-hand-side of \eqref{eq:Q-uni-2} as
\begin{align}\label{eq:Q-uni-21}
&t^{-\alpha} \sum_{j=t}^\infty \prod_{k=t+1}^j (1-\beta k^{-\alpha}) \nonumber \\ 
&=\sum_{j=t}^\infty (j+1)^{-\alpha} \prod_{k=t+1}^j (1-\beta k^{-\alpha}) + \sum_{j=t}^\infty (t^{-\alpha}-(j+1)^{-\alpha} ) \prod_{k=t+1}^j (1-\beta k^{-\alpha}).
\end{align}
Here, the first term is bounded by applying the telescoping technique
\begin{align}\label{eq:beta-telescope}
&\sum_{j=t}^\infty (j+1)^{-\alpha} \prod_{k=t+1}^j (1-\beta k^{-\alpha}) \nonumber \\ 
&= \sum_{j=t}^\infty \frac{1}{\beta}\left[1-(1-\beta(j+1)^{-\alpha})\right] \prod_{k=t+1}^j (1-\beta k^{-\alpha})\nonumber \\
&= \frac{1}{\beta} \sum_{j=t}^\infty \left(\prod_{k=t+1}^j (1-\beta k^{-\alpha}) - \prod_{k=t+1}^{j+1}(1-\beta k^{-\alpha})\right)\nonumber\\
&= \frac{1}{\beta}  \left(\sum_{j=t}^\infty \prod_{k=t+1}^j (1-\beta k^{-\alpha})- \sum_{j=t}^\infty\prod_{k=t+1}^{j+1}(1-\beta k^{-\alpha})\right)\nonumber\\
&= \frac{1}{\beta}  \left[\left(\prod_{k=t+1}^t (1-\beta k^{-\alpha}) +\sum_{j=t+1}^\infty \prod_{k=t+1}^{j} (1-\beta k^{-\alpha})\right)- \sum_{j=t}^\infty\prod_{k=t+1}^{j+1}(1-\beta k^{-\alpha})\right]\nonumber\\
&\overset{(i)}{=} \frac{1}{\beta} \left(1+\sum_{i=t}^\infty \prod_{k=t+1}^{i+1} (1-\beta k^{-\alpha})- \sum_{j=t}^\infty\prod_{k=t+1}^{j+1}(1-\beta k^{-\alpha})\right)= \frac{1}{\beta}.
\end{align}
Notice that in (i) we invoked the convention
\begin{align*}
\prod_{k=t+1}^t (1-\beta k^{-\alpha}) = 1,
\end{align*}
and a change of variable $i = j-1$.

Therefore, the proof boils down to showing that the second term is bounded by
\begin{align}\label{eq:Q-uni-22}
\sum_{j=t}^\infty ((j+1)^{-\alpha} - t^{-\alpha}) \prod_{k=t+1}^j (1-\beta k^{-\alpha}) = O(t^{\alpha-1}). 
\end{align}
Towards this end, we denote
\begin{align}\label{eq:defn-mtj}
m_t^j = \sum_{k=t}^j k^{-\alpha}.
\end{align}
In terms of $m_t^j$, the left-hand-side of \eqref{eq:Q-uni-22} is bounded by
\begin{align*}
&\sum_{j=t}^\infty ((j+1)^{-\alpha} - t^{-\alpha}) \prod_{k=t+1}^j (1-\beta k^{-\alpha})  \\ 
&\asymp \sum_{j=t}^\infty ((j+1)^{-\alpha} - t^{-\alpha}) \exp \left( - \beta m_t^j\right).
\end{align*}
Furthermore, we observe that on one hand, $m_t^j$ is bounded by
\begin{align}\label{eq:mtj-bound}
m_t^j = \sum_{k=t}^j k^{-\alpha} > \int_t^{j+1} x^{-\alpha} \mathrm{d}x = \frac{(j+1)^{1-\alpha} - t^{1-\alpha}}{1-\alpha};
\end{align}
on the other hand, the term $(j+1)^{-\alpha} - t^{-\alpha}$ can be bounded by
\begin{align*}
 t^{-\alpha} -(j+1)^{-\alpha} &= (j+1)^{-\alpha}\left[\left(\frac{j+1}{t}\right)^{\alpha} -1\right] \\ 
 &= (m_t^{j+1} - m_t^j) \cdot \left[\left(\frac{j+1}{t}\right)^{\alpha} -1\right].
\end{align*}
In what follows, we aim to bound the last term in terms of $m_t^j$. Specifically, since
\begin{align*}
\left(\frac{j+1}{t}\right)^{1-\alpha}-1 < \frac{(1-\alpha)m_t^j}{t^{1-\alpha}},
\end{align*}
we obtain
\begin{align*}
\left(\frac{j+1}{t}\right)^{\alpha} -1 &< \left[\frac{(1-\alpha)m_t^j}{t^{1-\alpha}}+1\right]^{\frac{\alpha}{1-\alpha}}-1 \\ 
&< t^{\alpha-1} \left((1-\alpha) m_t^j +1\right)^\frac{\alpha}{1-\alpha}.
\end{align*}
Therefore, the left-hand-side of \eqref{eq:Q-uni-22} is bounded by
\begin{align*}
&\sum_{j=t}^\infty ((j+1)^{-\alpha} - t^{-\alpha}) \prod_{k=t+1}^j (1-\beta k^{-\alpha})\\ 
&< \sum_{j=t}^\infty (m_t^{j+1} - m_t^j) t^{\alpha-1} \left((1-\alpha) m_t^j +1\right)^\frac{\alpha}{1-\alpha}\exp \left(\beta t^{-\alpha} - \beta m_t^j\right)\\ 
&<t^{\alpha-1}\sum_{j=t}^\infty \left((1-\alpha) m_t^j +1\right)^\frac{\alpha}{1-\alpha} \exp \left( - \beta m_t^j\right)(m_t^{j+1} - m_t^j) \\ 
&\asymp t^{\alpha-1}\int_0^{\infty} \left((1-\alpha) m +1\right)^\frac{\alpha}{1-\alpha} \exp(-\beta m) \mathrm{d}m = O(t^{\alpha-1});
\end{align*}
the integral on the last line is further bounded by
\begin{align*}
&\int_0^{\infty} \left((1-\alpha) m +1\right)^\frac{\alpha}{1-\alpha} \exp(-\beta m) \mathrm{d}m \\
&<  \int_0^{\infty} \left(m^{\frac{\alpha}{1-\alpha}} + 2^\frac{\alpha}{1-\alpha}\right)\exp(-\beta m) \mathrm{d}m \\ 
&= 2^\frac{\alpha}{1-\alpha} \frac{1}{\beta} + \left(\frac{1}{\beta}\right)^{\frac{1}{1-\alpha}}\Gamma\left(\frac{1}{1-\alpha}\right) \lesssim \left(\frac{1}{\beta}\right)^{\frac{1}{1-\alpha}}\Gamma\left(\frac{1}{1-\alpha}\right).
\end{align*}
The bound \eqref{eq:Q-uni-2} follows naturally. \qed

Next, we introduce additional quantities 
\begin{align}
	\overline{\bm{\Lambda}}_T=\frac{1}{T}\sum_{t=1}^T \bm{Q}_t \bm{\Gamma}\bm{Q}_t^\top,  \label{eq:defn-barLambdaT}
\end{align}
where $\bm{\Gamma}$ is defined in \eqref{eq:defn-Lambdastar} and, for each $t$, we set
\begin{align}
	\bm{Q}_t = \eta_t \sum_{j=t}^{T}\prod_{k=t+1}^{j} (\bm{I}-\eta_k \bm{A})\label{eq:defn-Qt}.
\end{align}

The matrix $\overline{\bm{\Lambda}}_T$ plays the role of a non-asymptotic covariance matrix for $\overline{\bm{\theta}}_T$. The following lemmas feature the properties of $\bm{Q}_t$ and $\bar{\bm{\Lambda}}_T$.

\begin{lemma}\label{lemma:Q}
Let $\eta_t = \eta_0 t^{-\alpha}$ with $\alpha \in (\frac{1}{2},1)$ and $0 \leq \eta_0 \leq \frac{1}{2\lambda_{\Sigma}}$. Recall that $\bm{Q}_t$ is defined as in \eqref{eq:defn-Qt}. Then for any positive integer $\ell < T$, the following relation hold:
\begin{subequations}
\begin{align}
\sum_{t=1}^T (\bm{Q}_t - \bm{A}^{-1}) = -\bm{A}^{-1} \sum_{j=1}^T \prod_{k=1}^j (\bm{I}-\eta_k \bm{A}),\label{eq:Qt-Ainv-bound-1}
\end{align}
Furthermore, there exist a sequence of matrix $\{\bm{S}_t\}_{1 \leq t \leq T}$, along with a constant $\widetilde{C}$, depending on $\alpha,\eta_0$ and $\gamma$, such that the difference between $\bm{Q}_t$ and $\bm{A}^{-1}$ can be represented as
\begin{align}\label{eq:Qt-Ainv-decompose}
\bm{Q}_t - \bm{A}^{-1}  = \bm{S}_t- \bm{A}^{-1} \prod_{k=t}^T (\bm{I}-\eta_k \bm{A}),
\end{align}
in which $\bm{S}_t$ is independent of $T$, and its norm is bounded by
\begin{align}\label{eq:St1-bound}
\|\bm{S}_t\| \lesssim \eta_0 \Gamma\left(\frac{1}{1-\alpha}\right)\left(\frac{2}{(1-\gamma)\lambda_0 \eta_0}\right)^{\frac{1}{1-\alpha}}t^{\alpha-1}.
\end{align}

\end{subequations}
\end{lemma}
\noindent\emph{Proof}: We address these equations and bounds separately.
\paragraph{Proof of \eqref{eq:Qt-Ainv-bound-1}} By definition, the sum of $\bm{Q}_t$ for $1 \leq t \leq T$ can be expressed as
\begin{align*}
\sum_{t=1}^T \bm{Q}_t &= \sum_{t=1}^T \eta_t \sum_{j=t}^T \prod_{k=t+1}^j (\bm{I} - \eta_k \bm{A}) \\ 
&= \sum_{j=1}^T \sum_{t=1}^j \eta_t \prod_{k=t+1}^j (\bm{I} - \eta_k \bm{A}),
\end{align*}
where the second line follows by switching the order of summation. Here, the summand can be represented as
\begin{align}\label{eq:A-telescope}
\eta_t \prod_{k=t+1}^j (\bm{I} - \eta_k \bm{A})&= \bm{A}^{-1}[\bm{I} - (\bm{I} - \eta_t \bm{A})]\prod_{k=t+1}^j (\bm{I} - \eta_k \bm{A}) \nonumber \\ 
&= \bm{A}^{-1} \left[\prod_{k=t+1}^j (\bm{I} - \eta_k \bm{A}) - \prod_{k=t}^j (\bm{I} - \eta_k \bm{A})\right].
\end{align}
Notice that summing the difference in the bracket from $t=1$ to $t=j$ would cancel all the terms in the middle and only the first and last terms would remain. Therefore,
\begin{align*}
\sum_{t=1}^T \bm{Q}_t &= \sum_{j=1}^T \sum_{t=1}^j \eta_t \prod_{k=t+1}^j (\bm{I} - \eta_k \bm{A})\\ 
&=\sum_{j=1}^T \sum_{t=1}^j \bm{A}^{-1} \left[\prod_{k=t+1}^j (\bm{I} - \eta_k \bm{A}) - \prod_{k=t}^j (\bm{I} - \eta_k \bm{A})\right] \\ 
&= \bm{A}^{-1}\sum_{j=1}^T \left[\bm{I} - \prod_{k=1}^j (\bm{I} - \eta_k \bm{A})\right],
\end{align*}
which directly implies \eqref{eq:Qt-Ainv-bound-1}.

\paragraph{Construction of the decomposition \eqref{eq:Qt-Ainv-decompose}.} 
The definition of $\bm{Q}_t$ \eqref{eq:defn-Qt} directly implies that on one hand,
\begin{align*}
\bm{Q}_t &= \eta_t \sum_{j=t}^{T}\prod_{k=t+1}^{j} (\bm{I}-\eta_k \bm{A})= \eta_t\bm{I} + \eta_t \sum_{j=t+1}^{T} \prod_{k=t+1}^{j} (\bm{I}-\eta_k \bm{A}),
\end{align*}
and that
\begin{align*}
(\bm{I}-\eta_t \bm{A})\bm{Q}_t = \eta_t \sum_{j=t}^{T}\prod_{k=t}^{j} (\bm{I}-\eta_k\bm{A}) &= \eta_t \sum_{j=t}^{T-1}\prod_{k=t}^{j} (\bm{I}-\eta_k\bm{A}) + \eta_t \prod_{k=t}^{T}(\bm{I}-\eta_k\bm{A})\\ 
&= \eta_t \sum_{j=t+1}^{T-1}\prod_{k=t}^{j-1} (\bm{I}-\eta_k\bm{A}) + \eta_t \prod_{k=t}^{T}(\bm{I}-\eta_k\bm{A})
\end{align*}
On the other hand,
\begin{align*}
&\prod_{k=t+1}^{j} (\bm{I}-\eta_k \bm{A}) - \prod_{k=t}^{j-1} (\bm{I}-\eta_k \bm{A})\\
&= \left[\prod_{k=t+1}^{j-1} (\bm{I}-\eta_k \bm{A})\right] (\bm{I}-\eta_{j}\bm{A}) - (\bm{I}-\eta_t \bm{A}) \left[\prod_{k=t+1}^{j-1} (\bm{I}-\eta_k \bm{A})\right]\\
&= \prod_{k=t+1}^{j-1} (\bm{I}-\eta_k \bm{A}) - \eta_{j} \left[\prod_{k=t+1}^{j-1} (\bm{I}-\eta_k \bm{A})\right] \bm{A} - \prod_{k=t+1}^{j-1} (\bm{I}-\eta_k \bm{A}) + \eta_t \bm{A}\prod_{k=t+1}^{j-1} (\bm{I}-\eta_k \bm{A})\\
&= \eta_t \bm{A}\prod_{k=t+1}^{j-1} (\bm{I}-\eta_k \bm{A})-\eta_{j} \prod_{k=t+1}^{j-1} (\bm{I}-\eta_k \bm{A}) \bm{A}\\
&= (\eta_t - \eta_{j})\bm{A}\prod_{k=t+1}^{j-1} (\bm{I}-\eta_k \bm{A}).
\end{align*}
Therefore, by taking a difference between $\bm{Q}_t$ and $(\bm{I}-\eta_t \bm{A}) \bm{Q}_t$, we obtain
\begin{align*}
\eta_t \bm{A}\bm{Q}_t &= \bm{Q}_t - (\bm{I}-\eta_t \bm{A})\bm{Q}_t\\
&= \eta_t \bm{I} + \eta_t \sum_{j=t+1}^{T-1} \left[\prod_{k=t+1}^{j} (\bm{I}-\eta_k \bm{A}) - \prod_{k=t}^{j-1} (\bm{I}-\eta_k \bm{A})\right] - \eta_t \prod_{k=t}^{T}(\bm{I}-\eta_k\bm{A})\\
&= \eta_t \bm{I} + \eta_t \sum_{j=t+1}^{T-1}(\eta_t - \eta_{j})\bm{A}\prod_{k=t+1}^{j-1} (\bm{I}-\eta_k \bm{A})- \eta_t \prod_{k=t}^{T}(\bm{I}-\eta_k\bm{A}).
\end{align*}
Consequently, the difference between $\bm{Q}_t$ and $\bm{A}^{-1}$ can be expressed as
\begin{align}
\bm{Q}_t - \bm{A}^{-1} =\underset{\bm{S}_t} {\underbrace{\sum_{j=t+1}^{T-1}(\eta_t - \eta_{j})\prod_{k=t+1}^{j-1} (\bm{I}-\eta_k \bm{A})}} -  \bm{A}^{-1} \prod_{k=t}^{T}(\bm{I}-\eta_k\bm{A}),
\end{align}
which takes the form of \eqref{eq:Qt-Ainv-decompose}.
\paragraph{Proof of \eqref{eq:St1-bound}.} Lemma \ref{lemma:A} implies that
\begin{align*}
\left\| \prod_{k=t+1}^{j-1} (\bm{I}-\eta_k \bm{A}) \right\| &\leq \prod_{k=t+1}^{j-1} \left(1-\frac{1}{2}\eta_k (1-\gamma) \lambda_0\right)\\
&\leq \prod_{k=t+1}^{j}\exp\left(-\frac{1}{2}\eta_k (1-\gamma) \lambda_0\right)\\
&= \exp \left(-\frac{1}{2}(1-\gamma) \lambda_0\sum_{k=t+1}^{j-1} \eta_k\right).
\end{align*}
Recalling the definition of $m_t^j$ \eqref{eq:defn-mtj},\eqref{eq:mtj-bound}, we can bound the norm $\bm{S}_t$ by
\begin{align*}
\|\bm{S}_t\|_2 &\leq \sum_{j=t}^{\infty} (\eta_t - \eta_{j}) \left\| \prod_{k=t+1}^{j-1} (\bm{I}-\eta_k \bm{A})\right\| _2 \\
&\leq \sum_{j=t}^{\infty} (\eta_t - \eta_{j})\exp \left(-\frac{1}{2}(1-\gamma)\lambda_0\sum_{k=t+1}^{j-1} \eta_k\right)\\
&=  \sum_{j=t}^{\infty} \eta_{j} \left[\left(\frac{j}{t}\right)^{\alpha} - 1\right]\exp \left(-\frac{1}{2}(1-\gamma)\lambda_0 \eta_0 m_t^j \right).
\end{align*}
The right-hand-side can then be bounded by a similar argument to that of Lemma \ref{lemma:Q-uni}. \qed

The following Lemma features the compressive rate of the matrix $\prod_{k=t}^T (\bm{I}-\eta_k\bm{A})$.
\begin{lemma}\label{thm:St2-bound}
When $\eta_0 < \frac{1}{2\lambda_{\Sigma}}$, for every positive definite matrix $\bm{Y} \in \mathbb{R}^{d \times d}$, it can be guaranteed that 
\begin{align}\label{eq:St2-bound}
T^{-\alpha} \sum_{t=1}^T \left(\prod_{k=t}^T (\bm{I}-\eta_k\bm{A}) \bm{Y} \prod_{k=t}^T (\bm{I}-\eta_k\bm{A}^\top) \right)- \bm{X}\leq O(T^{\alpha-1})\bm{Y}
\end{align}
where $\bm{X} \succeq \bm{0}$ is the solution to the Lyapunov equation
\begin{align}\label{eq:Lyapunov-St2}
\eta_0(\bm{A} \bm{X} + \bm{XA}^\top)= \bm{Y},
\end{align}
and the big $O$ notation for matrix sequences is defined as in \eqref{eq:defn-O-matrix}.
\end{lemma}
\noindent\emph{Proof}: For simplicity, we define a sequence of matrices $\{\bm{X}_i\}_{i=0}^T$ as
\begin{align*}
\bm{X}_i = \begin{cases}
&i^{-\alpha}\sum_{t=1}^i \left(\prod_{k=t}^i (\bm{I}-\eta_k\bm{A}) \bm{Y} \prod_{k=t}^i (\bm{I}-\eta_k\bm{A}^\top) \right), \quad \text{for} \quad 1 \leq i \leq T;\\ 
&\bm{0}, \quad \text{for} \quad i=0.
\end{cases}
\end{align*}

in this way, the left-hand-side of \eqref{eq:St2-bound} can be represented by $\bm{X}_T-\bm{X}$. It can be easily verified that the sequence $\{\bm{X}_i\}_{i =1}^T$ is featured by the iterative relation
\begin{align*}
i^{\alpha}\bm{X}_i &= (i-1)^{\alpha} (\bm{I}-\eta_i \bm{A})\bm{X}_{i-1} (\bm{I}-\eta_i \bm{A}^\top) +(\bm{I}-\eta_i \bm{A}) \bm{Y} (\bm{I}-\eta_i \bm{A}^\top), \quad \forall 1 \leq i \leq T.
\end{align*} 
Direct calculation yields that for every $1 \leq i \leq T$,
\begin{align*}
i^\alpha(\bm{X}_i - \bm{X}) &= (i-1)^{\alpha} (\bm{I}-\eta_i \bm{A})\bm{X}_{i-1} (\bm{I}-\eta_i \bm{A}^\top) +(\bm{I}-\eta_i \bm{A}) \bm{Y} (\bm{I}-\eta_i \bm{A}^\top) - i^{\alpha }\bm{X} \\ 
&= (i-1)^{\alpha} (\bm{I}-\eta_i \bm{A})(\bm{X}_{i-1}-\bm{X}) (\bm{I}-\eta_i \bm{A}^\top)\\
&+ (\bm{I}-\eta_i \bm{A}) \bm{Y} (\bm{I}-\eta_i \bm{A}^\top) - [i^{\alpha}-(i-1)^{\alpha}]\bm{X}-(i-1)^{\alpha}[\eta_i(\bm{AX}+\bm{XA}^\top) +\eta_i^2\bm{AXA}^\top].
% &+ \left(\frac{i-1}{i}\right)^{\alpha} (\bm{I}-\eta_i \bm{A})\bm{X} (\bm{I}-\eta_i \bm{A}^\top) - \bm{X} \\
% &+i^{-\alpha} (\bm{I}-\eta_i \bm{A}) \bm{Y} (\bm{I}-\eta_i \bm{A}^\top) \\ 
% &= \left(\frac{i-1}{i}\right)^{\alpha} (\bm{I}-\eta_i \bm{A})(\bm{X}_{i-1}-\bm{X}) (\bm{I}-\eta_i \bm{A}^\top) \\ 
% &+ i^{-\alpha}[\eta_0(\bm{AX} +\bm{XA}^\top)-\bm{Y}] + \frac{\bm{X}}{i} + O(i^{-2\alpha})\bm{X}.
\end{align*}
We now feature the terms on the last line. Firstly, we notice that \eqref{eq:Lyapunov-St2} guarantees
\begin{align*}
\eta_i(\bm{AX}+\bm{XA}^\top) = \eta_0 i^{-\alpha}(\bm{AX}+\bm{XA}^\top) = i^{-\alpha}\bm{Y};
\end{align*}
therefore, a simple reorganization yields
\begin{align*}
&(\bm{I}-\eta_i \bm{A}) \bm{Y} (\bm{I}-\eta_i \bm{A}^\top) - [i^{\alpha}-(i-1)^{\alpha}]\bm{X}-(i-1)^{\alpha}[\eta_i(\bm{AX}+\bm{XA}^\top) +\eta_i^2\bm{AXA}^\top] \\ 
&= \left[1-\left(\frac{i-1}{i}\right)^{\alpha}\right]\bm{Y} - [i^{\alpha}-(i-1)^{\alpha}]\bm{X} -\eta_i(\bm{AY} + \bm{YA}^\top) + \eta_i^2 \bm{A}(\bm{Y} - (i-1)^{\alpha} \bm{X})\bm{A}^\top.
\end{align*}
For these terms, we observe
\begin{align*}
&\left[1-\left(\frac{i-1}{i}\right)^{\alpha}\right]\bm{Y} \preceq \left[1-\left(\frac{i-1}{i}\right)\right]\bm{Y} = \frac{1}{i}\bm{Y};\\
&[i^{\alpha}-(i-1)^{\alpha}]\bm{X} = i^{\alpha}\left[1-\left(\frac{i-1}{i}\right)^{\alpha}\right]\bm{X} \preceq i^{\alpha}\left[1-\left(\frac{i-1}{i}\right)\right]\bm{X} = i^{\alpha-1} \bm{X} = O(i^{\alpha-1})\bm{Y};\\
&\eta_i(\bm{AY} + \bm{YA}^\top) = O(i^{-\alpha})\bm{Y};\\ 
&\eta_i^2 \bm{A}(\bm{Y} - (i-1)^{\alpha} \bm{X})\bm{A}^\top = O(i^{-\alpha})\bm{Y}.
\end{align*}
Hence, the leading term decreases by the rate of $O(i^{\alpha-1})\bm{Y}$. As a result,
\begin{align*}
i^\alpha(\bm{X}_i - \bm{X}) = (\bm{I}-\eta_i \bm{A})[(i-1)^{\alpha} (\bm{X}_{i-1}-\bm{X})] (\bm{I}-\eta_i \bm{A}^\top) + O(i^{\alpha-1})\bm{Y}
\end{align*}
For simplicity, we denote $\bm{Z}_i = i^{\alpha}(\bm{X}_i-\bm{X})$. The relation above can then be represented as
\begin{align*}
\bm{Z}_i = (\bm{I}-\eta_i \bm{A})\bm{Z}_{i-1}(\bm{I}-\eta_i \bm{A}^\top) + O(i^{\alpha-1})\bm{Y}.
\end{align*}
It is easy to verify by iteration that, there exists a quantity $\widetilde{C}$ independent of $T$, such that
\begin{align*}
&\sum_{i=1}^T i^{\alpha-1} \prod_{k=i+1}^T (\bm{I}-\eta_k \bm{A}) \bm{Y}\prod_{k=i+1}^T (\bm{I}-\eta_k \bm{A}^\top) \\ 
&\preceq \bm{Z}_T \preceq \sum_{i=1}^T i^{\alpha-1} \prod_{k=i+1}^T (\bm{I}-\eta_k \bm{A}) \bm{Y}\prod_{k=i+1}^T (\bm{I}-\eta_k \bm{A}^\top);
\end{align*}
In other words,
\begin{align*}
&T^{-\alpha} \sum_{i=1}^T i^{\alpha-1} \prod_{k=i+1}^T (\bm{I}-\eta_k \bm{A}) \bm{Y}\prod_{k=i+1}^T (\bm{I}-\eta_k \bm{A}^\top) \\
&\preceq \bm{X}_T-\bm{X} \\ 
&\preceq T^{-\alpha} \sum_{i=1}^T i^{\alpha-1} \prod_{k=i+1}^T (\bm{I}-\eta_k \bm{A}) \bm{Y}\prod_{k=i+1}^T (\bm{I}-\eta_k \bm{A}^\top).
\end{align*}
By letting $\beta = \frac{1-\gamma}{2}\lambda_0 \eta_0$, Lemma \ref{lemma:A} implies
\begin{align*}
&\left\|T^{-\alpha} \sum_{i=1}^T i^{\alpha-1} \prod_{k=i+1}^T (\bm{I}-\eta_k \bm{A}) \prod_{k=i+1}^T (\bm{I}-\eta_k \bm{A}^\top)\right\| \\ 
&\leq T^{-\alpha} \sum_{i=1}^T i^{\alpha-1} \prod_{k=i+1}^T  \|\bm{I}-\eta_k \bm{A}\|\prod_{k=i+1}^T \|\bm{I}-\eta_k \bm{A}^\top\| \\ 
&\leq T^{-\alpha} \sum_{i=1}^T i^{\alpha-1} \prod_{k=i+1}^T (1-\beta k^{-\alpha})^2 \\ 
&< T^{\alpha-1} \sum_{i=1}^T i^{-\alpha} \prod_{k=i+1}^T (1-\beta k^{-\alpha}) \\ 
&= T^{\alpha-1} \cdot \frac{1}{\beta} \sum_{i=1}^T \left(\prod_{k=i+1}^T (1-\beta k^{-\alpha})- \prod_{k=i}^T (1-\beta k^{-\alpha})\right) \\ 
&=\frac{1}{\beta} T^{\alpha-1} \left(\prod_{k=T+1}^T (1-\beta k^{-\alpha})- \prod_{k=1}^T (1-\beta k^{-\alpha})\right) < \frac{1}{\beta} T^{\alpha-1};
\end{align*}
notice that in the last two lines we invoked a similar approach to the telescoping method \eqref{eq:beta-telescope}. Combining this with \eqref{eq:lemma-O-matrix}, we obtain
\begin{align*}
T^{-\alpha} \sum_{i=1}^T i^{\alpha-1} \prod_{k=i+1}^T (\bm{I}-\eta_k \bm{A}) \bm{Y}\prod_{k=i+1}^T (\bm{I}-\eta_k \bm{A}^\top)= O(T^{\alpha-1})\bm{Y}.
\end{align*}
% Consequently, the difference between $\bm{X}_T$ and $\bm{X}$ is featured by
% \begin{align*}
% \bm{X}_T - \bm{X} = O(T^{-\alpha})\bm{X} + O(T^{\alpha-1})\bm{X} = O(T^{\alpha-1})\bm{X}.
% \end{align*}
% In the mean time, Lemma \ref{lemma:Lyapunov} essentially guarantees that $\bm{X} = O(1)\bm{Y}$. 
This completes the proof.
\qed

\begin{lemma}\label{lemma:Q-bound}
Let $\eta_t = \eta_0 t^{-\alpha}$ with $\alpha \in (\frac{1}{2},1)$ and $0 \leq \eta \leq \frac{1}{2\lambda_{\Sigma}}$. For $\bm{Q}_t$ defined as in \eqref{eq:defn-Qt}, it can be guaranteed that
\begin{align}
&\|\bm{Q}_t\| \leq 3\eta_0^{-\frac{\alpha}{1-\alpha}} \left(\frac{4\alpha}{\lambda_0(1-\gamma)}\right)^{\frac{1}{1-\alpha}}, \quad \forall t \in [T];  \label{eq:Q-bound} \\ 
&\|\bm{A} \bm{Q}_t\| \leq 2 + \eta_0 \left(\frac{1}{\beta}\right)^{\frac{1}{1-\alpha}}\Gamma \left(\frac{1}{1-\alpha}\right)t^{\alpha-1}; \label{eq:AQ-bound} \quad \text{and}\\ 
&\frac{1}{T}\sum_{t=1}^T \|\bm{Q}_t - \bm{A}^{-1}\|^2 \leq \frac{1}{\lambda_0(1-\gamma)}T^{\alpha-1} + \widetilde{C}T^{2\alpha-2}\label{eq:AQ-I-bound}
\end{align}
where $\beta = \frac{1-\gamma}{2}\lambda_0 \eta_0$  and $\widetilde{C}$ depends on $\alpha,\eta_0,\lambda_0,\gamma$.
\end{lemma}
\noindent \emph{Proof}: According to Lemma \ref{lemma:A}, since $0 < \eta_t \leq \eta_0 \leq \frac{1}{2\lambda_{\Sigma}}$ for all $t$, the norm of $\bm{Q}_t$ is bounded by
\begin{align*}
    \|\bm{Q}_t\|_2 &\leq \eta_t \sum_{j=t}^{T} \prod_{k=t+1}^{j} \left(1-\frac{1}{2}\lambda_0(1-\gamma)\eta_k\right)\\
    &=\eta_0 t^{-\alpha} \cdot \sum_{j=t}^{T} \prod_{k=t+1}^{j} \left(1-\frac{1}{2}\lambda_0(1-\gamma)\eta_0 k^{-\alpha}\right).
\end{align*}
The bound \eqref{eq:Q-bound} follows immediately by taking $\beta=\frac{1-\gamma}{2}\lambda_0 \eta_0$ in \eqref{eq:Q-uni-1}.
For the bound \eqref{eq:AQ-bound}, observe due to \eqref{eq:Qt-Ainv-decompose} that
\begin{align*}
\bm{AQ}_t - \bm{I} = \bm{AS}_t - \prod_{k=t}^T (\bm{I}-\eta_k \bm{A}).
\end{align*}
Therefore, the triangle inequality and \eqref{eq:St1-bound} directly imply
\begin{align*}
\|\bm{AQ}_t\| \leq \|\bm{I}\| + \|\bm{AS}_t\| + \left\|\prod_{k=t}^T (\bm{I}-\eta_k \bm{A})\right\| 
&\leq 1 + \eta_0 \left(\frac{1}{\beta}\right)^{\frac{1}{1-\alpha}}\Gamma \left(\frac{1}{1-\alpha}\right)t^{\alpha-1} + 1 \\ 
& = 2 + \eta_0 \left(\frac{1}{\beta}\right)^{\frac{1}{1-\alpha}}\Gamma \left(\frac{1}{1-\alpha}\right)t^{\alpha-1}.
\end{align*}
Furthermore, it can be easily verified that
\begin{align*}
\|\bm{AQ}_t-\bm{I}\|^2 
&\leq \|\bm{AS}_t\|^2 + \left\|\prod_{k=t}^T (\bm{I}-\eta_k \bm{A})\right\|^2 \\
&\leq \widetilde{C}' t^{2\alpha-2} + \prod_{k=t}^T(1-\beta k^{-\alpha})  \\
&< \widetilde{C}' t^{2\alpha-2} + T^{\alpha}(t-1)^{-\alpha}\prod_{k=t}^T(1-\beta k^{-\alpha}) \\ 
&= \widetilde{C}' t^{2\alpha-2} + T^{\alpha}\cdot \frac{1}{\beta} \left(\prod_{k=t}^T(1-\beta k^{-\alpha})-\prod_{k=t-1}^T(1-\beta k^{-\alpha})\right);
\end{align*}
The bound \eqref{eq:AQ-I-bound} follows by taking an average from $t=1$ to $t=T$ and applying the telescoping method as in \eqref{eq:beta-telescope}. 
\qed

The decomposition \eqref{eq:Qt-Ainv-decompose}, combined with the bounds \eqref{eq:St1-bound} Theorem \ref{thm:St2-bound}, implies the following bound on the difference between $\bar{\bm{\Lambda}}_T$ and $\bm{\Lambda}^\star$.

\begin{theorem}\label{thm:Lambda}
Let $\bar{\bm{\Lambda}}_T$ and $\bm{\Lambda}^\star$ be defined as in \eqref{eq:defn-barLambdaT} and \eqref{eq:defn-Lambdastar} respectively. Define $\bm{X}(\bm{\Lambda}^\star)$ as the unique solution to the Lyapunov equation
\begin{align}\label{eq:defn-X-Lambda}
\eta_0(\bm{AX+XA}^\top) = \bm{\Lambda}^\star;
\end{align}
then as $T \to \infty$, the difference between $\bar{\bm{\Lambda}_T}$ and $\bm{\Lambda}^\star$ satisfies
\begin{align*}
 \bar{\bm{\Lambda}}_T - \bm{\Lambda}^\star- T^{\alpha-1}\bm{X}(\bm{\Lambda}^\star)=O(T^{2\alpha-2})\bm{\Gamma}.
\end{align*}
Here, the $O$ notation for matrices is defined as in \eqref{eq:defn-O-matrix}.
%\textcolor{violet}{*ALE* Maybe better to say that $\| \bar{\bm{\Lambda}}_T - \bm{\Lambda}^\star \| = \bm{O}(T^{2\alpha-2})$ and then appeal to Weyl's inequality?}\weichen{please see the new notation.}
\end{theorem}
\noindent \emph{Proof:} By definition, the difference between $\bar{\bm{\Lambda}}_T$ and $\bm{\Lambda}^\star$ can be expressed as
\begin{align}\label{eq:delta-Lambda}
\bar{\bm{\Lambda}}_T - \bm{\Lambda}^\star 
&=\frac{1}{T}\sum_{t=1}^T (\bm{Q}_t \bm{\Gamma} \bm{Q}_t^\top - \bm{A}^{-1} \bm{\Gamma}\bm{A}^{-\top})\nonumber \\ 
&= \underset{I_1}{\underbrace{\frac{1}{T}\sum_{t=1}^T (\bm{Q}_t -\bm{A}^{-1}) \bm{\Gamma} \bm{A}^{-\top} + \frac{1}{T}\sum_{t=1}^T \bm{A}^{-1} \bm{\Gamma}  (\bm{Q}_t -\bm{A}^{-1})^\top}} \nonumber \\ 
&+ \underset{I_2}{\underbrace{\frac{1}{T}\sum_{t=1}^T (\bm{Q}_t - \bm{A}^{-1}) \bm{\Gamma} (\bm{Q}_t - \bm{A}^{-1})^{\top}}}.
\end{align}
We now proceed to bound the norms of $I_1$ and $I_2$ respectively.
\paragraph{Bounding the norm of $I_1$.} Notice that the first term can be expressed by
\begin{align}\label{eq:St-Lambda-norm}
\frac{1}{T} \sum_{t=1}^T (\bm{Q}_t - \bm{A}^{-1})\bm{\Gamma} \bm{A}^{-\top}&= -\frac{1}{T} \bm{A}^{-1}\sum_{j=1}^T \prod_{k=1}^j (\bm{I} -\eta_k \bm{A}) \cdot \bm{\Gamma} \bm{A}^{-\top}\nonumber \\ 
&=-\frac{1}{T} \sum_{j=1}^T \prod_{k=1}^j (\bm{I} -\eta_k \bm{A}) \cdot \bm{A}^{-1}\bm{\Gamma} \bm{A}^{-\top}\nonumber \\ 
& = -\frac{1}{T} \sum_{j=1}^T \prod_{k=1}^j (\bm{I} -\eta_k \bm{A}) \cdot \bm{\Lambda}^\star.
\end{align}
where we applied Equation \eqref{eq:Qt-Ainv-bound-1} the interchangeability of products. Similarly, the second term can be expressed by
\begin{align*}
\bm{A}^{-1}\bm{\Gamma} \frac{1}{T}\sum_{t=1}^T (\bm{Q}_t - \bm{A}^{-1})^{\top} = -\bm{\Lambda}^\star \cdot \frac{1}{T} \sum_{j=1}^T \prod_{k=1}^j (\bm{I} -\eta_k \bm{A})^\top.
\end{align*}
Lemma \ref{lemma:A} directly implies that
\begin{align*}
\frac{1}{T} \sum_{j=1}^T \prod_{k=1}^j (\bm{I} -\eta_k \bm{A})\leq \frac{1}{T}\prod_{k=1}^j \left(1-\frac{1-\gamma}{2}\lambda_0 \eta_0 k^{-\alpha}\right).
\end{align*}
For simplicity, we denote $\beta = \frac{1-\gamma}{2}\lambda_0 \eta_0$; it can then be seen that the summand on the right-hand-side can be bounded by
\begin{align}\label{eq:delta-Lambda-intersect-bound}
\sum_{j=1}^T \prod_{k=1}^j (1-\beta k^{-\alpha}) &< \sum_{j=1}^T \exp\left(\beta \sum_{k=1}^j k^{-\alpha}\right) \nonumber \\ 
&= \sum_{j=1}^T \exp\left(-\beta m_1^j\right) \\ 
&\asymp \sum_{j=1}^T \exp(-\beta m_1^j) (m_1^{j+1}-m_1^j)((1-\alpha)m_1^j)^{\frac{\alpha}{1-\alpha}}\nonumber \\ 
&\lesssim (1-\alpha)^{\frac{\alpha}{1-\alpha}} \int_0^\infty m^{\frac{\alpha}{1-\alpha}} \exp(-\beta m) \mathrm{d} m \nonumber \\ 
&= \beta^{-\frac{1}{1-\alpha}}(1-\alpha)^{\frac{\alpha}{1-\alpha}} \Gamma\left(\frac{1}{1-\alpha}\right),
\end{align}
a constant independent of $T$. Therefore, $I_1$ is featured by
\begin{align*}
\frac{1}{T}\sum_{t=1}^T (\bm{Q}_t -\bm{A}^{-1}) \bm{\Gamma} \bm{A}^{-\top} + \frac{1}{T}\sum_{t=1}^T \bm{A}^{-1} \bm{\Gamma}  (\bm{Q}_t -\bm{A}^{-1})^\top = O(T^{-1})\bm{\Lambda}^\star.
\end{align*}
\paragraph{Decomposing $I_2$.} The term $I_2$ can be further expressed by invoking the decomposition \eqref{eq:Qt-Ainv-decompose}. Specifically, we observe
\begin{align*}
I_2&= \underset{I_{21}}{\underbrace{\frac{1}{T}\sum_{t=1}^T\bm{S}_t \bm{\Gamma} \bm{S}_t^\top}} + \underset{I_{22}}{\underbrace{\frac{1}{T}\sum_{t=1}^T \prod_{k=t}^T (\bm{I}-\eta_k\bm{A})\bm{\Lambda}^\star \prod_{k=t}^T (\bm{I}-\eta_k\bm{A}^\top)}} \nonumber \\ 
&- \underset{I_{23}}{\underbrace{\left(\frac{1}{T} \sum_{t=1}^T \bm{S}_t \bm{\Gamma} \bm{A}^{-\top}\prod_{k=t}^T (\bm{I}-\eta_k \bm{A}^\top) +\frac{1}{T}\sum_{t=1}^T \prod_{k=t}^T (\bm{I}-\eta_k \bm{A})\bm{A}^{-1}\bm{\Gamma} \bm{S}_t^\top\right)}}.
\end{align*}
Since $\|\bm{S}_t\| = O(T^{\alpha-1})$, it is easy to verify that $I_{21} = O(T^{2\alpha-2})\bm{\Gamma}$; meanwhile, theorem \ref{thm:St2-bound} indicates that $I_{22} = T^{\alpha-1} \bm{X}(\bm{\Lambda}^\star) +O(T^{2\alpha-2})\bm{\Lambda}^\star$. As for the term $I_{23}$, we observe that 
\begin{align*}
\left\|\sum_{t=1}^T \bm{S}_t  \prod_{k=t}^T (\bm{I}-\eta_k \bm{A}^\top)\right\| 
&\lesssim \widetilde{C}\sum_{t=1}^T t^{\alpha-1} \prod_{k=t}^T (1-\beta k^{-\alpha}) \\ 
&<  \widetilde{C}T^{2\alpha-1} \sum_{t=1}^T t^{\alpha-1} \prod_{k=t+1}^T (1-\beta k^{-\alpha}) \\ 
&= \widetilde{C}T^{2\alpha-1} \cdot \frac{1}{\beta} \sum_{t=1}^T \left(\prod_{k=t+1}^T - \prod_{k=t}^T \right)(1-\beta k^{-\alpha})\\ 
&<\widetilde{C}\frac{T^{2\alpha-1}}{\beta},
\end{align*}
where $\widetilde{C}$ is a problem-related quantity related to $\eta_0, \alpha$ and $\gamma$. 
Consequently, it can be guaranteed that $I_{23} = O(T^{2\alpha-2})\bm{\Gamma}$. In combination, we have shown that
\begin{align*}
\bar{\bm{\Lambda}}_T - \bm{\Lambda}^\star &= I_1 + I_{21} + I_{22} + I_{23} \\ 
&= O(T^{-1})\bm{\Lambda}^\star + O(T^{2\alpha-2})\bm{\Gamma} + (T^{\alpha-1} \bm{X}(\bm{\Lambda}^\star) +O(T^{2\alpha-2})\bm{\Lambda}^\star) + O(T^{2\alpha-2})\bm{\Gamma}\\
&= T^{\alpha-1} \bm{X}(\bm{\Lambda}^\star) + O(T^{2\alpha-2})\bm{\Gamma} + O(T^{2\alpha-2})\bm{\Lambda}^\star;
\end{align*}
the theorem follows naturally by combining this with the fact that $\bm{\Lambda}^\star = O(1)\bm{\Gamma}$, due to \eqref{eq:lemma-A-6}.\qed

The following Lemma controls the minimum eigenvalue of $\bm{A}\bar{\bm{\Lambda}}_T \bm{A}^\top$.
\begin{lemma}\label{lemma:A-Lambda}
With $\bm{A}$ defined as \eqref{eq:defn-At-mean} and $\bar{\bm{\Lambda}}_T$ defined as in \eqref{eq:defn-barLambdaT}, when 
\begin{align}\label{eq:Lambda-T-condition}
T \geq 4 \left(\frac{2}{(1-\gamma)\lambda_0\eta_0}\right)^{\frac{1}{1-\alpha}}(1-\alpha)^{\frac{\alpha}{1-\alpha}}\Gamma(\frac{1}{1-\alpha})\mathsf{cond}(\Gamma),
\end{align}
it can be guaranteed that
\begin{align*}
\lambda_{\min}(\bm{A}\bm{\Lambda}_T \bm{A}^\top) \geq \frac{1}{2}\lambda_{\min}(\bm{\Gamma}) = \frac{1}{2\|\bm{\Gamma}^{-1}\|}.
\end{align*}
\end{lemma}
\noindent\emph{Proof}: We firstly decompose $\bm{A}\bm{\Lambda}_T \bm{A}^\top$ as
\begin{align*}
\bm{A}\bm{\Lambda}_T \bm{A}^\top &= \bm{A}\bm{\Lambda}^\star \bm{A}^\top + \bm{A}(\bm{\Lambda}_T - \bm{\Lambda}^\star) \bm{A}^\top \\ 
&= \bm{\Gamma} + \bm{A}(\bm{\Lambda}_T - \bm{\Lambda}^\star) \bm{A}^\top;
\end{align*} 
here, \eqref{eq:delta-Lambda} indicates that the difference $(\bm{\Lambda}_T - \bm{\Lambda}^\star)$ is bounded above by
\begin{align*}
\bm{\Lambda}_T - \bm{\Lambda}^\star & \succeq \frac{1}{T}\sum_{t=1}^T \left[(\bm{Q}_t - \bm{A}^{-1})\bm{\Gamma}\bm{A}^{-\top} + \bm{A}^{-1}\bm{\Gamma}(\bm{Q}_t - \bm{A}^{-1})^\top \right].
\end{align*}
Furthermore, \eqref{eq:Qt-Ainv-bound-1} reveals that
\begin{align*}
&\bm{A}(\bm{\Lambda}_T - \bm{\Lambda}^\star) \bm{A}^\top \\ 
&= \frac{1}{T}\bm{A} \sum_{t=1}^T (\bm{Q}_t - \bm{A}^{-1})\bm{\Gamma}\bm{A}^{-\top}\bm{A}^\top+\frac{1}{T}\bm{A}\bm{A}^{-1} \sum_{t=1}^T \bm{\Gamma}(\bm{Q}_t - \bm{A}^{-1})^\top\bm{A}^\top \\ 
&=-\frac{1}{T}\bm{A} \bm{A}^{-1}\sum_{j=1}^T \prod_{k=1}^j (\bm{I}-\eta_k \bm{A})\bm{\Gamma}-\frac{1}{T} \bm{\Gamma} \sum_{j=1}^T \prod_{k=1}^j (\bm{I}-\eta_k \bm{A})^\top \bm{A}^{-\top} \bm{A}^{\top} \\ 
&= -\frac{1}{T} \sum_{j=1}^T \prod_{k=1}^j (\bm{I}-\eta_k \bm{A})\bm{\Gamma} -\frac{1}{T} \bm{\Gamma} \sum_{j=1}^T \prod_{k=1}^j (\bm{I}-\eta_k \bm{A})^\top.
\end{align*}
It has been revealed in \eqref{eq:delta-Lambda-intersect-bound} that
\begin{align*}
\left\|\sum_{j=1}^T \prod_{k=1}^j (\bm{I}-\eta_k \bm{A})\right\| \leq \left(\frac{2}{(1-\gamma)\lambda_0\eta_0}\right)^{\frac{1}{1-\alpha}}(1-\alpha)^{\frac{\alpha}{1-\alpha}}\Gamma(\frac{1}{1-\alpha}).
\end{align*}
Consequently, the triangle inequality yields
\begin{align*}
\lambda_{\min}(\bm{A}\bm{\Lambda}_T \bm{A}^\top) &\geq \lambda_{\min}(\bm{\Gamma}) - \left\|\frac{1}{T} \sum_{j=1}^T \prod_{k=1}^j (\bm{I}-\eta_k \bm{A})\bm{\Gamma} +\frac{1}{T} \bm{\Gamma} \sum_{j=1}^T \prod_{k=1}^j (\bm{I}-\eta_k \bm{A})^\top\right\| \\ 
&\geq  \lambda_{\min}(\bm{\Gamma}) - \frac{2}{T}\left(\frac{2}{(1-\gamma)\lambda_0\eta_0}\right)^{\frac{1}{1-\alpha}}(1-\alpha)^{\frac{\alpha}{1-\alpha}}\Gamma(\frac{1}{1-\alpha}) \|\bm{\Gamma}\|.
\end{align*}
The lemma follows immediately by applying the condition \eqref{eq:Lambda-T-condition}.
\qed

\subsection{Gaussian approximation}
This section is concerned with Gaussian approximation of random variables. We will firstly introduce some Berry-Esseen bounds and then illustrate some properties of the multi-dimensional Gaussian distribution.

\subsubsection{Berry-Esseen bounds}
The following theorem, due to \citet{shao2021berry}, gives a Berry-Esseen bound for non-linear statistics which would be crucial for our analysis of the non-asymptotic distribution of the TD estimation error.
\begin{theorem}\label{thm:Shao}
(Theorem 2.1 of \citet{shao2021berry}) Let $\bm{X}_1,...,\bm{X}_n$ be independent random variables, and let $\bm{T}(\bm{X}_1,...,\bm{X}_n)$ be a $d$-dimensional statistic that can be represented as
\begin{align*}
\bm{T} = \bm{W}+\bm{D},
\end{align*}
where $\bm{W}$ is the sum of independent random variables
\begin{align*}
\bm{W} = \sum_{i=1}^n \bm{\xi}_i = \sum_{i=1}^n \bm{h}_i(\bm{X}_i)
\end{align*}
for Borel measurable functions $\{\bm{h}_i\}_{1\leq i\leq n}$ that satisfies
\begin{align*}
\mathbb{E}[\bm{\xi}_i] = 0, \text{ for each } 1\leq i \leq n \text{ and } \sum_{i=1}^n \mathbb{E}[\bm{\xi}_i \bm{\xi}_i^\top] = \bm{\Lambda}.
\end{align*}
Further assume that $\lambda_{\min}(\bm{\Lambda}) \geq \sigma > 0$. Let $\bm{z} \sim N(0,\bm{I}_d)$. Then for any convex set $\mathcal{A} \subseteq \mathbb{R}^d$,
\begin{align}\label{eq:thm-Shao}
d_{\mathsf{C}}(\bm{T},\bm{\Lambda}^{\frac{1}{2}}\bm{z}) \leq 259 \sqrt{d}\sigma^{-\frac{3}{2}} \cdot \sum_{i=1}^n \mathbb{E}\|\bm{\xi}_i\|_2^3 + 2\sigma^{-1}\mathbb{E}\{\|\bm{W}\|_2\Delta\} + 2\sigma^{-1}\sum_{i=1}^n \mathbb{E}\{\|\bm{\xi}_i\|_2|\Delta - \Delta^{(i)}|\},
\end{align}
for any random variables $\Delta$ and $\{\Delta^{(i)}\}_{1\leq i\leq n}$ such that $\Delta \geq \|D\|_2$ and $\Delta^{(i)}$ is independent of $\bm{X}_i$.
\end{theorem}

\subsubsection{Properties of multivariate Gaussian distributions}
The following theorem  by \citet{devroye2018total} gives an upper bound for the total variation (TV) distance between two Gaussian random variables with same means and different covariance matrices.
\begin{theorem}[\citet{devroye2018total}]\label{thm:DMR}
 Let $\bm{\Lambda}_1, \bm{\Lambda}_2 \in \mathbb{S}^{d \times d}$ be two positive definite matrices, and $\bm{\mu}$ be any vector in $\mathbb{R}^d$. Then the TV distance between Gaussian distributions $\mathcal{N}(\bm{\mu},\bm{\Lambda}_1)$ and $\mathcal{N}(\bm{\mu},\bm{\Lambda}_2)$ is bounded by
\begin{align*}
&\min\left\{1,\frac{1}{100}\left\|\bm{\Lambda}_1^{-1/2}\bm{\Lambda}_2 \bm{\Lambda}_1^{-1/2}-\bm{I}_d\right\|_{\mathsf{F}} \right\} \\ 
& \leq d_{\mathsf{TV}}(\mathcal{N}(\bm{\mu},\bm{\Lambda}_1),\mathcal{N}(\bm{\mu},\bm{\Lambda}_2)) \\ 
&\leq \frac{3}{2}\left\|\bm{\Lambda}_1^{-1/2}\bm{\Lambda}_2 \bm{\Lambda}_1^{-1/2}-\bm{I}_d\right\|_{\mathsf{F}}. 
\end{align*}
\end{theorem}

In the case of $\bm{\Lambda}_1 \succ \bm{\Lambda}_2$, the following proposition further specifies that the TV distance and the distance on convex sets are the same between $\mathcal{N}(\bm{0},\bm{\Lambda}_1)$ and $\mathcal{N}(\bm{0},\bm{\Lambda}_2)$.
\begin{proposition}\label{prop:Gaussian}
 Let $\bm{\Lambda}_1, \bm{\Lambda}_2 \in \mathbb{S}^{d \times d}$ be two positive definite matrices, and $\bm{\mu}$ be any vector in $\mathbb{R}^d$. When $\bm{\Lambda}_1 \succ \bm{\Lambda_2}$, it can be guaranteed that
 \begin{align*}
d_{\mathsf{C}}(\mathcal{N}(\bm{\mu},\bm{\Lambda}_1),\mathcal{N}(\bm{\mu},\bm{\Lambda}_2)) = d_{\mathsf{TV}}(\mathcal{N}(\bm{\mu},\bm{\Lambda}_1),\mathcal{N}(\bm{\mu},\bm{\Lambda}_2)).
 \end{align*}
\end{proposition}
\noindent \emph{Proof}: Without loss of generality, let $\bm{\mu} = \bm{0}$. Define $\mathcal{C}$ as
\begin{align*}
\mathcal{C} := \{\bm{x} \in \mathbb{R}^d\mid p_1(\bm{x}) < p_2(\bm{x})\},
\end{align*}
where $p_1(\cdot)$ and $p_2(\cdot)$ denote the probability density functions of $\mathcal{N}(\bm{0},\bm{\Lambda}_1)$ and $\mathcal{N}(\bm{0},\bm{\Lambda}_2)$ respectively; we aim to show that $\mathcal{C}$ is a convex set. Notice that the ratio between these two density functions is featured by
\begin{align*}
\frac{p_1(\bm{x})}{p_2(\bm{x})} &= \frac{\frac{1}{\sqrt{\mathsf{det}(\bm{\Lambda_1})}}\exp(-\bm{x}^\top \bm{\Lambda_1}^{-1}\bm{x})}{\frac{1}{\sqrt{\mathsf{det}(\bm{\Lambda_2})}}\exp(-\bm{x}^\top \bm{\Lambda_2}^{-1}\bm{x})} \\ 
&= \sqrt{\frac{\mathsf{det}(\bm{\Lambda}_2)}{\mathsf{det}(\bm{\Lambda_1})}}\exp(\bm{x}^\top (\bm{\Lambda_2}^{-1} - \bm{\Lambda_1}^{-1})\bm{x}).
\end{align*}
Therefore by definition, the set $\mathcal{C}$ is characterized by
\begin{align*}
\mathcal{C} = \left\{\bm{x} \in \mathbb{R}^d \bigg| \bm{x}^\top (\bm{\Lambda_2}^{-1} - \bm{\Lambda_1}^{-1})\bm{x} < \frac{1}{2} \log \frac{\mathsf{det}(\bm{\Lambda}_1)}{\mathsf{det}(\bm{\Lambda}_2)}\right\}.
\end{align*}
Since $\bm{\Lambda_1} \succ \bm{\Lambda}_2$, the matrix $(\bm{\Lambda_2}^{-1} - \bm{\Lambda_1}^{-1})$ must be positive definite; consequently, $\mathcal{C}$ is a $d$-dimensional hyper-ellipsoid and is therefore convex. With $\mathcal{C}$ being convex, it is seen that
\begin{align*}
&\left|\mathbb{P}_{z \sim \mathcal{N}(\bm{0},\bm{\Lambda}_1)}(\bm{z} \in \mathcal{C}) - \mathbb{P}_{z \sim \mathcal{N}(\bm{0},\bm{\Lambda}_2)}(\bm{z} \in \mathcal{C})\right| \\ 
&\leq \sup_{\mathcal{A} \in \mathscr{C}_d}\left|\mathbb{P}_{z \sim \mathcal{N}(\bm{\mu},\bm{\Lambda}_1)}(\bm{z} \in \mathcal{A}) - \mathbb{P}_{z \sim \mathcal{N}(\bm{\mu},\bm{\Lambda}_2)}(\bm{z} \in \mathcal{A})\right| \\ 
&= d_{\mathsf{C}}(\mathcal{N}(\bm{0},\bm{\Lambda}_1),\mathcal{N}(\bm{0},\bm{\Lambda}_2)),
\end{align*}
while in the meantime
\begin{align*}
 \left|\mathbb{P}_{z \sim \mathcal{N}(\bm{\mu},\bm{\Lambda}_1)}(\bm{z} \in \mathcal{C}) - \mathbb{P}_{z \sim \mathcal{N}(\bm{\mu},\bm{\Lambda}_2)}(\bm{z} \in \mathcal{C})\right| &=  d_{\mathsf{TV}}(\mathcal{N}(\bm{\mu},\bm{\Lambda}_1),\mathcal{N}(\bm{\mu},\bm{\Lambda}_2))  \\ 
 &\geq d_{\mathsf{C}}(\mathcal{N}(\bm{0},\bm{\Lambda}_1),\mathcal{N}(\bm{0},\bm{\Lambda}_2)). 
\end{align*}
Combining these two inequalities completes the proof. \qed

% \begin{theorem}\label{thm:Gaussian-dK}
% Let $\bm{\Lambda}_1, \bm{\Lambda}_2$ be two positive-definite matrices in $\mathbb{S}^{d \times d}$. Denote $\bm{D} = \mathsf{diag}(\bm{\Lambda}_1)$, and therefore $\bm{\Pi} = \bm{D}^{-\frac{1}{2}}\bm{\Lambda}_1\bm{D}^{-\frac{1}{2}}$ is the correlation matrix corresponding to $\bm{\Lambda}_1$. 
% \end{theorem}

\subsection{Concentration inequalities}
The following Azuma's inequality for matrix martingales is shown in \citet{Tropp2011matrixtails}.
\begin{theorem}\label{thm:matrix-Azuma}
Let $\{\bm{X}_k\}$ be a random sequence of matrices in $\mathbb{S}^{d \times d}$, and $\{\bm{A}_k\}$ be a sequence of fixed matrices in $\mathbb{S}^{d \times d}$. Assume that $\mathbb{E}_{k-1}[\bm{X}_k] = \bm{0}$ and $\bm{X}_k^2 \preceq \bm{A}_k^2$ almost surely. Define
\begin{align*}
\sigma^2 := \left\|\sum_k \bm{A}_k^2\right\|,
\end{align*}
then for all $\varepsilon >0$, 
\begin{align*}
\mathbb{P}\left(\left\|\sum_k \bm{X}_k\right\| \geq \varepsilon \right) \leq d \exp\left(-\frac{\varepsilon^2}{8\sigma^2}\right).
\end{align*}
\end{theorem}

Specifically, for vector-valued martingales, the following proposition is shown by \citet{kallenberg1991some}.
\begin{proposition}\label{prop:vector-mtg}
Let $\{\bm{x}_i\}_{i \geq 1}$ be a martingale in $\mathbb{R}^d$. Then there exists a martingale $\{\bm{y}_i\}_{i \geq 1}$ in $\mathbb{R}^2$ such that for every $i > 1$, $\|\bm{y}_i\|_2$ has the same distribution as $\|\bm{x}_i\|_2$ and $\|\bm{y}_i - \bm{y}_{i-1}\|_2$ has the same distribution as $\|\bm{x}_i - \bm{x}_{i-1}\|_2$.
\end{proposition}

The following Azuma's inequality on vector-valued martingales comes as a direct corollary.
\begin{corollary}\label{cor:vector-Azuma}
Let $\{\bm{x}_i\}_{i \geq 1}$ be a martingale in $\mathbb{R}^d$, and let $W_{\max}$ be a positive constant that satisfies
\begin{align*}
W_{\max} \geq \sum_{i=1}^t \|\bm{x}_i - \bm{x}_{i-1}\|_2^2 \quad \text{almost surely.}
\end{align*}
Then it can be guaranteed with probability at least $1-\delta$ that
\begin{align*}
\|\bm{x}_t\|_2 \leq 2\sqrt{2W_{\max}\log \frac{3}{\delta}}.
\end{align*}
\end{corollary}

The following Freedman's inequality on Hilbert spaces was shown by \cite{peng2024advances} in their Theorem 2.1. 
\begin{theorem}\label{thm:Hilbert-Freedman}
Let $\mathcal{X}$ be a Hilbert space, $\{\bm{X}_i\}_{i=1}^n$ be an $\mathcal{X}$-valued martingale difference sequence adapted to the filtration $\{\mathcal{F}_i\}_{i=1}^n$, $\bm{Y}_i := \sum_{j=1}^i \bm{X}_j$ be the corresponding martingale, and $W_i = \sum_{j=1}^i \mathbb{E}\|\bm{X}_j\|^2\mid\mathscr{F}_{j-1}$. Suppose that $\max_{i \in [n]}\|\bm{X}_i\| \leq b$ for some constant $b > 0$. Then for any $\varepsilon$ and $\sigma>0$, it can be guaranteed that
\begin{align*}
\mathbb{P}\left(\exists k \in [n], \text{ s.t. } \|\bm{Y}_k\|\geq \varepsilon \text{ and } W_k \leq \sigma^2 \right) \leq 2\exp\left(-\frac{\varepsilon^2/2}{\sigma^2 + b\varepsilon/3}\right).
\end{align*}
\end{theorem}

\section{Proof of main results}
\label{app:proofs}

\subsection{Proof of Theorem \ref{thm:bar-delta-t}}\label{app:proof-whp}
In order to derive the high-probability convergence guarantee for $\bar{\bm{\Delta}}_T$, we firstly propose the following result on the time-uniform convergence of the original TD estimator $\bm{\Delta}_t$.
\subsubsection{High-probability convergence guarantee for the original TD estimation error}
\begin{theorem}\label{thm:delta-t}
Consider the TD algorithm \eqref{eq:TD-update-all} with independent samples and decaying stepsizes $\eta_t = \eta_0t^{-\alpha}$, where $\alpha \in (\frac{1}{2},1)$.
For any $\delta \in (0,1)$, there exists $\eta_0 > 0$, such that with probability at least $1-\delta$, the norm of $\bm{\Delta}_t$ is bounded by
\begin{align}\label{eq:delta-t-alpha-bound-iid}
\|\bm{\Delta}_t\|_2 &\lesssim \frac{\eta_0}{\sqrt{2\alpha-1}}\left(\frac{(1-\gamma) \lambda_0 \eta_0}{4\alpha}\right)^{-\frac{\alpha}{2(1-\alpha)}}\sqrt{\log \frac{9T}{\delta}}(2\|\bm{\theta}^\star\|_2+1)t^{-\frac{\alpha}{2}} 
\end{align}
simultaneously for all $t \in [T]$.
\end{theorem}
\noindent \emph{Proof:} Under the TD update rule \eqref{eq:TD-update-all}, we represent $\bm{\Delta}_t$ as
\begin{align*}
\bm{\Delta}_t = \bm{\theta}_t - \bm{\theta}^\star
&= (\bm{\theta}_{t-1}-\eta_t(\bm{A}_{t}\bm{\theta}_{t-1}-\bm{b}_{t})) - \bm{\theta}^\star\\
&= \bm{\Delta}_{t-1} - \eta_t (\bm{A}\bm{\theta}_{t-1}-\bm{b} + \bm{\zeta}_{t})\\
&= (\bm{I}-\eta_t \bm{A})\bm{\Delta}_{t-1} -\eta_t \bm{\zeta}_{t},
\end{align*}
where $\bm{\zeta}_t$ is defined as
\begin{align}\label{eq:defn-zetat}
\bm{\zeta}_t = (\bm{A}_t -\bm{A})\bm{\theta}_{t-1} - (\bm{b}_t-\bm{b}).
\end{align}
Therefore by induction, $\bm{\Delta}_t$ can be expressed as a weighted sum of $\{\bm{\zeta}_i\}_{0 \leq i < t}$, namely
\begin{align}\label{eq:delta-t}
\bm{\Delta}_t &= \prod_{k=1}^t (\bm{I}-\eta_k \bm{A}) \Delta_0 -\sum_{i=1}^{t} \bm{R}_{i}^t \bm{\zeta}_{i} \nonumber \\ 
&= \prod_{k=1}^t (\bm{I}-\eta_k \bm{A})\Delta_0 - \sum_{i=1}^{t} \bm{R}_{i}^t (\bm{A}_i\bm{\theta}^\star -\bm{b}_i) - \sum_{i=1}^{t} \bm{R}_{i}^t (\bm{A}_i - \bm{A})\bm{\Delta}_{i-1}
\end{align}
in which the matrices $\{\bm{R}_i^t\}_{1 \leq i \leq t}$ are defined as 
\begin{align}\label{eq:defn-Rit}
\bm{R}_i^t = \eta_i\prod_{k=i+1}^t (\bm{I}-\eta_k \bm{A}).
\end{align}
% \begin{align}\label{eq:defn-Rit}
% \bm{R}_i^t = \eta_i \prod_{k=i+1}^t (\bm{I}-\eta_k \bm{A}).
% \end{align}

% The triangle inequality directly implies the following upper bound for $\|\bm{\Delta}_t\|_2$:
% \begin{align*}
% \|\bm{\Delta}_t\|_2 \leq \underset{I_1}{\underbrace{\left\|\prod_{k=1}^t (\bm{I}-\eta_k \bm{A}) \Delta_0 \right\|_2}} + \underset{I_2}{\underbrace{\left\|\sum_{i=1}^t \bm{R}_i^t (\bm{A}_i \bm{\theta}^\star - \bm{b}_i)\right\|_2}} + \underset{I_3} {\underbrace{\left\| \sum_{i=1}^t \bm{R}_i^t (\bm{A}_i - \bm{A}) \bm{\Delta}_{i-1} \right\|_2}}.
% \end{align*}
For simplicity, in this proof we denote
\begin{align*}
&\beta = \frac{1-\gamma}{2}\lambda_0\eta_0,\quad R = \frac{5}{2}\|\bm{\theta}^\star\|_2 + \frac{1}{4}, \quad \text{and}\\
&C = \frac{2\alpha-1}{512\eta_0^2} \left(\frac{\beta}{2\alpha}\right)^{\frac{\alpha}{1-\alpha}}.
\end{align*}
For any given $\delta$, we define $t^\star$ as a function of $\delta$, specifically
\begin{align}\label{eq:defn-tstar-alpha}
t^\star(\delta) = \inf\left\{t \in \mathbb{N}^+: \int_{t}^\infty \exp(-Cx^{\alpha})\mathrm{d}x \leq \frac{\delta}{9}\right\},
\end{align}
and assume that
\begin{align}\label{eq:delta-t-condition}
16\eta_0\sqrt{\frac{2}{2\alpha-1}} \sqrt{\log \frac{9t^\star}{\delta}}  \leq 1.
\end{align}

Furthermore, we denote, for simplicity,
\begin{align*}
\mathcal{H}_i = \{\max_{1 \leq j \leq i} \|\bm{\Delta}_j\| \leq R\},\quad \text{and} \quad \tilde{\bm{\zeta}}_i = \bm{\zeta}_i \mathds{1}(\mathcal{H}_{i-1}).
\end{align*}
It can then be easily verified that
\begin{align*}
\|\tilde{\bm{\zeta}}_i\|_2 \leq 4R + 2\|\bm{\theta}^\star\|_2 + 1.
\end{align*}
In what follows, we prove the desired result in the following 4 steps:
\begin{enumerate}
\item Showing the existence of $t^\star(\delta)$ and featuring its relationship with $\delta$;
\item Showing that $\mathbb{P}(\mathcal{H}_{t^\star}) \geq 1-\frac{1}{3}\delta$;
\item Showing that $\mathbb{P}(\mathcal{H}_{\infty}) \geq 1-\frac{2}{3}\delta$;
\item Showing the high-probability bound \eqref{eq:delta-t-alpha-bound-iid}.
\end{enumerate}
\paragraph{Step 1: the existence of $t^\star(\delta)$.} By definition, $t^\star$ depends on the integral
\begin{align*}
\int_{t}^\infty \exp(-Cx^{\alpha})\mathrm{d}x,
\end{align*}
which we handle with a change of variable
\begin{align*}
u = x^{\alpha}.
\end{align*}
It is easy to verify that
\begin{align*}
\mathrm{d}x = \frac{1}{\alpha}u^{\frac{1-\alpha}{\alpha}} \mathrm{d}u,
\end{align*}
and hence
\begin{align*}
\int_{t}^\infty \exp(-Cx^{\alpha})\mathrm{d}x &= \frac{1}{\alpha} \int_{t^{\alpha}}^{\infty} u^{\frac{1-\alpha}{\alpha}}\exp(-Cu)\mathrm{d}u.
\end{align*}
since $\alpha > 0.5$, it can be guaranteed that $u^{\frac{1-\alpha}{\alpha}} < u$, for any $u > t^{\alpha} \geq 1$. Hence, this integral is bounded by
\begin{align*}
\int_{t}^\infty \exp(-Cx^{\alpha})\mathrm{d}x &< \frac{1}{\alpha}\int_{t^{\alpha}}^{\infty} u\exp(-Cu) \mathrm{d}u \\ 
&= \frac{1}{\alpha C^2} (Cu+1) \exp(-Cu).
\end{align*}
Consequently, in order that
\begin{align*}
\int_{t}^\infty \exp(-Cx^{\alpha})\mathrm{d}x \leq \frac{\delta}{9},
\end{align*}
it suffices to let
\begin{align*}
(Ct^{\alpha}+1) \exp(-Ct^{\alpha}) \leq \frac{\alpha C^2\delta}{9};
\end{align*}
It is easy to verify that this condition is satisfied by taking
\begin{align*}
t = \left[\frac{2}{C} \left(\log \frac{9}{\alpha C^2 \delta} \vee \frac{5}{2}\right)\right]^{\frac{1}{\alpha}}.
\end{align*}
In other words, $t^{\star}(\delta)$ is a well-defined function of $\delta$ that scales as $O\left(\log^{\alpha} \frac{1}{\delta}\right)$ as $\delta \to 0$.
\paragraph{Step 2: Bounding $\mathbb{P}(\mathcal{H}_{t^\star})$.} We aim to prove, through an induction argument, that
\begin{align}\label{eq:delta-t-alpha-induction-1}
\mathbb{P}(\mathcal{H}_t) \geq 1-\frac{t}{t^\star}\cdot \frac{\delta}{3}.
\end{align}
Since $\bm{\theta}_0 = 0$, the norm of $\bm{\Delta}_0$ is equal to that of $\bm{\theta}^\star$, and therefore $\mathbb{P}(\mathcal{H}_0) = 1$ is trivial. When \eqref{eq:delta-t-alpha-induction-1} holds for $t-1$, we consider the probability of $\mathcal{H}_{t}$. 
We firstly observe that
\begin{align*}
\left\|\sum_{i=1}^{t} \bm{R}_i^{t} \tilde{\bm{\zeta}}_i\right\|_2^2 &\leq \sum_{i=1}^{t} \|\bm{R}_i^{t}\|^2 \|\tilde{\bm{\zeta}}_i\|_2^2 \\ 
&< \eta_0^2 (4R+2\|\bm{\theta}^\star\|_2+1)^2 \underset{W_t}{\underbrace{\sum_{i=1}^{t} i^{-2\alpha} \prod_{k=1}^t (1-\beta k^{-\alpha})}}.
\end{align*}
The term $W_t$ is bounded by $ \sum_{i=1}^t i^{-2\alpha} < \frac{1}{2\alpha-1}$.
Therefore, the vectors Azuma's inequality (Corollary \ref{cor:vector-Azuma}) shows that, with probability at least $1-\frac{\delta}{3t^\star}$,
\begin{align*}
\left\|\sum_{i=1}^{t} \bm{R}_i^{t} \tilde{\bm{\zeta}}_i\right\|_2 &< 2\eta_0(4R+2\|\bm{\theta}^\star\|_2+1)\sqrt{\frac{2}{2\alpha-1}\log \frac{9t^\star}{\delta}} \\ 
&\leq \frac{1}{8}(4R+2\|\bm{\theta}^\star\|_2+1),
\end{align*}
where we invoked the condition \eqref{eq:delta-t-condition} in the second inequality. The triangle inequality and a union bound argument yield that
\begin{align*}
\|\bm{\Delta}_{t}\| &\leq \left\|\prod_{k=1}^{t} (\bm{I}-\eta_k \bm{A}) \bm{\Delta}_0\right\|_2 + \left\|\sum_{i=1}^{t} \bm{R}_i^{t} \bm{\zeta}_i\right\|_2 \\ 
&= \left\|\prod_{k=1}^{t} (\bm{I}-\eta_k \bm{A}) \bm{\Delta}_0\right\|_2 + \left\|\sum_{i=1}^{t} \bm{R}_i^{t} \tilde{\bm{\zeta}}_i\right\|_2 \\ 
&< \|\bm{\theta}^\star\|_2 + \frac{1}{8}(4R+2\|\bm{\theta}^\star\|_2+1) = R,
\end{align*}
with probability at least $\mathbb{P}(\mathcal{H}_{t-1}) - \frac{\delta}{3t^\star} > 1-\frac{t}{t^\star} \cdot \frac{\delta}{3}$. The induction assumption then guarantees $\mathbb{P}(\mathcal{H}_{t^\star}) \geq 1-\frac{\delta}{3}$.
\paragraph{Step 3: Bounding $\mathbb{P}(\mathcal{H}_{\infty})$.} When $t > t^\star$, we invoke a tighter bound on $W_t$ to bound the norm of $\bm{\Delta}_t$. Namely, Lemma \ref{lemma:R} indicates that
\begin{align*}
W_t \leq \frac{1}{2\alpha-1}\left(\frac{2\alpha}{\beta}\right)^{\frac{\alpha}{1-\alpha}}t^{-\alpha}.
\end{align*}
Furthermore, define
\begin{align*}
\delta_t = 3\exp\left(-\frac{2\alpha-1}{512\eta_0^2}\left(\frac{\beta}{2\alpha}\right)^{\frac{\alpha}{1-\alpha}}t^{\alpha}\right)=3\exp(-Ct^{\alpha});
\end{align*}
it can then be seen that with probability at least $1-\delta_t$,
\begin{align*}
\left\|\sum_{i=1}^{t} \bm{R}_i^{t} \tilde{\bm{\zeta}}_i\right\|_2 &\leq 2\eta_0(4R + 2\|\bm{\theta}^\star\|_2+1)\sqrt{2W_t \log \frac{3}{\delta_t}} \\ 
&\leq 2\eta_0(4R + 2\|\bm{\theta}^\star\|_2+1) \cdot \frac{1}{16} = \frac{1}{2}R + \frac{1}{4}(2\|\bm{\theta}^\star\|_2+1).
\end{align*}
With a similar reasoning as the first step, the triangle inequality then indicates
\begin{align*}
\mathbb{P}(\mathcal{H}_{t}) \geq \mathbb{P}(\mathcal{H}_{t-1}) - \delta_t.
\end{align*}
Hence, the probability of $\mathcal{H}_{\infty}$ is bounded by
\begin{align*}
\mathbb{P}(\mathcal{H}_{\infty}) &\geq \mathbb{P}(\mathcal{H}_{t^\star}) -  \sum_{t=t^\star+1}^{\infty} \delta_t \\ 
&\geq \left(1-\frac{\delta}{3}\right)-\sum_{t=t^\star+1}^{\infty} 3\exp(-Ct^{\alpha}) \\ 
&\geq \left(1-\frac{\delta}{3}\right) - \int_{t^\star}^{\infty} 3\exp(-Cx^{\alpha}) \mathrm{d}x\\ 
&\geq \left(1-\frac{\delta}{3}\right) - \frac{\delta}{3} = 1-\frac{2}{3}\delta,
\end{align*}
where we applied the definition of $t^{\star}$ at the last inequality.
\paragraph{Step 4: Refining the bound on $\|\bm{\Delta}_t\|_2$.} With $\mathbb{P}(\mathcal{H}_{\infty})$ being bounded, we now proceed to refining the bound on $\|\bm{\Delta}_t\|_2$ for any $t \in \mathbb{N}$. Specifically, we observe, due to Lemma \ref{lemma:R}, that for every $t \in [T]$,
\begin{align*}
\left\|\prod_{k=1}^t (\bm{I}-\eta_k \bm{A})\bm{\Delta}_0\right\|_2 \leq \exp\left(-\frac{\beta}{1-\alpha}t^{1-\alpha}\right)\|\bm{\Delta}_0\|_2 \leq \left(\frac{\alpha}{2e\beta}\right)^{\frac{\alpha}{2(1-\alpha)}}\|\bm{\theta}^\star\|_2 t^{-\frac{\alpha}{2}}.
\end{align*}
Meanwhile, the vector Azuma's inequality (Corollary \ref{cor:vector-Azuma}) implies with probability at least $1-\frac{\delta}{3T}$,
\begin{align*}
\left\|\sum_{i=1}^t \bm{R}_i^t \tilde{\bm{\zeta}}_i\right\|_2 \leq 2\eta_0 \left(\frac{2\alpha}{\beta}\right)^{\frac{\alpha}{2(1-\alpha)}} \sqrt{\frac{2}{2\alpha-1} \log \frac{9T}{\delta}}(4R + 2\|\bm{\theta}^\star\|_2+1)t^{-\frac{\alpha}{2}}.
\end{align*}
The bound \eqref{eq:delta-t-alpha-bound-iid} then follows by invoking the triangle inequality and a union bound, and plugging in the definition of $R$ and $\beta$. \qed

\subsubsection{Completing the proof of Theorem \ref{thm:bar-delta-t}}\label{app:proof-bar-delta-t-iid}
Similar to the proof of Theorem \ref{thm:delta-t}, we firstly expand $\bar{\bm{\Delta}}_T$ into three terms: a non-random term capturing the error introduced by the initialization $\bm{\theta}_0$, a martingale, and a sum of $i.i.d.$ vectors. Specifically, \eqref{eq:delta-t} implies
\begin{align*}
\bar{\bm{\Delta}}_T &= \frac{1}{T}\sum_{t=1}^T \left(\prod_{k=1}^t (\bm{I}-\eta_k \bm{A})\Delta_0 - \sum_{i=1}^{t} \bm{R}_{i}^t (\bm{A}_i\bm{\theta}^\star -\bm{b}_i) - \sum_{i=1}^{t} \bm{R}_{i}^t (\bm{A}_i - \bm{A})\bm{\Delta}_{i-1}\right) \\ 
&= \frac{1}{T}\sum_{t=1}^T\prod_{k=1}^t (\bm{I}-\eta_k \bm{A})\Delta_0 - \frac{1}{T}\sum_{t=1}^T\sum_{i=1}^{t} \bm{R}_{i}^t (\bm{A}_i\bm{\theta}^\star -\bm{b}_i) - \frac{1}{T}\sum_{t=1}^T\sum_{i=1}^{t}\bm{R}_{i}^t (\bm{A}_i - \bm{A})\bm{\Delta}_{i-1} \\ 
&= \frac{1}{T}\sum_{t=1}^T\prod_{k=1}^t (\bm{I}-\eta_k \bm{A})\Delta_0 - \frac{1}{T} \sum_{i=1}^T \sum_{t=i}^T \bm{R}_{i}^t (\bm{A}_i\bm{\theta}^\star -\bm{b}_i) - \frac{1}{T} \sum_{i=1}^T \sum_{t=i}^T \bm{R}_{i}^t (\bm{A}_i - \bm{A})\bm{\Delta}_{i-1},
\end{align*}
where we change the order of summation in the last equation. Hence, the definition of $\bm{R}_i^t$ \eqref{eq:defn-Rit} and $\bm{Q}_t$ \eqref{eq:defn-Qt} indicates that
\begin{align*}
&\sum_{t=1}^T\prod_{k=1}^t (\bm{I}-\eta_k \bm{A}) = \frac{\bm{Q}_0}{\eta_0}, \quad \text{and} \\ 
&\sum_{t=i}^T \bm{R}_{i}^t = \sum_{t=i}^T \left(\eta_i \prod_{k=i+1}^t (\bm{I}-\eta_k \bm{A})\right) = \eta_i \sum_{t=i}^T\prod_{k=i+1}^t (\bm{I}-\eta_k \bm{A}) = \bm{Q}_i.
\end{align*}
Consequently, $\bar{\bm{\Delta}}_T$ can be expressed as
\begin{align}\label{eq:bar-delta-t-decompose}
\bar{\bm{\Delta}}_T = \underset{I_1}{\underbrace{\frac{1}{T\eta_0} \bm{Q}_0 \bm{\Delta}_0}} - \underset{I_2}{\underbrace{\frac{1}{T} \sum_{i=1}^T \bm{Q}_i(\bm{A}_i \bm{\theta}^\star-\bm{b}_i)}} - \underset{I_3}{\underbrace{\frac{1}{T}\sum_{i=1}^T \bm{Q}_i (\bm{A}_i-\bm{A})\bm{\Delta}_{i-1}}}.
\end{align}
Here, $I_1$ is a non-random term that captures the error introduced by $\bm{\Delta}_0$, $I_2$ is the sum of $i.i.d.$ random vectors, and $I_3$ is a martingale. We address the bounds on these terms in the following paragraphs in order. Through this proof, we denote, for simplicity, $\beta$ as
\begin{align*}
\beta:= \frac{1-\gamma}{2}\lambda_0 \eta_0
\end{align*}
\paragraph{Bounding the norm of $I_1$.} The norm of $I_1$ is bounded by the universal control over $\|\bm{Q}_t\|$ as indicated by Lemma \ref{lemma:Q-bound}. Specifically,
\begin{align}\label{eq:bar-delta-t-I1-bound}
\left\|\frac{1}{T\eta_0} \bm{Q}_0 \bm{\Delta}_0\right\|_2 \leq \frac{1}{T\eta_0} \|\bm{Q}_0\| \|\bm{\Delta}_0\|\leq \frac{3}{T}\left(\frac{2}{\beta}\right)^{\frac{1}{1-\alpha}} \|\bm{\Delta}_0\|_2.
\end{align}
\paragraph{Bounding the norm of $I_2$.} The term $I_2$ is a sum of $i.i.d.$ random vectors, and can hence be bounded by invoking the Freedman's inequality on Hilbert spaces (Theorem \ref{thm:Hilbert-Freedman}). Specifically, define
\begin{align*}
&W_{\max} := \sum_{i=1}^T \mathbb{E}\left\|\frac{1}{T}\bm{Q}_i(\bm{A}_i\bm{\theta}^\star-\bm{b}_i)\right\|_2^2, \quad \text{and} \\ 
&B_{\max}:= \max_{1 \leq i \leq T} \sup \left\|\frac{1}{T}\bm{Q}_i(\bm{A}_i\bm{\theta}^\star-\bm{b}_i)\right\|_2.
\end{align*}
It can then be guaranteed with probability at least $1-\delta$ that
\begin{align*}
\left\|\frac{1}{T} \sum_{i=1}^T \bm{Q}_i(\bm{A}_i \bm{\theta}^\star-\bm{b}_i)\right\|_2 \leq 2\sqrt{W_{\max} \log \frac{2}{\delta}}+\frac{4}{3}B_{\max}\log \frac{2}{\delta}.
\end{align*}
Furthermore, the terms $W_{\max}$ and $B_{\max}$ can be controlled by
\begin{align*}
W_{\max} &= \frac{1}{T^2} \sum_{i=1}^T \mathbb{E} \|\bm{Q}_i (\bm{A}_i\bm{\theta}^\star-\bm{b}_i)\|_2^2\\
 &= \frac{1}{T^2} \sum_{i=1}^T\mathbb{E}\left[(\bm{A}_i\bm{\theta}^\star-\bm{b}_i)^\top \bm{Q}_i^\top \bm{Q}_i (\bm{A}_i\bm{\theta}^\star-\bm{b}_i)\right]\\ 
&= \frac{1}{T^2} \sum_{i=1}^T\mathbb{E} \left[\mathsf{Tr}\left(\bm{Q}_i (\bm{A}_i\bm{\theta}^\star-\bm{b}_i)(\bm{A}_i\bm{\theta}^\star-\bm{b}_i)^\top \bm{Q}_i^\top \right)\right]\\ 
&= \frac{1}{T^2} \sum_{i=1}^T \mathsf{Tr}\left(\bm{Q}_i \mathbb{E}\left[(\bm{A}_i\bm{\theta}^\star-\bm{b}_i)(\bm{A}_i\bm{\theta}^\star-\bm{b}_i)^\top\right]\bm{Q}_i^\top\right)\\ 
&= \frac{1}{T} \mathsf{Tr}(\bar{\bm{\Lambda}}_T); \\ 
\text{and} \qquad B_{\max}&\leq \frac{1}{T}\sup_{1 \leq i \leq T} \|\bm{Q}_i\| \sup \|\bm{A}_i \bm{\theta}^\star - \bm{b}_i\|_2 \leq \frac{3\eta_0}{T} \left(\frac{2}{\beta}\right)^{\frac{1}{1-\alpha}}(2\|\bm{\theta}^\star\|_2+1).
\end{align*}
Hence, the norm of $I_2$ is bounded by
\begin{align}\label{eq:bar-delta-t-I2-bound}
\left\|\frac{1}{T} \sum_{i=1}^T \bm{Q}_i(\bm{A}_i \bm{\theta}^\star-\bm{b}_i)\right\|_2 \leq 2\sqrt{\frac{\mathsf{Tr}(\bar{\bm{\Lambda}}_T)}{T}\log \frac{6}{\delta}} + \frac{4\eta_0}{T} \left(\frac{2}{\beta}\right)^{\frac{1}{1-\alpha}}(2\|\bm{\theta}^\star\|_2+1) \log \frac{6}{\delta}.
\end{align}
with probability at least $1-\frac{\delta}{3}$.

\paragraph{Bounding the norm of $I_3$.}
The term $I_3$ is a martingale, to which the matrix Azuma's inequality is applicable. Towards this end, we need the universal bound on $\{\|\bm{\Delta}_{i-1}\|_2\}_{1 \leq i \leq T}$, for which Theorem \ref{thm:delta-t} comes in handy. Specifically, under the condition \eqref{eq:delta-t-condition}, it can be guaranteed with probability at least $1-\frac{\delta}{3}$ that
\begin{align*}
\left\|\bm{\Delta}_t\right\|_2 \lesssim \eta_0\left(\frac{2\alpha}{\beta}\right)^{\frac{\alpha}{2(1-\alpha)}}\sqrt{\frac{2}{2\alpha-1}\log \frac{27T}{\delta}}(2\|\bm{\theta}^\star\|_2+1)t^{-\frac{\alpha}{2}}, \quad \forall t \in [T].
\end{align*}
For simplicity, we denote
\begin{align*}
&R = \eta_0\left(\frac{2\alpha}{\beta}\right)^{\frac{\alpha}{2(1-\alpha)}}\sqrt{\frac{2}{2\alpha-1}\log \frac{27T}{\delta}}(2\|\bm{\theta}^\star\|_2+1),  \\ 
&\mathcal{H}_t = \left\{\max_{1 \leq j \leq t} j^{\frac{\alpha}{2}}\|\bm{\Delta}_j\|_2 \leq R\right\},\quad \text{and}  \\ 
&\tilde{\bm{\Delta}}_t = \bm{\Delta}_t\mathds{1}(\mathcal{H}_t).
\end{align*}
Then by definition, the probability of $\mathcal{H}_t$ is bounded by
\begin{align}\label{eq:bar-delta-t-Ht-bound}
\mathbb{P}(\mathcal{H}_1) \geq \mathbb{P}(\mathcal{H}_2) \geq ... \geq \mathbb{P}(\mathcal{H}_T) \geq 1-\frac{\delta}{3}.
\end{align}
Meanwhile, the vector Azuma's inequality can be invoked to obtain that with probability at least $1-\frac{\delta}{3}$,
\begin{align*}
\left\|\frac{1}{T}\sum_{i=1}^T \bm{Q}_i (\bm{A}_i-\bm{A})\tilde{\bm{\Delta}}_{i-1}\right\|_2\leq 2\sqrt{2W_{\max}\log \frac{9}{\delta}},
\end{align*}
where
\begin{align*}
W_{\max} = \sup \frac{1}{T^2} \sum_{i=1}^T \|\bm{Q}_i(\bm{A}_i-\bm{A})\tilde{\bm{\Delta}}_{i-1}\|_2^2 
&\lesssim \frac{1}{T^2} 3\eta_0^2 \left(\frac{2}{\beta}\right)^{\frac{1}{1-\alpha}}\cdot 4R^2\sum_{i=1}^T i^{-\alpha} \\ 
& \asymp \frac{12\eta_0^2}{1-\alpha}\left(\frac{2}{\beta}\right)^{\frac{1}{1-\alpha}}R^2 T^{-1-\alpha}.
\end{align*}
In other words, it can be guaranteed with probability at least $1-\frac{\delta}{3}$ that
\begin{align}\label{eq:bar-delta-t-I3-bound}
\left\|\frac{1}{T}\sum_{i=1}^T \bm{Q}_i (\bm{A}_i-\bm{A})\tilde{\bm{\Delta}}_{i-1}\right\|_2 
&\lesssim \eta_0 \left(\frac{2}{\beta}\right)^{\frac{1}{2(1-\alpha)}} R \sqrt{\frac{1}{1-\alpha}\log \frac{9}{\delta}} \nonumber\\ 
&\lesssim \eta_0^2 \left(\frac{2}{\beta}\right)^{\frac{2+\alpha}{2(1-\alpha)}}(6\|\bm{\theta}^\star\|_2+1)\sqrt{\frac{1}{(1-\alpha)(2\alpha-1)}}T^{-\frac{\alpha+1}{2}}\log \frac{27T}{\delta}.
\end{align}
Combining \eqref{eq:bar-delta-t-decompose}, \eqref{eq:bar-delta-t-I1-bound}, \eqref{eq:bar-delta-t-I2-bound}, \eqref{eq:bar-delta-t-Ht-bound} and \eqref{eq:bar-delta-t-I3-bound} with the triangle inequality and a union bound argument, we obtain with probability $1-\delta$ that
\begin{align*}
\|\bar{\bm{\Delta}}_T\|_2 &\leq \frac{3}{T}\left(\frac{2}{\beta}\right)^{\frac{1}{1-\alpha}} \|\bm{\Delta}_0\|_2 \\ 
&+ 2\sqrt{\frac{\mathsf{Tr}(\bar{\bm{\Lambda}}_T)}{T}\log \frac{6}{\delta}} + \frac{4\eta_0}{T} \left(\frac{2}{\beta}\right)^{\frac{1}{1-\alpha}}(2\|\bm{\theta}^\star\|_2+1) \log \frac{6}{\delta} \\ 
&+ \eta_0^2 \left(\frac{2}{\beta}\right)^{\frac{2+\alpha}{2(1-\alpha)}}(6\|\bm{\theta}^\star\|_2+1)\sqrt{\frac{1}{(1-\alpha)(2\alpha-1)}}T^{-\frac{\alpha+1}{2}}\log \frac{27T}{\delta}.
\end{align*}
In the mean time, since the difference between $\bar{\bm{\Lambda}}_T$ and $\bm{\Lambda}^\star$ is bounded by
\begin{align*}
\mathsf{Tr}(\bar{\bm{\Lambda}}_T) - \mathsf{Tr}(\bm{\Lambda}^\star) \lesssim T^{\alpha-1}\mathsf{Tr}(\bm{X}(\bm{\Lambda}^\star)) \lesssim \frac{\mathsf{Tr}(\bm{\Lambda}^\star)}{\beta} T^{\alpha-1},
\end{align*}
Theorem \ref{thm:bar-delta-t} follows by taking
\begin{align}\label{eq:bar-delta-t-tilde-C}
\tilde{C} &=  \eta_0^2 \left(\frac{2}{\beta}\right)^{\frac{2+\alpha}{2(1-\alpha)}}(6\|\bm{\theta}^\star\|_2+1)\sqrt{\frac{1}{(1-\alpha)(2\alpha-1)}}.
\end{align}

\subsubsection{The necessity of the condition \eqref{eq:delta-t-condition}}\label{app:delta-t-condition}
An important fact to note about Theorem \ref{thm:bar-delta-t} is that it relies on choosing an appropriate initial stepsize $\eta_0$ dependent of $\delta$. The relationship is specified in Equation \eqref{eq:delta-t-condition}.
This constraint, albeit only related to $\delta$ in logarithm terms, prevents $\delta$ to converge to $0$ for any \emph{fixed} $\eta_0$. One might naturally wonder if such a constraint could be relaxed and the convergence of $\bar{\bm{\Delta}}_T$ can be guaranteed with high probability for any fixed $\eta_0$ and arbitrarily small $\delta$. The following proposition shows that the answer is \emph{negative}, so some form of constraint on $\delta$ is necessary for a Chernoff-style convergence guarantee.

\begin{proposition}\label{proposition:alpha-explode}
Consider the TD algorithm \eqref{eq:TD-update-all} with independent samples and decaying stepsizes $\eta_t = \eta_0t^{-\alpha}$, where $\eta_0$ is fixed and $\alpha \in (\frac{1}{2},1)$. There exists an MDP, such that for any bi-variate polynomial function $\mathsf{poly}(\cdot,\cdot)$, there exists $\delta \in (0,1)$ and $T \in \mathbb{N}$, such that 
\begin{align*}
\mathbb{P}\left(\|\bar{\bm{\Delta}}_T\|_2 \geq \mathsf{poly}\left(T,\log \frac{1}{\delta}\right)\right) \geq \delta.
\end{align*}
\end{proposition}
\noindent\emph{Proof:} Consider the following MDP, where $|\mathcal{S}|=3$, $\gamma = 0.9$, while the transition matrix $\bm{P}$ and the reward vector $\bm{r}$ are
\begin{align*}
\bm{P} = \begin{pmatrix}
0.1 & 0.1 & 0.8 \\ 
0.1 & 0.1 & 0.8 \\ 
0.1 & 0.1 & 0.8 
\end{pmatrix}, \quad \text{and} \quad
\bm{r} = \begin{pmatrix}
0.1 \\ 
0.1 \\ 
1
\end{pmatrix}.
\end{align*}
It is easy to verify that $\bm{\mu} = (0.1,0.1,0.8)^\top$. Furthermore, let $d=1$ \footnote{In this case, many vectors and matrices reduce to scalars; however, we keep the boldface notation for simplicity.}, $\bm{\phi}(1) = -0.5$,$\bm{\phi}(2) = -1$ and $\bm{\phi}(3) = 1$. Direct calculation then yields
\begin{align*}
\bm{A} = 0.54475, \quad \bm{b} = 0.785, \quad \text{and} \quad \bm{\theta}^\star = 1.441.
\end{align*}
For any $T \in \mathbb{N}$, use $\mathcal{E}_T$ to denote the event that $s_t=1$ and $s_t' = 2$ for all $t \in [T]$. In the case of $i.i.d.$ samples, clearly $\mathbb{P}(\mathcal{E}_T) = 0.01^T$. For any bi-variate polynomial function $\mathsf{poly}(\cdot,\cdot)$, let $\delta = 0.01^T$, then $\log \frac{1}{\delta} = T \log 100$. Therefore, $\mathsf{poly}(T,\log \frac{1}{\delta}) = \mathsf{poly}(T,T\log 100)$ reduces to a univariate polynomial function of $T$. It suffices to show that under the event $\mathcal{E}_T$, the norm of $\bar{\bm{\Delta}}_T$ grows faster than any polynomial function of $T$.

Under the event $\mathcal{E}_T$, direct calculation yields $\bm{A}_t = -0.2$ and $\bm{b}_t = -0.05$. Hence, the estimation error $\bm{\Delta}_t$ is featured by the iterative relation
\begin{align*}
\bm{\Delta}_t &= \bm{\Delta}_{t-1} -\eta_t(\bm{A}_t \bm{\Delta}_{t-1} + \bm{A}_t \bm{\theta}^\star - \bm{b}_t) \\ 
&= (1+0.2 \eta_t) \bm{\Delta}_{t-1} +(0.2 \bm{\theta}^\star - 0.05)\eta_t.
\end{align*}
It is easy to show by induction that, when $\bm{\theta}_0 = 0$ and hence $\bm{\Delta}_0 = -\bm{\theta}^\star$,
\begin{align*}
\bm{\Delta}_t &=  (0.2 \bm{\theta}^\star - 0.05)\sum_{i=1}^t \eta_i \prod_{k=i+1}^t (1+0.2 \eta_k) + \prod_{k=1}^t (1+0.2\eta_k) \bm{\Delta}_0 \\ 
&= 5(0.2 \bm{\theta}^\star - 0.05)\sum_{i=1}^t \left(\prod_{k=i}^t - \prod_{k=i+1}^t \right)(1+0.2\eta_k)+ \prod_{k=1}^t (1+0.2\eta_k) (-\bm{\theta}^\star) \\
&= 5(0.2 \bm{\theta}^\star - 0.05)\left(\prod_{k=1}^t (1+0.2\eta_k) - 1\right)- \prod_{k=1}^t (1+0.2\eta_k) \bm{\theta}^\star\\ 
&= -0.25\prod_{k=1}^t (1+0.2\eta_k) + (0.25 - \bm{\theta}^\star) \\ 
&= -0.25\prod_{k=1}^t (1+0.2\eta_k) - 1.191.
\end{align*}
Consequently, the averaged TD estimation error is featured by
\begin{align*}
\bar{\bm{\Delta}}_T = -1.191 -0.25 \frac{1}{T}\sum_{t=1}^T \prod_{k=1}^t (1+0.2\eta_k),
\end{align*}
and therefore
\begin{align*}
\left\|\bar{\bm{\Delta}}_T\right\|_2 = 1.191 +0.25 \frac{1}{T}\sum_{t=1}^T \prod_{k=1}^t (1+0.2\eta_k)
\end{align*}
We now consider the growth of $\left\|\bar{\bm{\Delta}}_T\right\|_2$ with polynomial-decaying stepsizes $\eta_t = \eta_0 t^{-\alpha}$, where $\alpha \in (\frac{1}{2},1)$.
In this scenario, the norm of $\bar{\bm{\Delta}}_T$ is bounded by 
\begin{align*}
\left\|\bar{\bm{\Delta}}_T\right\|_2& = 1.191 +0.25 \frac{1}{T}\sum_{t=1}^T \prod_{k=1}^t (1+0.2\eta_k) \\ 
&= 1.191 + \frac{0.25}{T} \cdot \sum_{t=1}^T \frac{1}{0.2\eta_t} \left(\prod_{k=1}^{t} - \prod_{k=1}^{t-1}\right)(1+0.2\eta_k) \\ 
&> 1.191 + \frac{0.25}{0.2 \eta_T T}\sum_{t=1}^T\left(\prod_{k=1}^{t} - \prod_{k=1}^{t-1}\right)(1+0.2\eta_k)\\ 
&= 1.191 + \frac{5}{4\eta_0} T^{-\alpha-1} \left(\prod_{k=1}^T (1+0.2 \eta_0 k^{-\alpha}) - 1\right).
\end{align*}
Here, since $k^{-\alpha} \leq 1$, it can be guaranteed that
\begin{align*}
\prod_{k=1}^T (1+0.2 \eta_0 k^{-\alpha}) &> \prod_{k=1}^T (1+0.2\eta_0)^{k^{-\alpha}} \\ 
&= \exp\left(\log(1+0.2\eta_0) \sum_{k=1}^T k^{-\alpha}\right)\\ 
&> \exp\left(\log(1+0.2\eta_0) \cdot \frac{T^{1-\alpha}}{1-\alpha}\right).
\end{align*}
Take $u = T^{1-\alpha}$, then the norm of $\bar{\bm{\Delta}}_T$ is bounded below by
\begin{align*}
\left\|\bar{\bm{\Delta}}_T\right\|_2 > 1.191 + \frac{5}{4\eta_0}u^{-\frac{\alpha+1}{1-\alpha}}\left[\exp\left(\frac{\log(1+0.2\eta_0)}{1-\alpha} u\right)-1\right],
\end{align*}
which grows exponentially with $u$ and hence faster than any polynomial function of $T$. This completes the proof of the proposition.
\qed

\subsubsection{Comparing Theorem \ref{thm:bar-delta-t} with previous results}
\label{sec:morecomparions}

Table \ref{table:sample-complexity} provides a comparison of the guarantees of  Theorem \ref{thm:bar-delta-t} withexisting high-probability bounds for TD estimator with linear function approximation.

A recent line of works by \cite{chen2024concentration} and \cite{khodadadian2025general} has established high-probability convergence guarantees for the \emph{tabular} TD learning algorithm. Notably, their results do \emph{not} require the stepsizes to depend on the probability tolerance level $\delta$. For example, when translated to our notation, Theorem 6.1 of \cite{khodadadian2025general} stated that in the tabular case, with the choice of stepsizes $\eta_t = \eta_0 (t+h)^{-\frac{1}{2}}$ where $\eta_0 > \sqrt{h}$, it can be guaranteed for all $\delta \in (0,1)$ that with probability $1-\delta$, the averaged TD(n) estimator $\bar{\bm{V}}_T$ satisfies

\begin{align*}
\|\bar{\bm{V}}_T - \bm{V}^\star\|_{\infty}^2 \lesssim \frac{1}{(1-\gamma)^2(1-\gamma^n)^2\lambda_0^2} \frac{\log(1/\delta) + |\mathcal{S}|}{T}.
\end{align*}

We remark that this result is not applicable to our problem for the following reason.
\begin{enumerate}
\item When the discount factor $\gamma > 0$, this bound does not naturally apply to the TD(0) algorithm, since the $(1-\gamma^n)^2$ term in the denominator would be equal to $0$ when $n = 0$. 
\item Crucially, this bound depends on the fact that in the \emph{tabular} case, as long as the reward is bounded, the (original) TD iterates are \emph{uniformly bounded a.s.}; Please refer to Appendix B.8 of \cite{khodadadian2025general}, where Assumption 5.2 is verified. This fact cannot be generalized to TD with linear function approximation, as we have illustrated in the proof of Proposition \ref{proposition:alpha-explode}.
\end{enumerate}

In fact, in the tabular case and under our setting of TD(0) with stepsizes $\eta_t = \eta_0 t^{-\alpha}, \alpha \in (\frac{1}{2},1)$, one can verify by induction that  $\|\bm{V}_t\|_{\infty} \leq \frac{1}{1-\gamma}$ and therefore $\|\bm{V}_t\|_{2} \leq \frac{\sqrt{|\mathcal{S}|}}{1-\gamma}$ for all $t > 0$ almost surely. As a result, the guarantees in Theorem \ref{thm:delta-t} and of Theorem \ref{thm:bar-delta-t} can be straightened, by removing the dependence of $\eta_0$ on $\delta$. 
Specifically,  Theorem \ref{thm:delta-t} can be amended as:

For any $\eta_0 \in (0,\frac{1}{2\lambda_{\Sigma}})$ and $\delta \in (0,1)$, it holds with probability at least $1-\delta$ that
\begin{align*}
 \|\bm{V}_t - \bm{V}^\star\|_2 &\lesssim \frac{\eta_0}{\sqrt{2\alpha-1}}\left(\frac{(1-\gamma) \lambda_0 \eta_0}{4\alpha}\right)^{-\frac{\alpha}{2(1-\alpha)}}\sqrt{\log \frac{9T}{\delta}}\frac{\sqrt{|\mathcal{S}|}}{1-\gamma}t^{-\frac{\alpha}{2}} 
\end{align*}
simultaneously for all $t \in [T]$.  

In turn,  Theorem \ref{thm:bar-delta-t} can also be modified as follows:

For any $\eta_0 \in (0,\frac{1}{2\lambda_{\Sigma}})$ and $\delta \in (0,1)$, it holds with probability at least $1-\delta$ that
\begin{align*}
\|\bm{V}_t - \bm{V}^\star\|_2 \lesssim \sqrt{\frac{\mathsf{Tr}(\bm{\Lambda}^\star)}{T}\log \frac{1}{\delta}} + o(T^{-\frac{1}{2}}\log\frac{1}{\delta}).
\end{align*}

\begin{table*}[!t]
\centering
\renewcommand{\arraystretch}{2.3}
\begin{tabular}{c|c|c|c} 
\toprule
paper & stepsize & convergence guarantee & constraints on $t,T,\delta$\\ 
\toprule
	\citet{dalal2018finite}   & $\eta_t = (t+1)^{-1}$ &$\|\bm{\Delta}_t\|_2 \leq O\left(t^{-\min\{\frac{1}{2},\frac{\lambda}{\lambda+1}\}}\right)$ \footnotemark & $t \geq \tilde{C}(\log \frac{1}{\delta})$\\ \hline

	\citet{li2023sharp}   &{$\eta_t = \eta$} &{$\|\bar{\bm{\Delta}}_T\|_2 \leq O\left(T^{-\frac{1}{2}}\log^{\frac{1}{2}}\frac{T}{\delta}\right)$} & $\eta \sqrt{\log \frac{T}{\delta}} \leq \tilde{C} $\\ \hline

	\citet{samsonov2023finitesample}  & $\eta_t = \eta$ & $\|\bar{\bm{\Delta}}_T\|_2 \leq O\left(T^{-\frac{1}{2}}\log \frac{T}{\delta}\right)$ & $\eta \log \frac{T}{\delta} \leq \tilde{C} $ \\ \hline
	\textbf{Theorem \ref{thm:bar-delta-t}} & $\eta_t = \eta_0 t^{-\alpha},\alpha \in \left(\frac{1}{2},1\right)$ & $\|\bar{\bm{\Delta}}_T\|_2 \leq O\left(T^{-\frac{1}{2}}\sqrt{\log \frac{1}{\delta}}\right)$ & $\eta_0 \sqrt{\log \frac{1}{\delta\eta_0}} \leq \tilde{C}$\\ 
\toprule
\end{tabular}
\caption{Comparisons between our results on TD convergence rate with prior work. The quantity $\tilde{C}$, whose value may change at each occurrence, may depend on problem-related parameters other than $T$ and $\delta$. The conditions required in Theorem \ref{thm:bar-delta-t} are weaker than those in \citet{samsonov2023finitesample} and \citet{li2023sharp} while delivering the same rates; see discussion in the paragraph below Theorem \ref{thm:bar-delta-t}.} 
\label{table:sample-complexity}
\end{table*}

%\footnotetext{$\lambda$ can take any positive value smaller than $\min_{1 \leq i \leq d}\{\mathsf{Re}(\lambda_i(\bm{A}))\}$.}

\subsection{Proof of Theorem \ref{thm:Berry-Esseen-combined}}\label{app:proof-Berry-Essseen-combined}

By triangle inequality, we firstly decompose the convex distance between $\sqrt{T}\bar{\bm{\Delta}}_T$ and its asymptotic distribution $\mathcal{N}(\bm{0},\bm{\Lambda}^\star)$ by
\begin{align}\label{eq:Berry-Esseen-decompose}
d_{\mathsf{C}}(\sqrt{T}\bar{\bm{\Delta}}_T,\mathcal{N}(\bm{0},\bm{\Lambda}^\star)) \leq d_{\mathsf{C}}(\sqrt{T}\bar{\bm{\Delta}}_T,\mathcal{N}(\bm{0},\overline{\bm{\Lambda}}_T)) + d_{\mathsf{C}}(\mathcal{N}(\bm{0},\overline{\bm{\Lambda}}_T),\mathcal{N}(\bm{0},\bm{\Lambda}^\star)).
\end{align}
Here, recall that $\bar{\bm{\Lambda}}_T$ is defined as in \eqref{eq:defn-barLambdaT}. The following two theorems bound the two terms on the right-hand-side of \eqref{eq:Berry-Esseen-decompose} respectively; Theorem \ref{thm:Berry-Esseen-combined} follows directly by combining them with \eqref{eq:Berry-Esseen-decompose}.

\begin{theorem}\label{thm:Berry-Esseen}
Consider the TD with Polyak-Ruppert averaging with independent samples and decaying stepsizes $\eta_t = \eta_0 t^{-\alpha}$ for $\alpha \in (\frac{1}{2},1)$ and $\eta_0< 1/2$. 
% Let $\bm{Q}_t$ and $\overline{\bm{\Lambda}}_T$ be defined as in \eqref{eq:defn-Qt} and \eqref{eq:defn-barLambdaT} respectively, 
Then, %The distributions of $\bar{\bm{\Delta}}_T$ and the $d$-dimensional standard Gaussian random variable $\bm{z}$ are related by 
\begin{align*}
d_{\mathsf{C}}(\sqrt{T}\bar{\bm{\Delta}}_T,\mathcal{N}(\bm{0},\overline{\bm{\Lambda}}_T))\lesssim \sqrt{\frac{\eta_0 }{(1-\alpha)(1-\gamma) \lambda_0}} \mathsf{Tr}(\bm{\Gamma})\|\bm{\Gamma}^{-1}\| T^{-\frac{\alpha}{2}}  + \tilde{C} \; T^{-\frac{1}{2}},
\end{align*}
where $\tilde{C}$ is a problem related quantity independent of $T$.  %see definition~\eqref{} \yuting{add equation number} in Appendix \ref{sec:proof-thm-Berry-Esseen}.
\end{theorem}

\begin{theorem}
\label{thm:asymptotic}
Let $\bm{\Lambda}^\star$ and  $\overline{\bm{\Lambda}}_T$ be defined as in \eqref{eq:defn-Lambdastar} and \eqref{eq:defn-barLambdaT}, respectively, with the stepsizes $\eta_t = \eta_0 t^{-\alpha}$, for all $t>0$, any $\alpha \in (\frac{1}{2},1)$ and $\eta_0 < 1/2$.
Then, %the distance between Gaussian distributions $\mathcal{N}(\bm{0},\overline{\bm{\Lambda}}_T)$ and $\mathcal{N}(\bm{0},\bm{\Lambda}^\star)$ is bounded from above by 
\begin{align*}
&d_{\mathsf{C}}(\mathcal{N}(\bm{0},\overline{\bm{\Lambda}}_T),\mathcal{N}(\bm{0},\bm{\Lambda}^\star)) \leq d_{\mathsf{TV}}(\mathcal{N}(\bm{0},\overline{\bm{\Lambda}}_T),\mathcal{N}(\bm{0},\bm{\Lambda}^\star))  \lesssim   \frac{\sqrt{d\mathsf{cond}(\bm{\Gamma})}}{(1-\gamma)\lambda_0\eta_0} T^{\alpha-1}  
+  \tilde{C}' \; T^{2\alpha-2},
\end{align*}
where $\tilde{C}'$ is a problem related quantity independent of $T$.
\end{theorem}

\subsubsection{Proof of Theorem \ref{thm:Berry-Esseen}}\label{sec:proof-thm-Berry-Esseen}
This proof is based on Theorem \ref{thm:Shao}, the Berry-Esseen bound for multivariate non-linear statistics. Throughout the proof, we will firstly define corresponding statistics, and then bound the terms on the right-hand-side of \eqref{eq:thm-Shao}.
\paragraph{Defining $\bm{W}$,$\bm{D}$ and other statistics.} 
Recall from \eqref{eq:bar-delta-t-decompose} that the averaged TD error can be represented as
\begin{align*}
\overline{\bm{\Delta}}_T = \frac{1}{T \eta_0} \bm{Q}_0 \bm{\Delta}_0 - \frac{1}{T} \sum_{i=1}^{T-1} \bm{Q}_i \bm{\zeta}_i,
\end{align*}
where $\bm{Q}_t$ is defined as in \eqref{eq:defn-Qt}. Therefore, the statistics $\bm{W}$ and $\bm{D}$ can be constructed by

\begin{align}\label{eq:define-W-D}
\sqrt{T} \bm{A}\overline{\bm{\Delta}}_T &= \frac{\bm{A}}{\sqrt{T} \eta_0}  \bm{Q}_0 \bm{\Delta}_0 - \frac{\bm{A}}{\sqrt{T}}  \sum_{i=1}^{T} \bm{Q}_i \bm{\zeta}_i \nonumber \\
&= \underset{\bm{W}}{\underbrace{\sum_{i=1}^{T} \bm{\xi}_i}}  + \underset{\bm{D}}{\underbrace{\left(\frac{\bm{A}}{\sqrt{T} \eta_0}  \bm{Q}_0 \Delta_0 + \frac{\bm{A}}{\sqrt{T}} \sum_{i=1}^{T} \bm{Q}_i \bm{\omega}_i\right)}},
\end{align}
in which we define $\bm{\xi}_i$ and $\bm{\omega}_i$ as
\begin{align}
&\bm{\xi}_i = \frac{1}{\sqrt{T}}\bm{A} \bm{Q}_i(\bm{A}_i \bm{\theta}^\star - \bm{b}_i); \label{eq:defn-xii}\\
&\bm{\omega}_i = (\bm{A}_i-\bm{A}) \Delta_{i-1}. \label{eq:defn-omegai}
\end{align}
Notice that since $\{(s_i,s_i')\}_{i=1}^T$ are independent, $\{(\bm{A}_i,\bm{b}_i)\}_{i=1}^T$ are also independent. By definition, it is easy to verify that $\mathbb{E}[\bm{\xi}_i] = 0$ and that
\begin{align*}
\sum_{i=1}^T \mathbb{E}[\bm{\xi}_i\bm{\xi}_i^\top]
&= \frac{1}{T}\bm{A} \left\{\sum_{i=1}^T \bm{Q}_i \mathbb{E}[ (\bm{A}_i \bm{\theta}^\star - \bm{b}_i) (\bm{A}_i \bm{\theta}^\star - \bm{b}_i)^\top] \bm{Q}_i^\top\right\} \bm{A}^\top \\
&= \bm{A} \left\{\frac{1}{T} \sum_{i=1}^T \bm{Q}_i \bm{\Gamma} \bm{Q}_i^\top\right\} \bm{A}^\top = \bm{A}\bar{\bm{\Lambda}}_T \bm{A}^\top.
\end{align*}
With the minimum eigenvalue bounded below in Lemma \ref{lemma:A-Lambda} by
\begin{align*}
\lambda_{\min}(\bm{A}\bar{\bm{\Lambda}}_T \bm{A}^\top) \geq \frac{1}{2\|\bm{\Gamma}^{-1}\|},
\end{align*}
Theorem \ref{thm:Shao} guarantees that
\begin{align}\label{eq:Shao}
d_{\mathsf{C}}(\bm{A}\overline{\bm{\Delta}}_T,(\bm{A}\bar{\bm{\Lambda}}_T \bm{A}^\top)^{\frac{1}{2}}\bm{z})
& \lesssim \underset{I_1}{\underbrace{\sqrt{d}\|\bm{\Gamma}^{-\frac{3}{2}}\| \cdot \sum_{i=1}^n \mathbb{E}\|\bm{\xi}_i\|_2^3}} + \underset{I_2}{\underbrace{\|\bm{\Gamma}^{-1}\|\mathbb{E}\{\|\bm{W}\|_2\Delta\}}} + \underset{I_3}{\underbrace{\|\bm{\Gamma}^{-1}\|\sum_{i=1}^n \mathbb{E}\{\|\bm{\xi}_i\|_2 |\Delta - \Delta^{(i)}|\}}},
\end{align}
where $\Delta = \|\bm{D}\|$ and $\Delta^{(i)}$ is any random variable independent of $(\bm{A}_i,\bm{b}_i)$. Now, the proof boils down to bounding the three terms on the right-hand-side of \eqref{eq:Shao}.

\paragraph{Bounding $I_1$.} We firstly observe that
\begin{align*}
\|\bm{\xi}_i\|_2^3 &\leq T^{-3/2}  \|\bm{AQ}_i\|^3 \|\bm{A}_i \bm{\theta}^\star - \bm{b}_i\|_2^3\\
&\lesssim T^{-3/2} \cdot \left(2 + \eta_0\left(\frac{1}{\beta}\right)^{\frac{1}{1-\alpha}}\Gamma\left(\frac{1}{1-\alpha}\right)i^{\alpha-1}\right)^3 \cdot \|\bm{A}_i \bm{\theta}^\star - \bm{b}_i\|_2^3,
\end{align*}
in which  $\|\bm{AQ}_i\|$ is bounded by Lemma~\ref{lemma:Q-bound}. Furthermore, since $\max_{s \in \mathcal{S}} \|\bm{\phi}(s)\|_2 \leq 1$ and $\max_{s \in \mathcal{S}} r(s) \leq 1$, it is easy to verify that
\begin{align*}
\|\bm{A}_i \bm{\theta}^\star - \bm{b}_i\|_2 < 2\|\bm{\theta}^\star\|_2 + 1.
\end{align*}
Therefore, the term $I_1$ is bounded by 
\begin{align}\label{eq:bound-Shao-1}
\sqrt{d} \sum_{i=1}^T \mathbb{E} \|\bm{\xi}_i\|_2^3 
\lesssim \sqrt{d} \cdot   \left(\|\bm{\Gamma}^{ -1}\|\right)^{3/2}  \left(2\|\bm{\theta}^\star\|_2 + 1\right)^3   \left(\eta_0\left(\frac{1}{\beta}\right)^{\frac{1}{1-\alpha}}\Gamma\left(\frac{1}{1-\alpha}\right)\right)^3 T^{-\frac{1}{2}} = O(T^{-\frac{1}{2}}).
%&\leq  8 \sqrt{d} \left(\|\bm{\Lambda}_T^{\star -1}\|\right)^{3/2} \lambda_0^{-3}(1-\gamma)^{-3} \left(2\|\bm{\theta}^\star\|_2 + 1\right)^3 T^{-\frac{1}{2}} + \sqrt{d}o(T^{-\frac{1}{2}}).
\end{align}

\paragraph{Bounding $I_2$.} First of all, we define $\Delta = \|\bm{D}\|_2$. By Cauchy-Schwartz inequality, the second term of \eqref{eq:Shao} is upper bounded by 
\begin{align*}
\mathbb{E}\left\{\|\bm{W}\|_2\Delta\right\} \leq \sqrt{\mathbb{E} \|\bm{W}\|_2^2 \mathbb{E}\Delta^2}.
\end{align*}
Regarding $\mathbb{E} \|\bm{W}\|_2^2$, notice that
\begin{align*}
\mathbb{E} \|\bm{W}\|_2^2&=\mathbb{E}[\mathsf{Tr}(\bm{W}\bm{W}^\top)] 
= \mathsf{Tr}(\bm{A}\bar{\bm{\Lambda}}_T \bm{A}^\top) 
= \mathsf{Tr}(\bm{\Gamma}) + O(T^{\alpha-1}),
\end{align*}
where the last identity is a consequence of Theorem \ref{thm:Lambda}. In detail, 
\begin{align*}
\bm{A}\bar{\bm{\Lambda}}_T \bm{A}^\top - \bm{\Gamma} = \bm{A}(\bar{\bm{\Lambda}}_T - \bm{\Lambda}^\star)\bm{A}^\top  = T^{\alpha-1} \bm{X}(\bm{\Lambda}^\star)\bm{A}^\top + \bm{O}(T^{2\alpha-2}) = \bm{O}(T^{\alpha-1}),
\end{align*}
where $\bm{X}(\bm{\Lambda}^\star)$ is defined as in \eqref{eq:defn-X-Lambda}.

%\textcolor{violet}{Give details about the last step.}\weichen{checked.}

As for the expectation of $|\Delta|^2$, observe that
\begin{align*}
\mathbb{E}\Delta^2 &\leq 2\mathbb{E}\left\|\frac{1}{\sqrt{T} \eta_0} \bm{AQ}_0 \bm{\Delta}_0\right\|_2^2 + 2\mathbb{E} \left\|\frac{1}{\sqrt{T}} \sum_{i=1}^T \bm{AQ}_i\bm{\omega}_i\right\|_2^2 \\
&\leq \frac{2}{T\eta_0^2}  \|\bm{AQ}_0\|^2 \|\bm{\Delta}_0\|_2^2 + \frac{2}{T} \sum_{i=1}^T \|\bm{AQ}_i\|^2 \mathbb{E}\|\bm{\omega}_i\|_2^2\\
&\lesssim \frac{1}{T\eta_0^2}  \cdot \left(\frac{2}{(1-\gamma)\lambda_0\eta_0}\right)^{\frac{2}{1-\alpha}} \|\bm{\Delta}_0\|_2^2 \\ 
&+ \frac{1}{T} \sum_{i=1}^T (2+O(i^{\alpha-1}))^2 \mathbb{E} \|(\bm{A}_i-\bm{A})\|_2^2 \mathbb{E}\|\bm{\Delta}_{i-1}\|_2^2 \\
&\lesssim \frac{1}{T}  \sum_{i=1}^T  \mathbb{E}\|\bm{\Delta}_{i-1}\|_2^2 + O\left(\frac{1}{T}\right).
\end{align*}
The first line is the inequality $\|\bm{x} +\bm{y}\|_2^2 \leq 2\|\bm{x}\|_2^2 + 2\|\bm{y}\|_2^2$, for any vectors $\bm{x}$ and $\bm{y}$; in the third line, we applied the upper bounds for $\|\bm{AQ}_t\|$ established in Lemma \ref{lemma:Q-bound}. For the term $\mathbb{E}\|\bm{\Delta}_{i-1}\|_2^2$, the following lemma comes in handy.
\begin{lemma}\label{lemma:E-delta}
Under the TD iteration \eqref{eq:TD-update-all} with $i.i.d.$ samples and decaying stepsizes $\eta_t = \eta_0 t^{-\alpha}$ for $\alpha \in (\frac{1}{2},1)$ and $\eta_0 \leq \frac{1}{4}$, it can be guaranteed that
\begin{align*}
\mathbb{E}\|\bm{\Delta}_t\|_2^2 \leq \frac{2\eta_0\mathsf{Tr}(\bm{\Gamma})}{(1-\gamma) \lambda_0}   t^{-\alpha} + O(t^{-1}).
\end{align*}
holds for any $t >0$.
\end{lemma}
\noindent \emph{Proof}: See Appendix \ref{app:proof-lemma-E-delta}. \qed

As an immediate result of Lemma \ref{lemma:E-delta}, the last term is bounded by
\begin{align*}
\sum_{i=1}^T \mathbb{E}\|\bm{\Delta}_i\|_2^2 &\leq \sum_{i=1}^T \frac{2\eta_0}{(1-\gamma) \lambda_0} \mathsf{Tr}(\bm{\Gamma}) i^{-\alpha} + O(i^{-1})\\
&=\frac{2\eta_0}{(1-\alpha)(1-\gamma) \lambda_0}(2\|\bm{\theta}^\star\|_2+1)^2 T^{1-\alpha} + O(\log T).
\end{align*}
Therefore, $\mathbb{E}\Delta^2$ can be bounded by
\begin{align*}
\mathbb{E}\Delta^2 &\lesssim \frac{1}{T}  \sum_{i=1}^T  \mathbb{E}\|\bm{\Delta}_{i-1}\|_2^2 + O\left(\frac{1}{T}\right) \\ 
&\leq \frac{1}{T} \left(\frac{\eta_0}{(1-\alpha)(1-\gamma) \lambda_0}\mathsf{Tr}(\bm{\Gamma}) T^{1-\alpha} + O(\log T)\right) + O\left(\frac{1}{T}\right)\\ 
&= \frac{\eta_0\mathsf{Tr}(\bm{\Gamma})}{(1-\alpha)(1-\gamma) \lambda_0}T^{-\alpha} + O\left(\frac{\log T}{T}\right)
\end{align*}

So in conclusion, the second term of \eqref{eq:Shao} can be bounded by
\begin{align}\label{eq:bound-Shao-2}
\mathbb{E}\{\|W\|\Delta\} \leq \sqrt{\mathbb{E}\|W\|_2^2 \mathbb{E}\Delta^2} 
&\lesssim \sqrt{\frac{\eta_0 }{(1-\alpha)(1-\gamma) \lambda_0}} \mathsf{Tr}(\bm{\Gamma}) T^{-\frac{\alpha}{2}}
 + O\left(\sqrt{\frac{\log T}{T}}\right).
 \end{align}

\paragraph{Bounding $I_3$.} The third term of \eqref{eq:Shao} involves random variables $\{\Delta^{(i)}\}_{0 \leq i \leq T}$ such that $\Delta^{(i)}$ is independent of $(s_i,s_i')$. For this purpose, we define $\{(\check{s}_i,\check{s}_i')\}_{i=1}^n$ as independent copies of $\{(s_i,s_i')\}$ and let $\check{\bm{A}}_i,\check{\bm{b}}_i$ be generated by $(\check{s}_i,\check{s}_i')$. Consider the following iterations for any $i$:
\begin{align}\label{eq:define-theta-ji}
\bm{\theta}_j^{(i)} = \begin{cases}
\bm{\theta}_j, \text{ if }j < i;\\
\bm{\theta}_{j-1} - \eta_j(\check{\bm{A}}_j \bm{\theta}_{j-1}-\check{\bm{b}}_j), \text{ if }j = i;\\
\bm{\theta}_{j-1}^{(i)} - \eta_j (\bm{A}_j \bm{\theta}_{j-1}^{(i)} - \bm{b}_j), \text{ if }j > i.
\end{cases}
\end{align}
It is easy to verify that for all $j \in [T]$, $\bm{\theta}_j^{(i)}$ is independent of $(s_i,s_i')$. Therefore, with the definitions of
\begin{align}\label{eq:define-omega-ij}
\bm{\omega}_j^{(i)} = \begin{cases}
(\check{\bm{A}}_j - \bm{A}) (\bm{\theta}_{j-1}^{(i)} - \bm{\theta}^\star),\quad \text{ if } \quad j = i;\\
(\bm{A}_j - \bm{A})(\bm{\theta}_{j-1}^{(i)}- \bm{\theta}^\star), \quad \text{otherwise.}
\end{cases}
\end{align}
and
\begin{align*}
\Delta^{(i)} = \left\|\frac{1}{\sqrt{T} \eta_0}  \bm{AQ}_0 \bm{\Delta}_0- \frac{1}{\sqrt{T}}\sum_{j=1}^T \bm{AQ}_j\bm{\omega}_j^{(i)}\right\|_2,
\end{align*}

It can be guaranteed  that $\Delta^{(i)}$ is independent of $(s_i,s_i')$. The proof now boils down to bounding the difference between $\Delta$ and $\Delta^{(i)}$, for which the triangle inequality implies
\begin{align*}
|\Delta - \Delta^{(i)}| \leq \left\| \frac{1}{\sqrt{T}}  \sum_{j=1}^T \bm{AQ}_j (\bm{\omega}_j-\bm{\omega}_j^{(i)}) \right\| _2
&\leq \frac{1}{\sqrt{T}} \left\| \sum_{j=i}^T\bm{AQ}_j (\bm{\omega}_j-\bm{\omega}_j^{(i)}) \right\| _2.
\end{align*}
By squaring and taking expectations on both sides, we observe that
\begin{align}\label{eq:E-delta-deltai}
\mathbb{E}|\Delta- \Delta^{(i)}|^2
&\leq \frac{1}{T} \mathbb{E}\left\| \sum_{j=i}^T\bm{AQ}_j (\bm{\omega}_j-\bm{\omega}_j^{(i)}) \right\|_2^2 \nonumber \\ 
&= \frac{1}{T} \mathbb{E} \sum_{j=i}^T \left\|\bm{AQ}_j(\bm{\omega}_j-\bm{\omega}_j^{(i)})\right\|_2^2 + \frac{2}{T}  \mathbb{E} \left\{\sum_{i \leq j < j' < T} \left[\bm{AQ}_j\left(\bm{\omega}_j-\bm{\omega}_j^{(i)})\right)\right]^\top \left[\bm{AQ}_{j'}\left(\bm{\omega}_{j'}-\bm{\omega}_{j'}^{(i)}\right)\right]\right\}.
\end{align}
For the last term on the right-hand-side of \eqref{eq:E-delta-deltai}, recall from the definitions of $\bm{\omega}_j$ \eqref{eq:defn-omegai} and $\bm{\omega}_j^{(i)}$ \eqref{eq:define-omega-ij}, for any $j>i$,
\begin{align*}
\bm{\omega}_j-\bm{\omega}_j^{(i)} &= \left[(\bm{A}_j - \bm{A}) (\bm{\theta}_{j-1} - \bm{\theta}^\star)\right] - \left[(\bm{A}_j - \bm{A}) \left(\bm{\theta}_{j-1}^{(i)} - \bm{\theta}^\star\right)\right] \\ 
&= (\bm{A}_j - \bm{A}) \left(\bm{\theta}_{j-1} - \bm{\theta}_{j-1}^{(i)}\right)
\end{align*}
Hence, for $i \leq j < j' < T$, the law of total expectation implies 
\begin{align*}
&\mathbb{E}\left\{\left[\bm{AQ}_j\left(\bm{\omega}_j-\bm{\omega}_j^{(i)})\right)\right]^\top \left[\bm{AQ}_{j'}\left(\bm{\omega}_{j'}-\bm{\omega}_{j'}^{(i)}\right)\right]\right\} \\ 
&= \mathbb{E} \mathbb{E}_{j'-1}\left\{\left[\bm{AQ}_j\left(\bm{\omega}_j-\bm{\omega}_j^{(i)}\right)\right]^\top \left[\bm{AQ}_{j'}\left(\bm{\omega}_{j'}-\bm{\omega}_{j'}^{(i)}\right)\right]\right\} \\ 
&= \mathbb{E} \left\{\left[\bm{AQ}_j\left(\bm{\omega}_j-\bm{\omega}_j^{(i)}\right)\right]^\top \left[\bm{AQ}_{j'}\mathbb{E}_{j'-1}\left(\bm{\omega}_{j'}-\bm{\omega}_{j'}^{(i)}\right)\right]\right\}\\ 
&= \mathbb{E} \left\{\left[\bm{AQ}_j\left(\bm{\omega}_j-\bm{\omega}_j^{(i)}\right)\right]^\top \left[\bm{AQ}_{j'}\mathbb{E}_{j'-1} \left[(\bm{A}_{j'} - \bm{A}) \left(\bm{\theta}_{j'-1} - \bm{\theta}_{j'-1}^{(i)}\right)\right]\right]\right\} = 0,
\end{align*}
since $\mathbb{E}_{j'-1} [\bm{A}_{j'}- \bm{A}] = \bm{0}$. This effectively means that the last term on the right-hand-side of \eqref{eq:E-delta-deltai} is $0$. Therefore, the difference between $\Delta$ and $\Delta^{(i)}$ is bounded by
\begin{align}\label{eq:delta-deltai-decompose}
\mathbb{E}|\Delta- \Delta^{(i)}|^2 &\leq \frac{1}{T} \mathbb{E} \sum_{j=i}^T \left\|\bm{AQ}_j(\bm{\omega}_j-\bm{\omega}_j^{(i)})\right\|_2^2 \nonumber \\
&\leq \frac{1}{T} \sum_{j=i}^T \|\bm{AQ}_j\|^2 \mathbb{E} \left\|\bm{\omega}_j-\bm{\omega}_j^{(i)}\right\|_2^2 \nonumber \\ 
&\lesssim \frac{1}{T}  \cdot \sum_{j=i}^T (2+O(j^{\alpha-1}))\mathbb{E} \left\|\bm{\omega}_j-\bm{\omega}_j^{(i)}\right\|_2^2,
\end{align}
where in the last line we applied Lemma \ref{lemma:Q-bound} to bound the norm of $\bm{Q}_j$. The proof now boils down to bounding the expected difference between $\bm{\omega}_j$ and $\bm{\omega}_j^{(i)}$, for which we recall from definition that
\begin{align*}
\bm{\omega}_j - \bm{\omega}_j^{(i)} &= \begin{cases}
(\bm{A}_i - \check{\bm{A}}_i) (\bm{\theta}_{i-1} - \bm{\theta}^\star), \quad \text{for} \quad j = i;  \\
(\bm{A}_j - \bm{A}) \left(\bm{\theta}_{j-1} - \bm{\theta}_{j-1}^{(i)}\right), \quad \text{for} \quad j > i.
\end{cases}
\end{align*}
The norm of $(\bm{\theta}_{i-1} - \bm{\theta}^\star)$ can be bounded by Lemma \ref{lemma:E-delta}; for the norm of $\left(\bm{\theta}_{j-1} - \bm{\theta}_{j-1}^{(i)}\right)$, the following proposition comes in handy.
\begin{proposition}\label{proposition:theta-ji}
With $\bm{\theta}_j^{(i)}$ defined in \eqref{eq:define-theta-ji},it can be guaranteed that for every $i \in [T]$, 
\begin{subequations}
\begin{align}\label{eq:theta-ii}
\mathbb{E}\|\bm{\theta}_i-\bm{\theta}_i^{(i)}\|_2^2 \leq 2\eta_i^2 \mathsf{Tr}(\bm{\Gamma}) + 2\eta_i^2 \mathbb{E}\|\bm{\Delta}_{i-1}\|_2^2.
\end{align}
Furthermore, for every $j>i$, 
\begin{align}\label{eq:theta-ji}
\mathbb{E}\|\bm{\theta}_j-\bm{\theta}_j^{(i)}\|_2^2 &\leq \prod_{k=i+1}^j \left(1-\frac{1}{2}(1-\gamma)\lambda_0 \eta_k \right)\mathbb{E}\|\bm{\theta}_i-\bm{\theta}_i^{(i)}\|_2^2.
\end{align}
\end{subequations}
\end{proposition}  
\noindent \emph{Proof:} See appendix \ref{app:proof-proposition-theta-ji}. \qed

For $j=i$,  direct calculation yields 
\begin{align}\label{eq:omegai-omegaii}
\mathbb{E}\|\bm{\omega}_i-\bm{\omega}_i^{(i)}\|_2^2 &= \mathbb{E}\|(\bm{A}_i-\check{\bm{A}}_i)(\bm{\theta}_{i-1}-\bm{\theta}^\star)\|_2^2 \nonumber \\
&\lesssim \mathbb{E}\|\bm{\Delta}_{i-1}\|_2^2 ;
\end{align}
and similarly, for $j>i$, 
\begin{align}\label{eq:omegaj-omegaji}
\mathbb{E}\|\bm{\omega}_j-\bm{\omega}_j^{(i)}\|_2^2 &= \mathbb{E}\left\|(\bm{A}_j-\bm{A})\left(\bm{\theta}_{j-1}-\bm{\theta}_{j-1}^{(i)}\right)\right\|_2^2 \nonumber \\
&\lesssim \mathbb{E}\left\|\bm{\theta}_{j-1}-\bm{\theta}_{j-1}^{(i)}\right\|_2^2.
\end{align}
By combining \eqref{eq:delta-deltai-decompose}, \eqref{eq:omegai-omegaii},\eqref{eq:omegaj-omegaji} with \eqref{eq:theta-ji}, we obtain
\begin{align*}
\mathbb{E}|\Delta- \Delta^{(i)}|^2 &\leq\frac{1}{T} \left((2+O(i^{\alpha-1}))\mathbb{E}\|\bm{\omega}_i-\bm{\omega}_i^{(i)}\|_2^2  + \sum_{j=i+1}^T (2+O(j^{\alpha-1}))\mathbb{E}\|\bm{\omega}_j-\bm{\omega}_j^{(i)}\|_2^2  \right)\\
&\lesssim \frac{1}{T} \cdot (2+O(i^{\alpha-1})) \underset{I_1}{\underbrace{\mathbb{E}\|\bm{\omega}_i-\bm{\omega}_i^{(i)}\|_2^2}}\\ 
& +\frac{1}{T}\underset{I_2}{\underbrace{\sum_{j=i+1}^T (2+O(j^{\alpha-1})) \prod_{k=i+1}^j \left(1-\frac{1-\gamma}{2}\lambda_0 \eta_k\right) \mathbb{E}\|\bm{\theta}_i-\bm{\theta}_i^{(i)}\|_2^2}}.
\end{align*}
With $I_1$ bounded by \eqref{eq:omegai-omegaii}, the proof now boils down to bounding $I_2$. Notice that \eqref{eq:theta-ii} implies
\begin{align*}
&\sum_{j=i+1}^T  \prod_{k=i+1}^j \left(1-\frac{1-\gamma}{2}\lambda_0 \eta_k\right) \mathbb{E}\|\bm{\theta}_i-\bm{\theta}_i^{(i)}\|_2^2 \\ 
&\lesssim  \left[\mathsf{Tr}(\bm{\Gamma})+2\mathbb{E}\|\bm{\Delta}_{i-1}\|_2^2\right]\eta_0^2 i^{-2\alpha}\sum_{j=i+1}^T  \prod_{k=i+1}^j \left(1-\frac{1-\gamma}{2}\lambda_0 \eta_k\right),
\end{align*}
where we can invoke \eqref{eq:Q-uni-2} in Lemma \ref{lemma:Q-uni} to obtain 
\begin{align*}
i^{-2\alpha}\sum_{j=i+1}^T  \prod_{k=i+1}^j \left(1-\frac{1-\gamma}{2}\lambda_0 \eta_k\right) < \frac{2}{(1-\gamma) \lambda_0 \eta_0 } i^{-\alpha}+ O(i^{-1}).
\end{align*} 
Combining these results with Lemma \ref{lemma:E-delta}, we establish the following bound for $I_1+I_2$:
\begin{align*}
I_1 + I_2 \lesssim \frac{\eta_0}{(1-\gamma) \lambda_0}\mathsf{Tr}(\bm{\Gamma}) i^{-\alpha} + O(i^{-1}).
\end{align*}
Hence, 
\begin{align*}
\mathbb{E}|\Delta- \Delta^{(i)}|^2 \lesssim \frac{1}{T} \frac{\eta_0}{(1-\gamma) \lambda_0}\mathsf{Tr}(\bm{\Gamma}) i^{-\alpha} + O(i^{-1}).
\end{align*}

Therefore, by Cauchy-Schwartz inequality,
\begin{align}\label{eq:Shao-3-decompose}
\left(\mathbb{E}\sum_{i=1}^T \|\bm{\xi}_i\| |\Delta - \Delta^{(i)}|\right)^2 \leq \left(\sum_{i=1}^T \mathbb{E}\|\bm{\xi}_i\|_2^2 \right) \left(\sum_{i=1}^T \mathbb{E}|\Delta - \Delta^{(i)}|^2\right),
\end{align}
in which the two terms on the right-hand-side can be bounded respectively by 
\begin{align}
\label{eq:Shao-31}
\sum_{i=1}^T \mathbb{E}\|\bm{\xi}_i\|_2^2&= \sum_{i=1}^T \mathbb{E}\mathsf{Tr}[\bm{\xi}_i \bm{\xi}_i^\top] = \mathsf{Tr}(\bm{A}\bar{\bm{\Lambda}}_T \bm{A}^\top) = \mathsf{Tr}(\bm{\Gamma}) + O(T^{\alpha-1})
\end{align}
and
\begin{align}\label{eq:Shao-32}
\sum_{i=1}^T \mathbb{E}|\Delta - \Delta^{(i)}|^2
&\lesssim \frac{1}{T} \frac{\eta_0}{(1-\gamma) \lambda_0}\mathsf{Tr}(\bm{\Gamma}) \sum_{i=1}^T \left(i^{-\alpha} + O(i^{-1})\right) \notag \\
&= \frac{\eta_0\mathsf{Tr}(\bm{\Gamma})}{(1-\alpha)(1-\gamma) \lambda_0}T^{-\alpha} + O\left(\frac{\log T}{T}\right).
\end{align}
A direct combination of \eqref{eq:Shao-3-decompose}, \eqref{eq:Shao-31} and \eqref{eq:Shao-32} yields the following bound for the third term of \eqref{eq:Shao}:
\begin{align}\label{eq:bound-Shao-3}
\mathbb{E}\sum_{i=1}^T \|\bm{\xi}_i\| |\Delta - \Delta^{(i)}| 
\lesssim \sqrt{\frac{\eta_0 }{(1-\alpha)(1-\gamma) \lambda_0}} \mathsf{Tr}(\bm{\Gamma}) T^{-\frac{\alpha}{2}}+ O\left(\sqrt{\frac{\log T}{T}}\right).
\end{align}

\paragraph{Completing the proof.} A direct combination of \eqref{eq:Shao}, \eqref{eq:bound-Shao-1}, \eqref{eq:bound-Shao-2} and \eqref{eq:bound-Shao-3} yields
\begin{align*}
&\left|\mathbb{P}(\sqrt{T} \bm{A}\overline{\bm{\Delta}}_T \in \mathcal{A})-\mathbb{P}((\bm{A}\bar{\bm{\Lambda}}_T\bm{A}^\top)^{\frac{1}{2}}\bm{z} \in \mathcal{A})\right|\leq \sqrt{\frac{\eta_0 }{(1-\alpha)(1-\gamma) \lambda_0}} \|\bm{\Gamma}^{-1}\| \mathsf{Tr}(\bm{\Gamma}) T^{-\frac{\alpha}{2}}  + O\left(\sqrt{\frac{\log T}{T}}\right).
\end{align*}
Furthermore, for any convex set $\mathcal{A} \in \mathbb{R}^d$, the set
\begin{align*}
\mathcal{A}' = \{\bm{A}\bm{x}:\bm{x} \in \mathcal{A} \}
\end{align*}
must also be convex; by definition, $\mathcal{A}$ and $\mathcal{A}'$ are related by
\begin{align*} 
&\mathbb{P}(\sqrt{T} \overline{\bm{\Lambda}}_T^{-1/2}\overline{\bm{\Delta}}_T \in \mathcal{A}) = \mathbb{P}(\sqrt{T}\overline{\bm{\Delta}}_T \in \mathcal{A}'), \quad \text{and} \\ 
&\mathbb{P}((\bm{A}\bar{\bm{\Lambda}}_T\bm{A}^\top)^{\frac{1}{2}}\bm{z} \in \mathcal{A}) = \mathbb{P}(\overline{\bm{\Lambda}}_T^{1/2}\bm{z} \in \mathcal{A}').
\end{align*}
Therefore, for any convex set $\mathcal{A} \in \mathbb{R}^d$, it can be guaranteed that
\begin{align*}
&\left|\mathbb{P}(\sqrt{T}\overline{\bm{\Delta}}_T\in \mathcal{A})-\mathbb{P}(\overline{\bm{\Lambda}}_T^{1/2}\bm{z} \in \mathcal{A})\right| \leq \sqrt{\frac{\eta_0 }{(1-\alpha)(1-\gamma) \lambda_0}}\|\bm{\Gamma}^{-1}\| \mathsf{Tr}(\bm{\Gamma}) T^{-\frac{\alpha}{2}} + O\left(\sqrt{\frac{\log T}{T}}\right).
\end{align*}
Taking supremum over $\mathcal{A}$ completes the proof. \qed

\subsubsection{Proof of Theorem \ref{thm:asymptotic}}\label{app:proof-thm-asymptotic}
For any Borel set $\mathcal{A} \subset \mathbb{R}^d$, define $\mathcal{A}' = \{\bm{Ax}: \bm{x} \in \mathbb{R}^d\}$. By definition, when $\bm{z}$ is the $d$-dimensional standard Gaussian distribution,
\begin{align*}
\left|\mathbb{P}(\bar{\bm{\Lambda}}_T^{\frac{1}{2}}\bm{z} \in \mathcal{A}) - \mathbb{P}({\bm{\Lambda}}^{\star \frac{1}{2}}\bm{z} \in \mathcal{A})\right| = \left|\mathbb{P}(\bm{A}\bar{\bm{\Lambda}}_T^{\frac{1}{2}}\bm{z} \in \mathcal{A}') - \mathbb{P}(\bm{A}{\bm{\Lambda}}^{\star \frac{1}{2}}\bm{z} \in \mathcal{A}')\right|.
\end{align*}
Since $\bm{A}$ is non-singular, the map $\mathcal{A} \mapsto \mathcal{A}'$ is a bijection. Hence by taking supremum over $\mathcal{A}$ and $\mathcal{A}'$ on both sides of the equation, we obtain
\begin{align*}
d_{\mathsf{TV}}(\bar{\bm{\Lambda}}_T^{\frac{1}{2}}\bm{z},{\bm{\Lambda}}^{\star \frac{1}{2}}\bm{z}) = d_{\mathsf{TV}}(\bm{A}\bar{\bm{\Lambda}}_T^{\frac{1}{2}}\bm{z},\bm{A}{\bm{\Lambda}}^{\star \frac{1}{2}}\bm{z});
\end{align*}
In other words,
\begin{align*}
d_{\mathsf{TV}}(\mathcal{N}(\bm{0},\bar{\bm{\Lambda}}_T),\mathcal{N}(\bm{0},\bm{\Lambda}^\star)) = d_{\mathsf{TV}}(\mathcal{N}(\bm{0},\bm{A}\bar{\bm{\Lambda}}_T\bm{A}^\top),\mathcal{N}(\bm{0},\bm{\Gamma}));
\end{align*}
Therefore, the proof boils down to bounding the TV distance between $\mathcal{N}(\bm{0},\bm{A}\bar{\bm{\Lambda}}_T\bm{A}^\top)$ and $\mathcal{N}(\bm{0},\bm{\Gamma})$. Towards this end, Theorem \ref{thm:DMR} indicates that this TV distance converges at the same rate of
\begin{align*}
\left\|\bm{\Gamma}^{-\frac{1}{2}}\bm{A}\bar{\bm{\Lambda}}_T\bm{A}^\top\bm{\Gamma}^{-\frac{1}{2}} - \bm{I}_d\right\|_{\mathsf{F}} = \left\|\bm{\Gamma}^{-\frac{1}{2}}\bm{A}(\bar{\bm{\Lambda}}_T - \bm{\Lambda}^\star)\bm{A}^\top\bm{\Gamma}^{-\frac{1}{2}}\right\|_{\mathsf{F}}.
\end{align*}
Here, $\bar{\bm{\Lambda}}_T - \bm{\Lambda}^\star$ featured by Theorem \ref{thm:Lambda} as
\begin{align*}
\bar{\bm{\Lambda}}_T - \bm{\Lambda}^\star = T^{\alpha-1}\bm{X} + \bm{O}(T^{2\alpha-2}),
\end{align*}
where $\bm{X}$ satisfies the Lyapunov equation
\begin{align*}
\eta_0 (\bm{AX}+\bm{XA}^\top) = \bm{\Lambda}^\star;
\end{align*}
Therefore it is easy to verify that
\begin{align*}
\bm{\Gamma}^{-\frac{1}{2}}\bm{A}(\bar{\bm{\Lambda}}_T - \bm{\Lambda}^\star)\bm{A}^\top\bm{\Gamma}^{-\frac{1}{2}}=T^{\alpha-1}\bm{Y} + \bm{O}(T^{2\alpha-2}),
\end{align*}
in which
\begin{align*}
\eta_0(\bm{\Gamma}^{-\frac{1}{2}}\bm{A}\bm{\Gamma}^{\frac{1}{2}}\bm{Y} + \bm{Y}\bm{\Gamma}^{\frac{1}{2}}\bm{A}^\top\bm{\Gamma}^{-\frac{1}{2}}) = \bm{I}.
\end{align*}
Let $\lambda$ be any eigenvalue of $\bm{Y}$ and $\bm{x}$ be its corresponding eigenvector with $\|\bm{x}\|_2=1$. Then by definition,
\begin{align*}
1&=\eta_0 \bm{x}^\top (\bm{\Gamma}^{-\frac{1}{2}}\bm{A}\bm{\Gamma}^{\frac{1}{2}}\bm{Y} + \bm{Y}\bm{\Gamma}^{\frac{1}{2}}\bm{A}^\top\bm{\Gamma}^{-\frac{1}{2}}) \bm{x} \\ 
&= \eta_0 \lambda \bm{x}^\top (\bm{\Gamma}^{-\frac{1}{2}}\bm{A}\bm{\Gamma}^{\frac{1}{2}} + \bm{\Gamma}^{\frac{1}{2}}\bm{A}^\top\bm{\Gamma}^{-\frac{1}{2}})\bm{x} \\ 
&\geq \eta_0 \lambda \cdot \lambda_{\min}(\bm{\Gamma}^{-\frac{1}{2}}\bm{A}\bm{\Gamma}^{\frac{1}{2}} + \bm{\Gamma}^{\frac{1}{2}}\bm{A}^\top\bm{\Gamma}^{-\frac{1}{2}}) \\ 
&\geq \frac{2(1-\gamma)\lambda_0 \eta_0}{\mathsf{cond}(\bm{\Gamma}^{\frac{1}{2}})} \lambda.
\end{align*}
Therefore, the Frobenius norm of $\bm{Y}$ is bounded by
\begin{align*}
\|\bm{Y}\|_{\mathsf{F}} \leq \sqrt{d}\lambda_{\max}(\bm{Y}) \leq \frac{\sqrt{d\mathsf{cond}(\bm{\Gamma}^{\frac{1}{2}})}}{2(1-\gamma)\lambda_0\eta_0} T^{\alpha-1}.
\end{align*}
Hence the distance between $\mathcal{N}(\bm{0},\bar{\bm{\Lambda}}_T)$ and $\mathcal{N}(\bm{0},\bm{\Lambda}^\star)$ is bounded by
\begin{align*}
d_{\mathsf{C}}(\mathcal{N}(\bm{0},\bar{\bm{\Lambda}}_T),\mathcal{N}(\bm{0},\bm{\Lambda}^\star)) &\leq \|\bm{Y}\|_{\mathsf{F}}T^{\alpha-1} + \|\bm{O}(T^{2\alpha-2})\|_{\mathsf{F}} = \frac{\sqrt{d\mathsf{cond}(\bm{\Gamma})}}{2(1-\gamma)\lambda_0\eta_0} T^{\alpha-1} + O(T^{2\alpha-2}).
\end{align*}
\qed
\subsubsection{Tightness of Theorem \ref{thm:asymptotic}}\label{sec:tightness.thm:asymptotic} 

As is indicated by Lemma \ref{thm:Lambda}, the difference between $\bar{\bm{\Lambda}}_T$ and $\bm{\Lambda}^\star$ is featured by
\begin{align*}
\bar{\bm{\Lambda}}_T - \bm{\Lambda}^\star \geq \bm{O}(T^{\alpha-1})
\end{align*}
when $T$ is sufficiently large. Therefore, since $\bar{\bm{\Lambda}}_T \succ \bm{\Lambda}^\star$, Proposition \ref{prop:Gaussian} guarantees that
\begin{align*}
d_{\mathsf{C}}(\mathcal{N}(\bm{0},\bar{\bm{\Lambda}}_T),\mathcal{N}(\bm{0},\bm{\Lambda}^\star)) = d_{\mathsf{TV}}(\mathcal{N}(\bm{0},\bar{\bm{\Lambda}}_T),\mathcal{N}(\bm{0},\bm{\Lambda}^\star)).
\end{align*}
Meanwhile, Theorem \ref{thm:DMR} indicates the TV distance between $\mathcal{N}(\bm{0},\bar{\bm{\Lambda}}_T)$ and $\mathcal{N}(\bm{0},\bm{\Lambda}^\star))$ converges at the same rate, i.e.
\begin{align*}
d_{\mathsf{TV}}(\mathcal{N}(\bm{0},\bar{\bm{\Lambda}}_T),\mathcal{N}(\bm{0},\bm{\Lambda}^\star))&\approx \left\|\bm{\Lambda}^{\star -\frac{1}{2}}\bar{\bm{\Lambda}}_T\bm{\Lambda}^{\star -\frac{1}{2}}-\bm{I}\right\|_{\mathsf{F}} \\ 
&= \left\|\bm{\Lambda}^{\star -\frac{1}{2}}(\bar{\bm{\Lambda}}_T - \bm{\Lambda}^\star)\bm{\Lambda}^{\star -\frac{1}{2}}\right\|_{\mathsf{F}} = O(T^{\alpha-1}).
\end{align*}
Consequently, the $O(T^{\alpha-1})$ rate by which $\mathcal{N}(\bm{0},\bar{\bm{\Lambda}}_T)$ converges to $\mathcal{N}(\bm{0},\bm{\Lambda}^\star)$ cannot be further improved. As a direct consequence, when $\alpha \in (\frac{2}{3},1)$, since $\alpha-1 > -\frac{1}{3}> -\frac{\alpha}{2}$, it can be guaranteed through triangle inequality that
\begin{align*}
d_{\mathsf{C}}(\sqrt{T}\bar{\bm{\Delta}}_T,\mathcal{N}(\bm{0},\bm{\Lambda}^\star)) &\geq  d_{\mathsf{C}}(\mathcal{N}(\bm{0},\overline{\bm{\Lambda}}_T),\mathcal{N}(\bm{0},\bm{\Lambda}^\star))-d_{\mathsf{C}}(\sqrt{T}\bar{\bm{\Delta}}_T,\mathcal{N}(\bm{0},\overline{\bm{\Lambda}}_T)) \\ 
&\geq \Theta(T^{\alpha-1}) - O(T^{-\frac{\alpha}{2}}) \geq \Theta(T^{-\frac{1}{3}}).
\end{align*}

\subsection{Proof of Theorem \ref{thm:plug-in-iid}}\label{app:proof-plug-in-iid}
Applying the same reasoning as in the proof of Theorem \ref{thm:asymptotic},
\begin{align*}
d_{\mathsf{TV}}(\mathcal{N}(\bm{0},\hat{\bm{\Lambda}}_T),\mathcal{N}(\bm{0},\bm{\Lambda}^\star))&= d_{\mathsf{TV}}(\mathcal{N}(\bm{0},\bm{A}\hat{\bm{\Lambda}}_T\bm{A}^\top),\mathcal{N}(\bm{0},\bm{\Gamma})) \\
&\lesssim \|\bm{\Gamma}^{-\frac{1}{2}}\bm{A}(\hat{\bm{\Lambda}}_T - \bm{\Lambda}^\star)\bm{A}^\top \bm{\Gamma}^{-\frac{1}{2}}\|_{\mathsf{F}} \\ 
&\leq \|\bm{\Gamma}^{-1}\| \|\bm{A}(\hat{\bm{\Lambda}}_T - \bm{\Lambda}^\star)\bm{A}^\top\|_{\mathsf{F}}
\end{align*}

In order to bound the right-hand-side of the inequality above, we firstly present the difference between $\hat{\bm{\Lambda}}$ and $\bm{\Lambda}^\star$ as 
\begin{align}\label{eq:delta-Lambda-decompose}
\hat{\bm{\Lambda}}_T - \bm{\Lambda}^\star &= \bar{\bm{A}}_T^{-1}\hat{\bm{\Gamma}}_T \bar{\bm{A}}_T^{-\top} - \bm{A}^{-1} \bm{\Gamma}\bm{A}^{-\top} \nonumber \\ 
&= \bar{\bm{A}}_T^{-1} (\hat{\bm{\Gamma}}_T - \bm{\Gamma}) \bar{\bm{A}}_T^{-\top} + \left[\bar{\bm{A}}_T^{-1}\bm{\Gamma} \bar{\bm{A}}_T^{-\top} - \bm{A}^{-1} \bm{\Gamma} \bm{A}^{-\top}\right].
\end{align}
The proof now boils down to bounding the estimation error of $\bar{\bm{A}}^{-1}$ and $\hat{\bm{\Gamma}}$, for which the following lemmas come in handy.

\begin{lemma}\label{lemma:bar-A-inv}
Let $\bar{\bm{A}}_T$ be defined as in \eqref{eq:defn-hat-Gamma} with $\{(s_t,s_t')\}_{0 \leq t \leq T}$ being $i.i.d.$ samples. It can then be guaranteed with probability at least $1-\delta$ that
\begin{align*}
\|(\bar{\bm{A}}_T^{-1}-\bm{A}^{-1})\bm{A}\|_{\mathsf{F}} \leq \frac{4\sqrt{2}}{\lambda_0(1-\gamma)\sqrt{T}} \sqrt{\log \frac{3}{\delta}},
\end{align*}
with the proviso that
\begin{align}\label{eq:A-inv-condition}
T \geq \frac{128}{\lambda_0(1-\gamma)}\log \frac{3}{\delta}.
\end{align}
\end{lemma}
\noindent \emph{Proof}: See Appendix \ref{app:proof-lemma-bar-A-inv}. \qed
\begin{lemma}\label{lemma:Gamma}
Let $\bm{\Gamma}$ be defined as in \eqref{eq:defn-Lambdastar} and $\hat{\bm{\Gamma}}_T$ be defined as in \eqref{eq:defn-hat-Gamma}. Then the difference between $\hat{\bm{\Gamma}}_T$ and $\bm{\Gamma}$ can be bounded by
\begin{align*}
\left\|\hat{\bm{\Gamma}}_T - \bm{\Gamma}\right\|_{\mathsf{F}} \leq \frac{\sqrt{\mathsf{Tr}(\bm{\Gamma})}}{\lambda_0(1-\gamma)}(2\|\bm{\theta}^\star\|_2+1)\|\bm{\Gamma}^{-1}\| T^{-\frac{1}{3}} + o(T^{-\frac{1}{3}}),
\end{align*}
with probability at least $1-2T^{-\frac{1}{3}}-\frac{\mathsf{Tr}(\bm{\Gamma})}{\lambda_0(1-\gamma)}T^{-\frac{1}{2}}$.
\end{lemma}
% \begin{align*}
% \left\|\hat{\bm{\Gamma}}_T - \bm{\Gamma}\right\|_{\mathsf{F}} \leq \left(\frac{\sqrt{\mathsf{Tr}(\bm{\Gamma})}}{\lambda_0(1-\gamma)}+(2\|\bm{\theta}^\star\|_2+1)\right)(2\|\bm{\theta}^\star\|_2+1)\sqrt{\frac{\log T}{T}} + o\left(\sqrt{\frac{\log T}{T}}\right)
% \end{align*}
% with probability at least $1-2T^{-\frac{1}{2}}$.

\noindent \emph{Proof}: See Appendix \ref{app:proof-lemma-Gamma}. \qed

Based on Lemmas \ref{lemma:bar-A-inv} and \ref{lemma:Gamma}, we now consider the estimation error of $\hat{\bm{\Lambda}}_T$. By triangle inequality, \eqref{eq:delta-Lambda-decompose} implies that
\begin{align*}
\left\|\bm{A}(\hat{\bm{\Lambda}}_T - \bm{\Lambda}^\star)\bm{A}^\top \right\|_{\mathsf{F}} \leq \underset{I_1}{\underbrace{\|\bar{\bm{A}}_T^{-1}\bm{A}\|^2 \|\hat{\bm{\Gamma}}_T - \bm{\Gamma}\|_{\mathsf{F}}}} + \underset{I_2}{\underbrace{\left\|\bm{A}\bar{\bm{A}}_T^{-1}\bm{\Gamma} \bar{\bm{A}}_T^{-\top}\bm{A}^\top -  \bm{\Gamma} \right\|_{\mathsf{F}}}}.
\end{align*}
In what follows, we bound $I_1$ and $I_2$ in order.
\paragraph{Bounding $I_1$.} The norm of $\hat{\bm{\Gamma}}_T - \bm{\Gamma}$ is bounded by Lemma \ref{lemma:Gamma}, while the norm of $\bm{A}\bar{\bm{A}}^{-1}$ is bounded by the fact that 
\begin{align*}
\|\bm{A}\bar{\bm{A}}_T^{-1}\| \leq \|\bm{I}\| + \|\bm{A}(\bar{\bm{A}}_T^{-1} - \bm{A}^{-1})\| \leq 3
\end{align*}
with probability at least $1-\delta$, according to Lemma \ref{lemma:bar-A-inv} under condition \eqref{eq:A-inv-condition}. In combination, $I_1$ can be bounded by
\begin{align}\label{eq:bound-I1-Lambda}
I_1 \leq  \left(\frac{\sqrt{\mathsf{Tr}(\bm{\Gamma})}}{\lambda_0(1-\gamma)}+(2\|\bm{\theta}^\star\|_2+1)\right)(2\|\bm{\theta}^\star\|_2+1)\sqrt{\frac{\log T}{T}} + o\left(\sqrt{\frac{\log T}{T}}\right)
\end{align}
with probability at least $1-2T^{-\frac{1}{2}}$.

\paragraph{Bounding $I_2$.} The norm of $I_2$ is bounded by estimation error of $\bar{\bm{A}}^{-1}$. Namely, we further decompose $I_2$ by triangle inequality as
\begin{align*}
I_2 &\leq \|(\bm{A}\bar{\bm{A}}_T^{-1} - \bm{I}) \bm{\Gamma} \|_{\mathsf{F}} + \|\bm{\Gamma} (\bar{\bm{A}}_T^{-\top}\bm{A}^{\top} - \bm{I})\|_{\mathsf{F}} + \|(\bar{\bm{A}}_T^{-1} - \bm{A}^{-1})\bm{\Gamma} (\bar{\bm{A}}_T^{-\top} - \bm{A}^{-\top})\|_{\mathsf{F}}\\ 
&\leq 2 \|\bm{\Gamma}\| \|\bm{A}(\bar{\bm{A}}_T^{-1} - \bm{A}^{-1})\|_{\mathsf{F}} + \|\bm{\Gamma}\| \|\bm{A}(\bar{\bm{A}}_T^{-1} - \bm{A}^{-1})\|_{\mathsf{F}}^2,
\end{align*}
where we can apply Lemma \ref{lemma:bar-A-inv} to conclude that
\begin{align}\label{eq:bound-I2-Lambda}
I_2 \lesssim \frac{\|\bm{\Gamma}\|}{\lambda_0(1-\gamma)} \sqrt{\frac{\log T}{T}}.
\end{align}
with probability at least $1-T^{-\frac{1}{2}}$.
Theorem \ref{thm:plug-in-iid} follows by combining \eqref{eq:bound-I1-Lambda} and \eqref{eq:bound-I2-Lambda}.

\subsection{Proof of Theorem \ref{cor:confidence-region}}\label{app:proof-inference-iid}
We approach the theorem in two steps: first measuring the difference between the distribution of $\sqrt{T}\bar{\bm{\Delta}}_T$ and $\mathcal{N}(\bm{0},\hat{\bm{\Lambda}}_T)$, and then bounding the probability that $\bm{\theta}$ is within the region $\mathcal{C}_{\delta}$.

\paragraph{Step 1: Measuring the difference between the distribution of $\sqrt{T}\bar{\bm{\Delta}}_T$ and $\mathcal{N}(\bm{0},\hat{\bm{\Lambda}}_T)$.} We firstly observe that by triangle inequality, for any convex set $\mathcal{A} \subset \mathbb{R}^d$ and a $d$-dimensional standard Gaussian random variable $\bm{z}$,
\begin{align*}
\left|\mathbb{P}(\sqrt{T}\bar{\bm{\Delta}}_T \in \mathcal{A}) - \mathbb{P}(\hat{\bm{\Lambda}}_T^{\frac{1}{2}}\bm{z} \in \mathcal{A}) \right| 
&\leq \underset{I_1}{\underbrace{\left|\mathbb{P}(\sqrt{T}\bar{\bm{\Delta}}_T \in \mathcal{A}) - \mathbb{P}(\bar{\bm{\Lambda}}_T^{\frac{1}{2}}\bm{z} \in \mathcal{A}) \right|}} \\ 
&\quad+ \underset{I_2}{\underbrace{\left|\mathbb{P}(\bar{\bm{\Lambda}}_T^{\frac{1}{2}}\bm{z} \in \mathcal{A}) - \mathbb{P}({\bm{\Lambda}}^{\star \frac{1}{2}}\bm{z} \in \mathcal{A}) \right|}} + \underset{I_3}{\underbrace{\left|\mathbb{P}({\bm{\Lambda}}^{\star \frac{1}{2}}\bm{z} \in \mathcal{A}) - \mathbb{P}(\hat{\bm{\Lambda}}_T^{\frac{1}{2}}\bm{z} \in \mathcal{A}) \right|}}.
\end{align*}
Here, $I_1$ and $I_2$ measures the difference between the non-asymptotic and asymptotic distributions of $\bar{\bm{\Delta}}_T$, and are bounded by Theorems \ref{thm:Berry-Esseen} and \ref{thm:asymptotic} respectively. Specifically, 
\begin{align}\label{eq:plugin-I1-I2-bound}
&\left|\mathbb{P}(\sqrt{T}\bar{\bm{\Delta}}_T \in \mathcal{A}) - \mathbb{P}({\bm{\Lambda}}^{\star \frac{1}{2}}\bm{z} \in \mathcal{A}) \right|\nonumber \\ 
&\lesssim \sqrt{\frac{\eta_0 }{(1-\alpha)(1-\gamma) \lambda_0}} \mathsf{Tr}(\bm{\Gamma})\|\bm{\Gamma}^{-1}\| T^{-\frac{\alpha}{2}}+ \frac{\sqrt{d\mathsf{cond}(\bm{\Gamma})}}{2(1-\gamma)\lambda_0\eta_0} T^{\alpha-1} + O(T^{2\alpha-2}+T^{-\frac{1}{2}}).
\end{align} 
where $\bm{z}$ is a $d$-dimensional standard Gaussian random variable independent of the sample trajectory $\{(s_t,s_t')\}_{t \geq 0}$. Hence, the proof boils down to bounding $I_3$, which measures between the difference the centered Gaussian distribution with variance matrices $\hat{\bm{\Lambda}}_T$ and $\bm{\Lambda}^\star$. Towards this end, we use $\mathcal{E}$ to denote the event of 
\begin{align*}
&d_{\mathsf{TV}}(\mathcal{N}(\bm{0},\hat{\bm{\Lambda}}_T),\mathcal{N}(\bm{0},\bm{\Lambda}^\star))\\ 
&\lesssim \left(\frac{\sqrt{\mathsf{Tr}(\bm{\Gamma})}}{\lambda_0(1-\gamma)}+(2\|\bm{\theta}^\star\|_2+1)\right)(2\|\bm{\theta}^\star\|_2+1)\|\bm{\Gamma}^{-1}\|\sqrt{\frac{\log T}{T}} + o\left(\sqrt{\frac{\log T}{T}}\right).
\end{align*}
it has been illustrated in Theorem \ref{thm:plug-in-iid} that
\begin{align*}
\mathbb{P}(\mathcal{E}) \geq 1-3T^{-\frac{1}{2}};
\end{align*}
meanwhile by definition,
\begin{align*}
&\left|\mathbb{P}\left({\bm{\Lambda}}^{\star \frac{1}{2}}\bm{z} \in \mathcal{A}\right) - \mathbb{P}\left(\hat{\bm{\Lambda}}_T^{\frac{1}{2}}\bm{z} \in \mathcal{A} \mid \mathcal{E}\right) \right| \\ 
&\lesssim \left(\frac{\sqrt{\mathsf{Tr}(\bm{\Gamma})}}{\lambda_0(1-\gamma)}+(2\|\bm{\theta}^\star\|_2+1)\right)(2\|\bm{\theta}^\star\|_2+1)\|\bm{\Gamma}^{-1}\|\sqrt{\frac{\log T}{T}} + o\left(\sqrt{\frac{\log T}{T}}\right).
\end{align*}
Therefore, since $\bm{z}$ is independent of the sample trajectory, the law of total probability implies
\begin{align*}
\left|\mathbb{P}\left(\hat{\bm{\Lambda}}_T^{\frac{1}{2}}\bm{z} \in \mathcal{A}\right) -  \mathbb{P}\left(\hat{\bm{\Lambda}}_T^{\frac{1}{2}}\bm{z} \in \mathcal{A} \mid \mathcal{E}\right)\right| 
&= \left|\mathbb{P}\left(\hat{\bm{\Lambda}}_T^{\frac{1}{2}}\bm{z} \in \mathcal{A} \mid \mathcal{E}\right) (\mathbb{P}(\mathcal{E})-1) + \mathbb{P}\left(\hat{\bm{\Lambda}}_T^{\frac{1}{2}}\bm{z} \in \mathcal{A} \mid \mathcal{E}^{\complement}\right) \mathbb{P}\left(\mathcal{E}^\complement\right)\right| \\ 
&= \mathbb{P}(\mathcal{E}^\complement) \cdot \left| \mathbb{P}\left(\hat{\bm{\Lambda}}_T^{\frac{1}{2}}\bm{z} \in \mathcal{A} \mid \mathcal{E}\right) - \mathbb{P}\left(\hat{\bm{\Lambda}}_T^{\frac{1}{2}}\bm{z} \in \mathcal{A} \mid \mathcal{E}^\complement\right)\right| \\ 
&\leq \mathbb{P}(\mathcal{E}^\complement) \leq 3T^{-\frac{1}{2}}.
\end{align*}
Consequently, the triangle inequality directly implies that 
\begin{align}\label{eq:plugin-I3-bound}
& \left|\mathbb{P}\left({\bm{\Lambda}}^{\star \frac{1}{2}}\bm{z} \in \mathcal{A}\right) -  \mathbb{P}\left(\hat{\bm{\Lambda}}_T^{\frac{1}{2}}\bm{z} \in \mathcal{A} \right)\right|\nonumber \\ 
&\leq \left|\mathbb{P}\left({\bm{\Lambda}}^{\star \frac{1}{2}}\bm{z} \in \mathcal{A}\right) -  \mathbb{P}\left(\hat{\bm{\Lambda}}_T^{\frac{1}{2}}\bm{z} \in \mathcal{A} \mid \mathcal{E}\right)\right|+ \left|\mathbb{P}\left(\hat{\bm{\Lambda}}_T^{\frac{1}{2}}\bm{z} \in \mathcal{A}\right) -  \mathbb{P}\left(\hat{\bm{\Lambda}}_T^{\frac{1}{2}}\bm{z} \in \mathcal{A} \mid \mathcal{E}\right)\right| \nonumber \\ 
&\leq \left(\frac{\sqrt{\mathsf{Tr}(\bm{\Gamma})}}{\lambda_0(1-\gamma)}+(2\|\bm{\theta}^\star\|_2+1)\right)(2\|\bm{\theta}^\star\|_2+1)\|\bm{\Gamma}^{-1}\|\sqrt{\frac{\log T}{T}} + o\left(\sqrt{\frac{\log T}{T}}\right) + 3T^{-\frac{1}{2}} \nonumber \\ 
&=\left(\frac{\sqrt{\mathsf{Tr}(\bm{\Gamma})}}{\lambda_0(1-\gamma)}+(2\|\bm{\theta}^\star\|_2+1)\right)(2\|\bm{\theta}^\star\|_2+1)\|\bm{\Gamma}^{-1}\|\sqrt{\frac{\log T}{T}} + o\left(\sqrt{\frac{\log T}{T}}\right).
\end{align}
Combining \eqref{eq:plugin-I1-I2-bound}, \eqref{eq:plugin-I3-bound} and taking the supremum of $\mathcal{A}$ over all convex sets, we obtain 
\begin{align}\label{eq:inference-bound}
d_{\mathsf{C}}\left(\sqrt{T}\bar{\bm{\Delta}}_T,\mathcal{N}(\bm{0},\hat{\bm{\Lambda}}_T)\right)
\lesssim \frac{\sqrt{d\mathsf{cond}(\bm{\Gamma})}}{2(1-\gamma)\lambda_0\eta_0} T^{-\frac{1}{3}}+  \frac{\sqrt{\mathsf{Tr}(\bm{\Gamma})}}{\lambda_0(1-\gamma)}(2\|\bm{\theta}^\star\|_2+1)\|\bm{\Gamma}^{-1}\| T^{-\frac{1}{3}}+ o(T^{-\frac{1}{3}}).
\end{align}

\paragraph{Step 2: Bounding $\mathbb{P}(\bm{\theta}^\star \in \mathcal{C}_{\delta})$.} 
Define $\mathcal{C}_{\delta}'$ as
\begin{align*}
\mathcal{C}_{\delta}':= \left\{y=\sqrt{T}(\bar{\bm{\theta}}_T - \bm{x}): \bm{x} \in \mathcal{C}_{\delta}\right\}. 
\end{align*}
Then since $\mathcal{C}_{\delta}$ is convex, it is easy to verify that $\mathcal{C}_{\delta}'$ is also convex. Meanwhile on one hand,
\begin{align*}
\mathbb{P}(\bm{\theta}^\star \in \mathcal{C}_{\delta}) = \mathbb{P}\left(\sqrt{T}(\bar{\bm{\theta}}_T - \bm{\theta}^\star) \in \mathcal{C}_{\delta}'\right); 
\end{align*}
On the other hand,
\begin{align*}
\mathbb{P}_{\bm{z} \sim \mathcal{N}(\bm{0},\bm{I}_d)} \left(\bar{\bm{\theta}}_T - \frac{\hat{\bm{\Lambda}}_T^{\frac{1}{2}}\bm{z}}{\sqrt{T}} \in \mathcal{C}_{\delta} \bigg| \bar{\bm{\theta}}_T,\hat{\bm{\Lambda}}_T \right) = \mathbb{P}_{\bm{z} \sim \mathcal{N}(\bm{0},\bm{I}_d)}\left(\hat{\bm{\Lambda}}_T^{\frac{1}{2}}\bm{z} \in \mathcal{C}_{\delta}'\right).
\end{align*}
As a direct consequence, \eqref{eq:inference-bound} and the triangle inequality indicate that
\begin{align*}
\mathbb{P}(\bm{\theta}^\star \in \mathcal{C}_{\delta}) &= \mathbb{P}\left(\sqrt{T}(\bar{\bm{\theta}}_T - \bm{\theta}^\star) \in \mathcal{C}_{\delta}'\right) \\ 
&\geq \mathbb{P}_{\bm{z} \sim \mathcal{N}(\bm{0},\bm{I}_d)}\left(\hat{\bm{\Lambda}}_T^{\frac{1}{2}}\bm{z} \in \mathcal{C}_{\delta}'\right) - d_{\mathsf{C}}\left(\sqrt{T}\bar{\bm{\Delta}}_T,\mathcal{N}(\bm{0},\hat{\bm{\Lambda}}_T)\right) \\ 
&= \mathbb{P}_{\bm{z} \sim \mathcal{N}(\bm{0},\bm{I}_d)} \left(\bar{\bm{\theta}}_T - \frac{\hat{\bm{\Lambda}}_T^{\frac{1}{2}}\bm{z}}{\sqrt{T}} \in \mathcal{C}_{\delta} \bigg| \bar{\bm{\theta}}_T ,\hat{\bm{\Lambda}}_T\right)- d_{\mathsf{C}}\left(\sqrt{T}\bar{\bm{\Delta}}_T,\mathcal{N}(\bm{0},\hat{\bm{\Lambda}}_T)\right) \\ 
&\geq 1-\delta - \frac{\sqrt{\mathsf{Tr}(\bm{\Gamma})}}{\lambda_0(1-\gamma)}(2\|\bm{\theta}^\star\|_2+1)\|\bm{\Gamma}^{-1}\| T^{-\frac{1}{3}}-\frac{\sqrt{d\mathsf{cond}(\bm{\Gamma})}}{2(1-\gamma)\lambda_0\eta_0} T^{-\frac{1}{3}}  - o(T^{-\frac{1}{3}}).
\end{align*}
\qed

\section{Proof of supportive lemmas and propositions}

\subsection{Proof of Lemma \ref{lemma:E-delta}}\label{app:proof-lemma-E-delta}
The lemma is proved by developing an iterative relation for the series $\{\mathbb{E}\|\bm{\Delta}_t\|_2^2\}$. By definition, 
\begin{align*}
\bm{\Delta}_t &= (\bm{\theta}_{t-1} - \bm{\theta}^\star) - \eta_t (\bm{A}_t \bm{\theta}_{t-1}-\bm{b}_t)\\
&= (\bm{\theta}_{t-1} - \bm{\theta}^\star) - \eta_t \bm{A}_t (\bm{\theta}_{t-1} - \bm{\theta}^\star)-\eta_t (\bm{A}_t\bm{\theta}^\star - \bm{b}_t)\\
&=(\bm{I}-\eta_t \bm{A}_t) \bm{\Delta}_{t-1} - \eta_t (\bm{A}_t \bm{\theta}^\star -\bm{b}_t).
\end{align*}
Therefore, under $i.i.d.$ samples, we can relate $\mathbb{E}\|\bm{\Delta}_t\|_2^2$ with $\mathbb{E}\|\bm{\Delta}_{t-1}\|_2^2$ by 
\begin{align*}
\mathbb{E}\|\bm{\Delta}_t\|_2^2 &= \mathbb{E}\|(\bm{I}-\eta_t \bm{A}_t)\bm{\Delta}_{t-1}\|_2^2 + 2\eta_t \mathbb{E}\langle (\bm{I}-\eta_t \bm{A}_t)\bm{\Delta}_{t-1}, \bm{A}_t\bm{\theta}^\star-\bm{b}_t \rangle + \eta_t^2 \mathbb{E}\|\bm{A}_t\bm{\theta}^\star-\bm{b}_t\|_2^2\\
&= \mathbb{E}\|(\bm{I}-\eta_t \bm{A}_t) \bm{\Delta}_{t-1}\|_2^2 + 2\eta_t^2 \mathbb{E}\langle \bm{A}_t\bm{\Delta}_{t-1},\bm{A}_t\bm{\theta}^\star-\bm{b}_t\rangle + \eta_t^2 \mathbb{E}\|\bm{A}_t\bm{\theta}^\star-\bm{b}_t\|_2^2\\
&= \mathbb{E}\|(\bm{I}-\eta_t \bm{A}_t) \bm{\Delta}_{t-1}\|_2^2 + \eta_t^2 \left(\mathbb{E}\|\bm{A}_t\bm{\Delta}_{t-1}\|_2^2 + \mathbb{E}\|\bm{A}_t\bm{\theta}^\star-\bm{b}_t\|_2^2\right)\\
&+ \eta_t^2 \mathbb{E}\|\bm{A}_t\bm{\theta}^\star-\bm{b}_t\|_2^2\\
&= \|\bm{\Delta}_{t-1}\|_2^2 -\eta_t \bm{\Delta}_{t-1}^\top (\bm{A} + \bm{A}^\top) \bm{\Delta}_{t-1}\\
&+ 2\eta_t^2 \bm{\Delta}_{t-1}^\top \mathbb{E}[\bm{A}_t^\top \bm{A}_t]\bm{\Delta}_{t-1} + \frac{3}{2} \eta_t^2  \mathbb{E}\|\bm{A}_t\bm{\theta}^\star-\bm{b}_t\|_2^2\\
&\leq \|\bm{\Delta}_{t-1}\|_2^2 -(\eta_t -2\eta_t^2)\bm{\Delta}_{t-1}^\top (\bm{A} + \bm{A}^\top) \bm{\Delta}_{t-1} + 2\eta_t^2  \mathbb{E}\|\bm{A}_t\bm{\theta}^\star-\bm{b}_t\|_2^2.
\end{align*}
Notice that we applied Lemma \ref{lemma:A} in the last inequality. Now, since the stepsize $\eta_t$ satisfies $\eta_t \leq \eta_0 \leq \frac{1}{4}$, Lemma \ref{lemma:A} implies
\begin{align*}
\mathbb{E}\|\bm{\Delta}_t\|_2^2 &\leq \mathbb{E}\|\bm{\Delta}_{t-1}\|_2^2 - \frac{1}{2}\eta_t \mathbb{E}[(\bm{\theta}_{t-1}-\bm{\theta}^\star)^\top (\bm{A} + \bm{A}^\top) (\bm{\theta}_{t-1}-\bm{\theta}^\star) ] + 2 \eta_t^2  \mathbb{E}\|\bm{A}_t\bm{\theta}^\star-\bm{b}_t\|_2^2\\
&< \left(1-(1-\gamma)\lambda_0 \eta_t \right)\mathbb{E}\|\bm{\Delta}_{t-1}\|_2^2+ 2 \eta_t^2 \mathsf{Tr}(\bm{\Gamma}).
\end{align*}
Therefore by induction, $\mathbb{E}\|\bm{\Delta}_t\|_2^2$ is bounded by
\begin{align}\label{eq:E-deltat-iid-bound}
\mathbb{E}\|\bm{\Delta}_t\|_2^2 < \prod_{i=1}^t \left(1-(1-\gamma)\lambda_0 \eta_i \right)\|\bm{\Delta}_0\|_2^2 + 2\sum_{i=1}^n \eta_i^2 \prod_{j=i+1}^n \left(1-(1-\gamma)\lambda_0 \eta_j \right) \mathsf{Tr}(\bm{\Gamma}).
\end{align}
The proof now boils down to bounding the two terms on the right hand side of \eqref{eq:E-deltat-iid-bound}. For the first term, we observe
\begin{align}\label{eq:E-deltat-iid-bound-1}
\prod_{i=1}^t \left(1-(1-\gamma)\lambda_0 \eta_i \right) &< \exp \left((1-\gamma)\lambda_0 \eta_0 \sum_{i=1}^t i^{-\alpha}\right) \nonumber \\ 
&\lesssim \exp \left(-\frac{1-\gamma}{(1-\alpha)}\lambda_0 \eta_0 t^{1-\alpha}\right) = o(t^{-1});
\end{align}
for the second term, we can apply Lemma \ref{lemma:R} with $\beta = (1-\gamma) \lambda_0 \eta_0$ to show that
\begin{align}\label{eq:E-deltat-iid-bound-2}
\sum_{i=1}^n \eta_i^2 \prod_{j=i+1}^n \left(1-(1-\gamma)\lambda_0 \eta_j \right) = \frac{\eta_0}{(1-\gamma) \lambda_0}t^{-\alpha} + O(t^{-1}).
\end{align}
The lemma follows immediately by plugging \eqref{eq:E-deltat-iid-bound-1} and \eqref{eq:E-deltat-iid-bound-2} into \eqref{eq:E-deltat-iid-bound}.

\subsection{Proof of Proposition \ref{proposition:theta-ji}}\label{app:proof-proposition-theta-ji}
We address the two properties in order.
\paragraph{Proof of Equation \eqref{eq:theta-ii}.}Recall from \eqref{eq:TD-update-all} and \eqref{eq:define-theta-ji} that for any $i \in [T]$, 
\begin{align*}
&\bm{\theta}_i = \bm{\theta}_{i-1} -\eta_i (\bm{A}_i \bm{\theta}_{i-1}-\bm{b}_i);\quad \text{and}\\
&\bm{\theta}_i^{(i)} = \bm{\theta}_{i-1} - \eta_i (\check{\bm{A}}_i \bm{\theta}_{i-1} - \check{\bm{b}}_i).
\end{align*}
Therefore, the difference between $\bm{\theta}_i$ and $\bm{\theta}_i^{(i)}$ can be bounded by
\begin{align*}
\mathbb{E}\|\bm{\theta}_i - \bm{\theta}_{i}^{(i)}\|_2^2 &\leq \eta_i^2 \mathbb{E} \|(\bm{A}_i-\check{\bm{A}}_i)\bm{\theta}_{i-1} - (\bm{b}_i - \check{\bm{b}}_i)\|_2^2\\
&=  \eta_i^2 \mathbb{E} \|(\bm{A}_i - \bm{A})\bm{\theta}^\star + (\bm{A}_i-\bm{A}) \bm{\Delta}_{i-1} -(\bm{b}_i-\bm{b})\|_2^2\\
&\leq \eta_i^2 (2\mathbb{E}\|(\bm{A}_i - \bm{A})\bm{\Delta}_{i-1}\|_2^2 + 2\mathbb{E} \|(\bm{A}_i - \bm{A})\bm{\theta}^\star - (\bm{b}_i - \bm{b})\|_2^2) \\ 
&\leq 2\eta_i^2 \mathsf{Tr}(\bm{\Gamma}) + 2\eta_i^2 \mathbb{E}\|\bm{\Delta}_{i-1}\|_2^2.
\end{align*}

% Notice that the penultimate line follows from the fact that $(\check{s}_i,\check{s}_i')$ and $(s_i,s_i')$ have the same distribution. With $\mathbb{E}\|\bm{\Delta}_{i-1}\|_2$ bounded by Lemma \ref{lemma:E-delta}, we can conclude that
% \begin{align*}
% \mathbb{E}\|\bm{\theta}_i - \bm{\theta}_{i}^{(i)}\|_2^2 &\leq \widetilde{C}i^{-2\alpha} (\|\bm{\theta}^\star\|_2+1)^2[1+\lambda_0^{-\frac{\alpha}{1-\alpha}} i^{-\alpha}],
% \end{align*}
% with $\widetilde{C}$ being a constant depending on $\alpha,\eta_0$ and $\gamma$. 

\paragraph{Proof of Equation \eqref{eq:theta-ji}.} Recall from \eqref{eq:TD-update-all} and \eqref{eq:define-theta-ji} that
\begin{align*}
&\bm{\theta}_j = \bm{\theta}_{j-1}-\eta_j (\bm{A}_j \bm{\theta}_{j-1}-\bm{b}_j);\\
&\bm{\theta}_j^{(i)} = \bm{\theta}_{j-1}^{(i)} - \eta_j (\bm{A}_j \bm{\theta}_{j-1}^{(i)}-\bm{b}_j).
\end{align*}
Therefore, the difference between $\bm{\theta}_j$ and $\bm{\theta}_{j}^{(i)}$ can be expressed as
\begin{align*}
\mathbb{E}\|\bm{\theta}_j-\bm{\theta}_j^{(i)}\|_2^2 &= \mathbb{E}\|(\bm{I}-\eta_j \bm{A}_j) (\bm{\theta}_{j-1}-\bm{\theta}_{j-1}^{(i)})\|_2^2\\
&\leq \left(1-\frac{1}{2}(1-\gamma)\lambda_0 \eta_j \right)\mathbb{E}\|\bm{\theta}_{j-1}-\bm{\theta}_{j-1}^{(i)}\|_2^2,
\end{align*}
where the inequality follows from Lemma \ref{lemma:A}. By induction, we obtain
\begin{align*}
\mathbb{E}\|\bm{\theta}_j-\bm{\theta}_j^{(i)}\|_2^2 &\leq \prod_{k=i+1}^j \left(1-\frac{1}{2}(1-\gamma)\lambda_0 \eta_k \right)\mathbb{E}\|\bm{\theta}_i-\bm{\theta}_i^{(i)}\|_2^2.
\end{align*}

\subsection{Proof of Lemma \ref{lemma:bar-A-inv}}\label{app:proof-lemma-bar-A-inv}
We firstly observe that since $\|\bm{\phi}(s)\|_2 \leq 1$ for all $s \in \mathcal{S}$, it can be guaranteed that
\begin{align*}
\|\textbf{vec}(\bm{A}_t)\|_2^2 = \|\bm{A}_t\|_{\mathsf{F}}^2 
&= \mathsf{Tr}(\bm{A}_t^\top \bm{A}_t) \\ 
&=\mathsf{Tr}\left([\bm{\phi}(s_t) - \gamma \bm{\phi}(s_t')]\bm{\phi}(s_t)^\top \bm{\phi}(s_t)[\bm{\phi}(s_t) - \gamma \bm{\phi}(s_t')]^\top \right) \\ 
&\leq \mathsf{Tr}\left([\bm{\phi}(s_t) - \gamma \bm{\phi}(s_t')][\bm{\phi}(s_t) - \gamma \bm{\phi}(s_t')]^\top \right) < 4
\end{align*}
holds uniformly for all $0 \leq t \leq T$. Hence the vector Azuma's inequality \ref{cor:vector-Azuma} directly implies that
\begin{align*}
\|\bar{\bm{A}}_T - \bm{A}\|_{\mathsf{F}} &= \left\|\mathbf{vec}(\bar{\bm{A}}_T-\bm{A})\right\| \leq \frac{4\sqrt{2}}{\sqrt{T}} \sqrt{\log \frac{3}{\delta}}
\end{align*}
with probability at least $1-\delta$. In what follows, we aim to transform this error bound to the error bound for $\bar{\bm{A}}_T^{-1}$. Towards this end, we observe that condition \eqref{eq:A-inv-condition} guarantees with probability at least $1-\delta$, 
\begin{align*}
\left\|\bm{A}^{-1} (\bar{\bm{A}}_T- \bm{A})\right\| &\leq \|\bm{A}^{-1}\| \|\bar{\bm{A}}_T - \bm{A}\| \\ 
&\leq \lambda_0^{-1} (1-\gamma)^{-1} \frac{4\sqrt{2}}{\sqrt{T}} \sqrt{\log \frac{3}{\delta}} < \frac{1}{2}
\end{align*}
where the last inequality is guaranteed by the condition \eqref{eq:A-inv-condition}.
Hence, the condition of Lemma \ref{lemma:delta-inv} is satisfied with probability at least $1-\delta$ by taking $\bm{Y}=\bm{A}^{-1}\bar{\bm{A}}$ and $\bm{X} = \bm{I}$. Consequently, the difference between $\bar{\bm{A}}^{-1}\bm{A}$ and $\bm{I}$ is bounded by
\begin{align*}
\|(\bar{\bm{A}}_T^{-1} - \bm{A}^{-1})\bm{A}\|_{\mathsf{F}} 
\leq 2 \|(\bar{\bm{A}}_T - \bm{A})\bm{A}^{-1}\|_{\mathsf{F}} 
& \leq \frac{4\sqrt{2}}{\sqrt{T}}\sqrt{\log \frac{3}{\delta}}\|\bm{A}^{-1}\| \\ 
&\leq \frac{4\sqrt{2}}{\lambda_0(1-\gamma)\sqrt{T}} \sqrt{\log \frac{3}{\delta}}.
\end{align*}
\qed

\subsection{Proof of Lemma \ref{lemma:Gamma}}\label{app:proof-lemma-Gamma}
We firstly represent $\hat{\bm\Gamma}_T - \bm{\Gamma}$ as

\begin{align*}
\hat{\bm{\Gamma}}_T - \bm{\Gamma} &= \underset{I_1}{\underbrace{\sum_{t=1}^T w_{t,T}\left[(\bm{A}_t \bar{\bm{\theta}}_t - \bm{b}_t)(\bm{A}_t \bar{\bm{\theta}}_t - \bm{b}_t)^\top - (\bm{A}_t {\bm{\theta}}^\star - \bm{b}_t)(\bm{A}_t {\bm{\theta}^\star} - \bm{b}_t)^\top\right]}} \\ 
&+ \underset{I_2}{\underbrace{ \sum_{t=1}^T w_{t,T}\left[(\bm{A}_t {\bm{\theta}}^\star - \bm{b}_t)(\bm{A}_t {\bm{\theta}^\star} - \bm{b}_t)^\top] - \mathbb{E}[(\bm{A}_t {\bm{\theta}}^\star - \bm{b}_t)(\bm{A}_t {\bm{\theta}^\star} - \bm{b}_t)^\top\right]}},
\end{align*}

where $I_1$ is mainly controlled by the estimation error of $\bar{\bm{\theta}}_t$, and $I_2$ only involves the randomness of $\{(\bm{A}_t,\bm{b}_t)\}$. We now control the norms of these two terms in order.

\paragraph{Bounding the Frobenius norm of $I_1$.} By definition, $I_1$ can be further decomposed as
% \begin{align*}
% I_1 &= \frac{1}{T} \sum_{t=1}^T \left[(\bm{A}_t \bar{\bm{\theta}}_T - \bm{b}_t)(\bm{A}_t \bar{\bm{\theta}}_T - \bm{b}_t)^\top - (\bm{A}_t {\bm{\theta}}^\star - \bm{b}_t)(\bm{A}_t {\bm{\theta}^\star} - \bm{b}_t)^\top\right] \\ 
% &= \underset{I_{11}}{\underbrace{\frac{1}{T} \sum_{t=1}^T \left[(\bm{A}_t \bar{\bm{\Delta}}_T) (\bm{A}_t \bar{\bm{\Delta}}_T)^\top \right]}} + \underset{I_{12}}{\underbrace{\frac{1}{T} \sum_{t=1}^T \left[(\bm{A}_t \bar{\bm{\Delta}}_T)(\bm{A}_t \bm{\theta}^\star - \bm{b}_t)^\top + (\bm{A}_t \bm{\theta}^\star - \bm{b}_t) (\bm{A}_t \bar{\bm{\Delta}}_T)^\top \right]}}.
% \end{align*}
\begin{align*}
I_1 &= \sum_{t=1}^T w_{t,T} \left[(\bm{A}_t \bar{\bm{\theta}}_t - \bm{b}_t)(\bm{A}_t \bar{\bm{\theta}}_t - \bm{b}_t)^\top - (\bm{A}_t {\bm{\theta}}^\star - \bm{b}_t)(\bm{A}_t {\bm{\theta}^\star} - \bm{b}_t)^\top\right] \\ 
&= \underset{I_{11}}{\underbrace{\sum_{t=1}^T w_{t,T}\left[(\bm{A}_t \bar{\bm{\Delta}}_t) (\bm{A}_t \bar{\bm{\Delta}}_t)^\top \right]}} + \underset{I_{12}}{\underbrace{\sum_{t=1}^T w_{t,T}\left[(\bm{A}_t \bar{\bm{\Delta}}_t)(\bm{A}_t \bm{\theta}^\star - \bm{b}_t)^\top + (\bm{A}_t \bm{\theta}^\star - \bm{b}_t) (\bm{A}_t \bar{\bm{\Delta}}_t)^\top \right]}}.
\end{align*}
By triangle inequality, the Frobenius norm of $I_{11}$ is bounded by
% \begin{align}\label{eq:Gamma-bound-I11-norm}
% \left\|\frac{1}{T} \sum_{t=1}^T \left[(\bm{A}_t \bar{\bm{\Delta}}_T) (\bm{A}_t \bar{\bm{\Delta}}_T)^\top \right]\right\|_{\mathsf{F}} 
% &\leq \frac{1}{T}\sum_{t=1}^T \left\|(\bm{A}_t \bar{\bm{\Delta}}_T) (\bm{A}_t \bar{\bm{\Delta}}_T)^\top\right\|_{\mathsf{F}} \nonumber \\ 
% &\leq \frac{1}{T}\sum_{t=1}^T \sqrt{\mathsf{Tr}\left((\bm{A}_t \bar{\bm{\Delta}}_T) (\bm{A}_t \bar{\bm{\Delta}}_T)^\top(\bm{A}_t \bar{\bm{\Delta}}_T) (\bm{A}_t \bar{\bm{\Delta}}_T)^\top\right)} \nonumber \\ 
% &\leq \frac{1}{T}\sum_{t=1}^T 4\|\bar{\bm{\Delta}}_T\|_2^2 = 4\|\bar{\bm{\Delta}}_T\|_2^2.
% \end{align}
\begin{align}\label{eq:Gamma-bound-I11-norm}
\left\|\sum_{t=1}^T w_{t,T}\left[(\bm{A}_t \bar{\bm{\Delta}}_t) (\bm{A}_t \bar{\bm{\Delta}}_t)^\top \right]\right\|_{\mathsf{F}} 
&\leq \sum_{t=1}^T  w_{t,T}\left\|(\bm{A}_t \bar{\bm{\Delta}}_t) (\bm{A}_t \bar{\bm{\Delta}}_t)^\top\right\|_{\mathsf{F}} \nonumber \\ 
&\leq \sum_{t=1}^T w_{t,T}\sqrt{\mathsf{Tr}\left((\bm{A}_t \bar{\bm{\Delta}}_t) (\bm{A}_t \bar{\bm{\Delta}}_t)^\top(\bm{A}_t \bar{\bm{\Delta}}_t) (\bm{A}_t \bar{\bm{\Delta}}_t)^\top\right)} \nonumber \\ 
&\leq \sum_{t=1}^T w_{t,T} \cdot 4\|\bar{\bm{\Delta}}_t\|_2^2.
\end{align}

% Since 
% \begin{align*}
% \mathbb{E}\|\bar{\bm{\Delta}}_T\|_2^2 &= \frac{1}{T} \mathsf{Tr}(\bar{\bm{\Lambda}}_T) + \tilde{C}T^{-\alpha-1} = \frac{1}{T} \mathsf{Tr}(\bm{\Lambda}^\star) + o(T^{-1})  \leq \frac{\mathsf{Tr}(\bm{\Gamma})}{\lambda_0^2(1-\gamma)^2 T} +o(T^{-1}),
% \end{align*}
Since 
\begin{align*}
\mathbb{E}\|\bar{\bm{\Delta}}_t\|_2^2 &= \frac{1}{t} \mathsf{Tr}(\bar{\bm{\Lambda}}_t) + \tilde{C}t^{-\alpha-1} = \frac{1}{t} \mathsf{Tr}(\bm{\Lambda}^\star) + o(t^{-1})  \leq \frac{\mathsf{Tr}(\bm{\Gamma})}{\lambda_0^2(1-\gamma)^2 t} +o(t^{-1}),
\end{align*}
the expectation of this bound can be bounded by
\begin{align*}
\mathbb{E}\left\|\sum_{t=1}^T w_{t,T}\left[(\bm{A}_t \bar{\bm{\Delta}}_t) (\bm{A}_t \bar{\bm{\Delta}}_t)^\top \right]\right\|_{\mathsf{F}} 
&\leq 4\sum_{t=1}^T w_{t,T} \cdot \mathbb{E}\|\bar{\bm{\Delta}}_t\|_2^2 \\ 
&\leq 4\sum_{t=1}^T \frac{2t}{T(T+1)} \cdot \left(\frac{\mathsf{Tr}(\bm{\Gamma})}{\lambda_0^2(1-\gamma)^2 t} +o(t^{-1})\right) \\ 
&\asymp\frac{\mathsf{Tr}(\bm{\Gamma})}{\lambda_0^2(1-\gamma)^2 T}+o(T^{-1});
\end{align*}

Consequently, the Markov's inequality indicates
\begin{align}\label{eq:Gamma-bound-I11-whp}
\left\|I_{11}\right\|_{\mathsf{F}} \leq \frac{\sqrt{\mathsf{Tr}(\bm{\Gamma})}}{\lambda_0(1-\gamma) \sqrt{T}} + o(T^{-\frac{1}{2}})
\end{align}
with probability at least $1-\frac{\sqrt{\mathsf{Tr}(\bm{\Gamma})}}{\lambda_0(1-\gamma) \sqrt{T}}$.

Meanwhile, the norm of $I_{12}$ is bounded by
\begin{align}\label{eq:Gamma-bound-I12}
&\left\|\sum_{t=1}^T w_{t,T}\left[(\bm{A}_t \bar{\bm{\Delta}}_t)(\bm{A}_t \bm{\theta}^\star - \bm{b}_t)^\top + (\bm{A}_t \bm{\theta}^\star - \bm{b}_t) (\bm{A}_t \bar{\bm{\Delta}}_t)^\top \right]\right\|_{\mathsf{F}} \nonumber \\
&\leq \sum_{t=1}^T w_{t,T}\left\|(\bm{A}_t \bar{\bm{\Delta}}_t)(\bm{A}_t \bm{\theta}^\star - \bm{b}_t)^\top + (\bm{A}_t \bm{\theta}^\star - \bm{b}_t) (\bm{A}_t \bar{\bm{\Delta}}_t)^\top\right\|_{\mathsf{F}} \nonumber \\ 
&\leq 2\sum_{t=1}^T w_{t,T}\left\|(\bm{A}_t \bar{\bm{\Delta}}_t)(\bm{A}_t \bm{\theta}^\star - \bm{b}_t)^\top\right\|_{\mathsf{F}} \nonumber \\
&\leq 2\sum_{t=1}^T w_{t,T}\sqrt{\mathsf{Tr}\left((\bm{A}_t \bm{\theta}^\star - \bm{b}_t)(\bm{A}_t \bar{\bm{\Delta}}_t)^\top(\bm{A}_t \bar{\bm{\Delta}}_t)(\bm{A}_t \bm{\theta}^\star - \bm{b}_t)^\top \right)}\nonumber \\ 
&\leq 4\sum_{t=1}^T w_{t,T}(2\|\bm{\theta}^\star\|_2+1)\|\bar{\bm{\Delta}}_t\|_2\nonumber \\ 
&= 4(2\|\bm{\theta}^\star\|_2+1)\sum_{t=1}^T w_{t,T}\|\bar{\bm{\Delta}}_t\|_2.
\end{align}
% \begin{align}\label{eq:Gamma-bound-I12}
% &\left\|\frac{1}{T} \sum_{t=1}^T \left[(\bm{A}_t \bar{\bm{\Delta}}_T)(\bm{A}_t \bm{\theta}^\star - \bm{b}_t)^\top + (\bm{A}_t \bm{\theta}^\star - \bm{b}_t) (\bm{A}_t \bar{\bm{\Delta}}_T)^\top \right]\right\|_{\mathsf{F}}\\
% &\leq \frac{1}{T}\sum_{t=1}^T \left\|(\bm{A}_t \bar{\bm{\Delta}}_T)(\bm{A}_t \bm{\theta}^\star - \bm{b}_t)^\top + (\bm{A}_t \bm{\theta}^\star - \bm{b}_t) (\bm{A}_t \bar{\bm{\Delta}}_T)^\top\right\|_{\mathsf{F}} \nonumber \\ 
% &\leq \frac{2}{T}\sum_{t=1}^T\left\|(\bm{A}_t \bar{\bm{\Delta}}_T)(\bm{A}_t \bm{\theta}^\star - \bm{b}_t)^\top\right\|_{\mathsf{F}} \nonumber \\
% &\leq \frac{2}{T}\sum_{t=1}^T \sqrt{\mathsf{Tr}\left((\bm{A}_t \bm{\theta}^\star - \bm{b}_t)(\bm{A}_t \bar{\bm{\Delta}}_T)^\top(\bm{A}_t \bar{\bm{\Delta}}_T)(\bm{A}_t \bm{\theta}^\star - \bm{b}_t)^\top \right)}\nonumber \\ 
% &\leq \frac{4}{T}\sum_{t=1}^T 4(2\|\bm{\theta}^\star\|_2+1)\|\bar{\bm{\Delta}}_T\|_2\nonumber \\ 
% &= 4(2\|\bm{\theta}^\star\|_2+1)\|\bar{\bm{\Delta}}_T\|_2.
% \end{align}
Furthermore, since $\sum_{t=1}^T w_{t,T} = 1$, the Jensen's inequality guarantees
\begin{align*}
\mathbb{E} \left(\sum_{t=1}^T w_{t,T}\|\bar{\bm{\Delta}}_t\|_2\right)^2 &\leq \sum_{t=1}^T w_{t,T} \mathbb{E}\|\bar{\bm{\Delta}}_t\|_2^2 \\ 
&\leq \sum_{t=1}^T \frac{2t}{T(T+1)} \left(\frac{\mathsf{Tr}(\bm{\Gamma})}{\lambda_0^2(1-\gamma)^2 t} +o(t^{-1})\right) \\ 
&\asymp\frac{\mathsf{Tr}(\bm{\Gamma})}{\lambda_0^2(1-\gamma)^2 T}+o(T^{-1})
\end{align*}
Therefore, the Chebyshev's inequality indicates
\begin{align}\label{eq:Gamma-bound-I12-whp}
&\left\|I_{22}\right\|_{\mathsf{F}} \lesssim 4(2\|\bm{\theta}^\star\|_2+1)\frac{\sqrt{\mathsf{Tr}(\bm{\Gamma})}}{\lambda_0(1-\gamma) }T^{-\frac{1}{3}} + o(T^{-\frac{1}{3}})
\end{align}
with probability at least $1-T^{-\frac{1}{3}}$.

Hence by combining \eqref{eq:Gamma-bound-I11-whp} and \eqref{eq:Gamma-bound-I12-whp}, we obtain
\begin{align}\label{eq:bound-I1-Gamma}
\|I_1\|_{\mathsf{F}} \lesssim 4(2\|\bm{\theta}^\star\|_2+1)\frac{\sqrt{\mathsf{Tr}(\bm{\Gamma})}}{\lambda_0(1-\gamma) }T^{-\frac{1}{3}} + o(T^{-\frac{1}{3}})
\end{align} 
with probability at least $1-T^{-\frac{1}{3}}-\frac{\sqrt{\mathsf{Tr}(\bm{\Gamma})}}{\lambda_0(1-\gamma) \sqrt{T}}$. 

\paragraph{Bounding the Frobenius norm of $I_2$.} Notice that $I_2$ only involves the randomness of $i.i.d.$ samples $\{(\bm{A}_t,\bm{b}_t)\}$, and 
\begin{align*}
&\left\|\textbf{vec}\left((\bm{A}_t {\bm{\theta}}^\star - \bm{b}_t)(\bm{A}_t {\bm{\theta}^\star} -\bm{b}_t)^\top\right)\right\|_2\\ 
&= \left\|(\bm{A}_t {\bm{\theta}}^\star - \bm{b}_t)(\bm{A}_t {\bm{\theta}^\star} -\bm{b}_t)^\top\right\|_{\mathsf{F}} \\ 
&= \sqrt{\mathsf{Tr}((\bm{A}_t {\bm{\theta}}^\star - \bm{b}_t)(\bm{A}_t {\bm{\theta}^\star} -\bm{b}_t)^\top(\bm{A}_t {\bm{\theta}}^\star - \bm{b}_t)(\bm{A}_t {\bm{\theta}^\star} -\bm{b}_t)^\top)} \\ 
&< (2\|\bm{\theta}^\star\|_2+1)^2
\end{align*}
holds uniformly for all $t \in [T]$. Consequently, the Frobenius norm of $I_2$ can be bounded by directly applying the vector Azuma's inequality: for every $\delta \in (0,1)$, 
\begin{align}\label{eq:bound-I2-Gamma}
\left\|I_2\right\|_{\mathsf{F}} \leq 2\sqrt{2} \sqrt{\sum_{t=1}^T w_{t,T}^2}(2\|\bm{\theta}^\star\|_2+1)^2 \sqrt{\log \frac{3}{\delta}};
\end{align}
with probability at least $1-\delta$. 
% \begin{align}\label{eq:bound-I2-Gamma}
% \left\|I_2\right\|_{\mathsf{F}} \leq \frac{2\sqrt{2}}{\sqrt{T}}(2\|\bm{\theta}^\star\|_2+1)^2 \sqrt{\log \frac{3}{\delta}}
% \end{align}
Since  
\begin{align*}
\sum_{t=1}^T w_{t,T}^2 &= \frac{4}{T^2(T+1)^2} \sum_{t=1}^T t^2 \\ 
&= \frac{4}{T^2(T+1)^2} \cdot \frac{T(T+1)(2T+1)}{6} \asymp \frac{1}{T},
\end{align*}
Lemma \ref{lemma:Gamma} follows by combining \eqref{eq:bound-I1-Gamma} and \eqref{eq:bound-I2-Gamma} with $\delta = T^{-\frac{1}{3}}$.\qed

\section{Experimental details}\label{app:experiment-details}

\subsection{Online algorithm to construct the variance estimator}\label{app:online-alg}

The code for our experiments can be found at \url{https://github.com/Weichen-Wu-1996/TD_inference}. Experiments were performed on a personal laptop (MacOS). No external GPU clusters or cloud resources were used.

\begin{algorithm}[htbp]
\caption{Online algorithm for computing $\bar{\bm{\theta}}_T$ and $\hat{\bm{\Lambda}}_T$}\label{alg:online}
\begin{algorithmic}[1]
\Require MDP transition kernel $\mathcal{P}$, stationary distribution $\mu$, discount factor $\gamma$, feature map $\bm{\phi}$, initial stepsize $\eta_0$, stepsize decay parameter $\alpha \in (0.5, 1)$, sample size $T$
\State Initialize $\bm{\theta}= \bm{0}$, $\bar{\bm{\theta}} = \bm{0}$, $\bar{\bm{A}} = \bm{0}$, $\hat{\bm{\Gamma}} = \bm{0}$.
\State \textbf{for} $t=1,\dots,T$ \textbf{do}
\Statex \quad Sample \(s \sim \mu\), \(s' \sim \mathcal{P}(\cdot \mid s)\).
\Statex \quad Compute: \(\bm{A} \gets \bm{\phi}(s)[\bm{\phi}(s)- \gamma \bm{\phi}(s')],\ \bm{b} \gets r(s)\,\bm{\phi}(s)\).
\Statex \quad Perform TD update: \(\bm{\theta} \gets \bm{\theta} - \eta_0 t^{-\alpha} (\bm{A}\bm{\theta} - \bm{b})\).
\Statex \quad Perform online updates:
\Statex \quad \(\bar{\bm{\theta}} \gets \bar{\bm{\theta}} + \tfrac{1}{t}\,(\bm{\theta} - \bar{\bm{\theta}})\).
\Statex \quad \(\bar{\bm{A}} \gets \bar{\bm{A}} + \tfrac{1}{t}\,(\bm{A} - \bar{\bm{A}})\).
\Statex \quad \(\hat{\bm{\Gamma}} \gets \hat{\bm{\Gamma}} + \tfrac{2}{t+1}\big[(\bm{A}\bar{\bm{\theta}} - \bm{b})(\bm{A}\bar{\bm{\theta}} - \bm{b})^\top - \hat{\bm{\Gamma}}\big]\).
\State \textbf{end for}
\State Construct the variance estimator:
\State \(\hat{\bm{\Lambda}} \gets \bar{\bm{A}}^{-1}\,\hat{\bm{\Gamma}}\,\bar{\bm{A}}^{-\top}\).
\Ensure Averaged TD estimator \(\bar{\bm{\theta}}\), covariance matrix estimator \(\hat{\bm{\Lambda}}\).
\end{algorithmic}
\end{algorithm}

It is easy to verify that by definition \eqref{eq:defn-hat-Gamma}, the estimators $\bar{\bm{A}}_T$ and $\hat{\bm{\Gamma}}_T$ can be updated by
\begin{align*}
&\bar{\bm{A}}_T = \bar{\bm{A}}_{T-1} + \frac{1}{T}(\bm{A}_T - \bar{\bm{A}}_{T-1}), \quad \text{and} \\ 
&\hat{\bm{\Gamma}}_T = \hat{\bm{\Gamma}}_{T-1} + \frac{2}{T+1} \left((\bm{A}_T\bar{\bm{\theta}}_T - \bm{b}_T)(\bm{A}_T\bar{\bm{\theta}}_T - \bm{b}_T)^\top - \hat{\bm{\Gamma}}_{T-1}\right).
\end{align*}

Algorithm \ref{alg:online} reveals the complete process of generating both the averaged TD estimator $\bar{\bm{\theta}}_T$ and the variance estimator $\hat{\bm{\Lambda}}_T$. Since Algorithm \ref{alg:online} iterates from $t=0$ to $t=T-1$ in a single loop and does not store the samples for each iteration, it has a space complexity of $O(Td^2+d^3)$ and a space complexity of $O(d^2)$. Here, the $d^3$ term stems from calculating matrix inverse.

\subsection{The Markov Decision Process and the feature map}
Given a finite state space $\mathcal{S}$, define a sequence of MDP $\{\mathcal{M}_{\bm{q}}\}$ indexed by $\bm{q} \in \mathcal{Q} \subset \{q_+, q_-\}^{d-1}$ where for each $\bm{q}$, the transition kernel equals
\begin{align}
\label{eqn:lb-kernel}
&P_{\bm{q}}(s' \mid s) \nonumber \\ 
&= 
\left\{ \begin{array}{lcc}
q_s\mathds{1}(s' = s) + \frac{1-q_s}{|\mathcal{S}|-d+1}\mathds{1}(s' \ge d) & \text{for} & s < d; \\[0.2cm]
\frac{\gamma}{|\mathcal{S}|-d+1}\mathds{1}(s' \ge d) + \frac{1-q_{s'}}{d-1}\mathds{1}(s' < d) & \text{for} & s \ge d.
\end{array}\right.
\end{align}
and the reward function equals $r(s) = \mathds{1}(s \ge d)$.  

Here, for each $i \in [d-1]$, $q_{i}$ is taken to be either $q_+$ or $q_{-}$ where  
\begin{align*}
q_+ := \gamma + (1-\gamma)^2\varepsilon,\qquad\text{and}\qquad q_- := \gamma - (1-\gamma)^2\varepsilon.
\end{align*}
% In addition, the additional constants in expression~\eqref{eqn:lb-kernel} are chosen as 
% \begin{align*}
% p_s \defn \frac{1-q_s}{(d-1)(1-q)}\qquad\text{for } q \defn \frac{1}{d - 1} \sum_{s = 1}^{d-1} q_s.
% \end{align*}

We further impose the constraint that the number of $q_+$'s and $q_-$'s in $\bm{q}$ are the same, namely, 
\begin{align}
\label{eqn:brahms}
\sum_{s=1}^{d-1} \mathds{1}(q_s = q_+) = \sum_{s=1}^{d-1} \mathds{1}(q_s = q_-)= (d-1)/2.
\end{align}
%which in turn implies $q = (q_+ + q_-)/2 = \gamma.$ 
Here without loss of generality, assume $d$ is an odd number. 
With these definitions in place, it can be easily verified that the stationary distribution for $\bm{P}$ obeys  
\begin{align}
\mu(s) = \left\{ \begin{array}{lcc}
\frac{1}{2(d-1)} & \text{for} & s < d; \\[0.2cm]
\frac{1}{2(|\mathcal{S}| - d + 1)} & \text{for} & s \ge d. 
\end{array}\right.
\end{align}
% where $\mu_0^{-1} = 1 + \frac{1-\gamma}{1-q}$. 
Moreover, suppose the feature map is taken to be   
\begin{align*}
	\bm{\phi}(s) = \bm{e}_{s \wedge d} \in \mathbb{R}^d, 
\end{align*}
then one can further verify that
\begin{align}\label{eq:simulate-theta-star}
\theta^{\star}(d) &= V^{\star}(s) = \frac{1}{1-\gamma^2 - \sum_{i = 1}^{d-1}\frac{\gamma^2(1 - q_i)^2}{(d-1)(1-\gamma q_i)}}, \nonumber \\
\theta^{\star}(i) &= V^{\star}(i) = \frac{\gamma(1 - q_i)}{1-\gamma q_i}V^{\star}(s), 
\text{ for } s \ge d \text{ and } i < d.
\end{align}
From the expressions above, we remark that, the values of $q$ and $V^\star(s)$ with $s \geq d$ are fixed for all $\bm{q} \in \mathcal{Q}$ which is ensured by the construction~\eqref{eqn:brahms}. It is shown in Theorem 2 of \cite{li2023sharp} that the sequence $\{\mathcal{M}_{\bm{q}}\}_{q \in \mathcal{Q}}$ is hard to distinguish for a considerably large set $\mathcal{Q}$. 

\subsection{Parameter specification}
In our experiments, we fix the size of the state space to $|\mathcal{S}| = 10$, the discount factor to be $\gamma = 0.2$, and let $\varepsilon = 0.01$. For the TD learning algorithm, we set the initial stepsize to be $\eta_0 = 5$, and total number of iterations to be $T = 2^{20}$. Data is recorded at every power of 2, so that the log plot has constant intervals on the x-axis. In order to show the role played by the stepsize decaying speed $\alpha$, we fix $d = 3$ and let $\alpha$ iterate through the group $(\frac{1}{2},\frac{2}{3},\frac{3}{4},1$; and in order to illustrate the influence of the dimension of the feature map, we fix $\alpha = \frac{2}{3}$ and let $d$ iterate through the group of $(3,5,7,9)$. For every given $d$, we set $\bm{q}$ to be
\begin{align*}
q_i = \begin{cases}
q_+, \quad \text{for } 1 \leq i \leq \frac{d-1}{2}; \\ 
q_-, \quad \text{for } \frac{d+1}{2} \leq i \leq d.
\end{cases}
\end{align*}

We verify our theoretical results through the following:
\begin{enumerate}
\item In order to illustrate the high-probability convergence of the TD estimation error as indicated by Theorem \ref{thm:bar-delta-t}, we measure the $95\%$ quantile of the $L_2$ norm of $\bar{\bm{\Delta}}_T$ in $10,000$ independent trials;
\item In order to illustrate the Berry-Esseen bound as indicated by Theorems \ref{thm:Berry-Esseen} and \ref{thm:asymptotic}, we measure the difference between the empirical and asymptotic distributions of $\bar{\bm{\Delta}}_T$ by
\begin{align*}
d(\sqrt{T}\bar{\bm\Delta}_T,\mathcal{N}(\bm{0},\bm{\Lambda}^\star)) 
:= \max_{1 \leq j \leq d} \sup_{x \in \mathbb{R}} \left|\mathbb{P}(\sqrt{T}(\bar{\bm\Delta}_T)_j \leq x) - \mathbb{P}((\bm{\Lambda}^{\star \frac{1}{2}}\bm{z})_j \leq x)\right|,
\end{align*}
averaged among $10,000$ independent trials.
Besides, as an intuitive demonstration, Figure \ref{fig:iid-d3-alpha067-deltaT1-converge} in the introduction section records the empirical and asymptotic distributions of $(\bar{\bm{\Delta}}_T)_1$ when $d = 3$ and $\alpha = \frac{2}{3}$.
\item In order to illustrate the efficiency of our variance estimator $\hat{\bm{\Lambda}}_T$, we measure the averaged Frobenius norm of the difference between $\hat{\bm{\Lambda}}_T$ and $\bm{\Lambda}^\star$ among $10,000$ independent trials. As is indicated by \ref{thm:DMR}, this is a direct proxy for the TV distance between the zero-mean Gaussian distributions with the two corresponding variance matrices.
\item In order to validate our statistical inference procedure, we run Algorithm \ref{alg:online} on the MDP with $d=3$ using $\alpha = \frac{2}{3}$, construct both individual and simultaneous confidence intervals for $\bm{\theta}^\star$, and measure the coverage rates of these confidence intervals in $10,000$ independent trials. Details to follow in section \ref{app:coverage}.
\end{enumerate}

\subsection{Full experimental results}
\subsubsection{High-probability convergence guarantee for the TD estimator}
\begin{figure*}[h]
\begin{center}
\begin{tabular}{cc}
\includegraphics[width = 0.48\textwidth]{imgs/figure_3a.pdf} & \includegraphics[width = 0.48\textwidth]{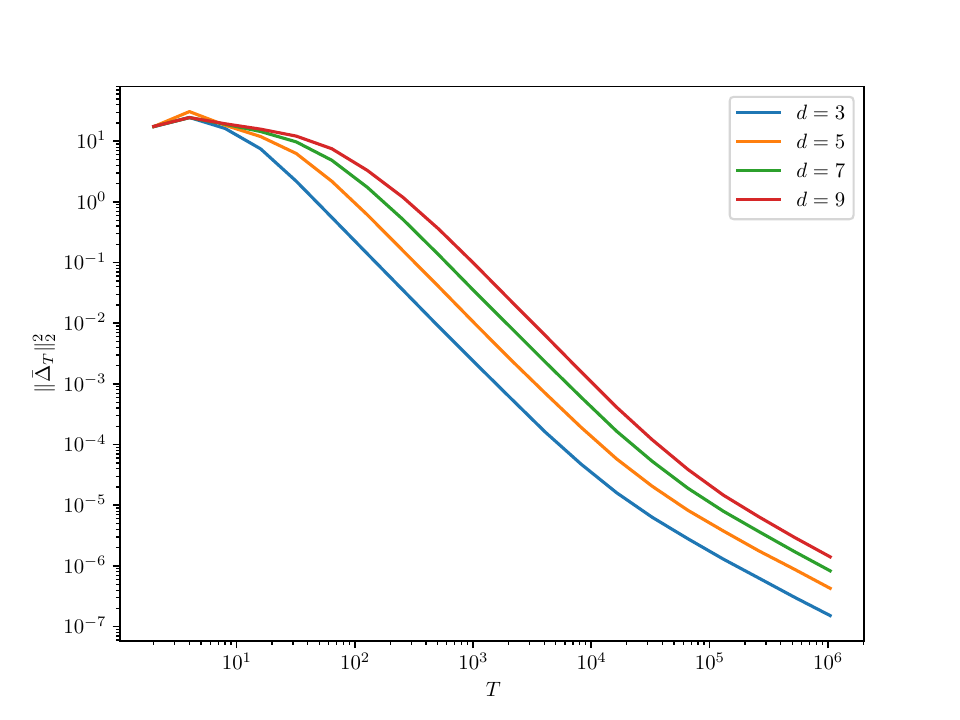}\\
\end{tabular}
\end{center}
\caption{The decay of the $95\%$ quantile of $\|\bar{\bm{\Delta}}_T\|_2^2$ when (a) $d = 3$, for different choices of $\alpha$; and (b) $\alpha = \frac{2}{3}$, for different choices of $d$.}
\label{fig:L2-converge}
\end{figure*}

Figure \ref{fig:L2-converge} shows the decay of the $95\%$ quantile of $\|\bar{\bm{\Delta}}_T\|_2^2$; the slope of this log-log plot verifies our theoretical claim in Theorem \ref{thm:bar-delta-t} that $\|\bar{\bm{\Delta}}_T\|_2$ converges by the rate of $O(T^{-\frac{1}{2}})$ with high probability. 

{%\color{red}
Plot (a) illustrates the behavior of the estimation error under different choices of $\alpha$. As a reference, the black dotted line represents the $95\%$ quantile of $\{\bm{Z}_T\}$ where $\bm{Z}_T \sim \mathcal{N}(\bm{0},\bm{\Lambda}^\star/T)$. Our theory predicts that when $T$ grows large, this should be the asymptotic line for the empirical $95\%$ quantiles when $\alpha \in (\frac{1}{2},1)$. This is exactly what we observe that $\alpha=\frac{2}{3}$ and $\alpha=\frac{3}{4}$. On the contrary, when $\alpha = 1$, the empirical quantiles do \emph{not} converge to the asymptotic quantiles. This is also predicted by our theory that since $\bar{\bm{\Lambda}}_T-\bm{\Lambda}^\star = \bm{O}(T^{\alpha-1})$ (which does not converge when $\alpha=1$), the variance of $\bar{\bm{\Delta}}_T$ will \emph{not} converge to $\bm{\Lambda}^\star$ when $\alpha=1$. This fact is also reflected in the $O(T^{\frac{\alpha}{2}-1})$ term in the bound of Theorem \ref{thm:bar-delta-t}, which when $\alpha = 1$, converges by $O(T^{-\frac{1}{2}})$, the same rate as the leading term. As for the case of $\alpha=\frac{1}{2}$, we observe that when $T$ is small,  $\|\bar{\bm{\Delta}}_T\|_2$ is relatively large; it is not until a very large number of iterations (i.e., $T > 10^5$), that the empirical and asymptotic lines begin to overlap. This phenomenon is also predicted by the $O(T^{-\frac{\alpha+1}{2}})$ term in the bound of Theorem \ref{thm:bar-delta-t}, which converges slower when $\alpha = \frac{1}{2}$. A more detailed explanation follows in Appendix \ref{app:experiment-decompose}.
}

Meanwhile, plot(b) shows that, fixing $\alpha = \frac{2}{3}$, the convergence is slower when $d$ is larger. Both are in accordance with our conclusion in Theorem (1) that $\|\bar{\bm{\Delta}}_T\|_2$ converges by the rate of 
\begin{align*}
O\left(\sqrt{\frac{\mathsf{Tr}(\bm{\Lambda}^\star)}{T}\log \frac{1}{\delta}}\right)
\end{align*}
with probability at least $1-\delta$.

% Figure \ref{fig:iid-d3-converge} (b) shows the rate by which the empirical distribution of $\sqrt{T}\bar{\bm{\Delta}}_T$ converges to its asymptotic distribution $\mathcal{N}(\bm{0},\bm{\Lambda}^\star)$. 
% As is indicated by Figure \ref{fig:iid-d3-converge} (b), the distirbution of $\sqrt{T}\bar{\bm{\Delta}}_T$ converges to $\mathcal{N}(\bm{0},\bm{\Lambda}^\star)$ by a similar rate when $\alpha = \frac{2}{3}, \frac{3}{4}$ and $\frac{4}{5}$; when $\alpha = \frac{1}{2}$, the convergence is slower. We point out that this result is in contradictory to the proposal by \cite{samsonov2024gaussian}. 

\subsubsection{Berry-Esseen bound}
\begin{figure*}[h]
\begin{center}
\begin{tabular}{cc}
\includegraphics[width = 0.48\textwidth]{imgs/figure_2.pdf} & \includegraphics[width = 0.48\textwidth]{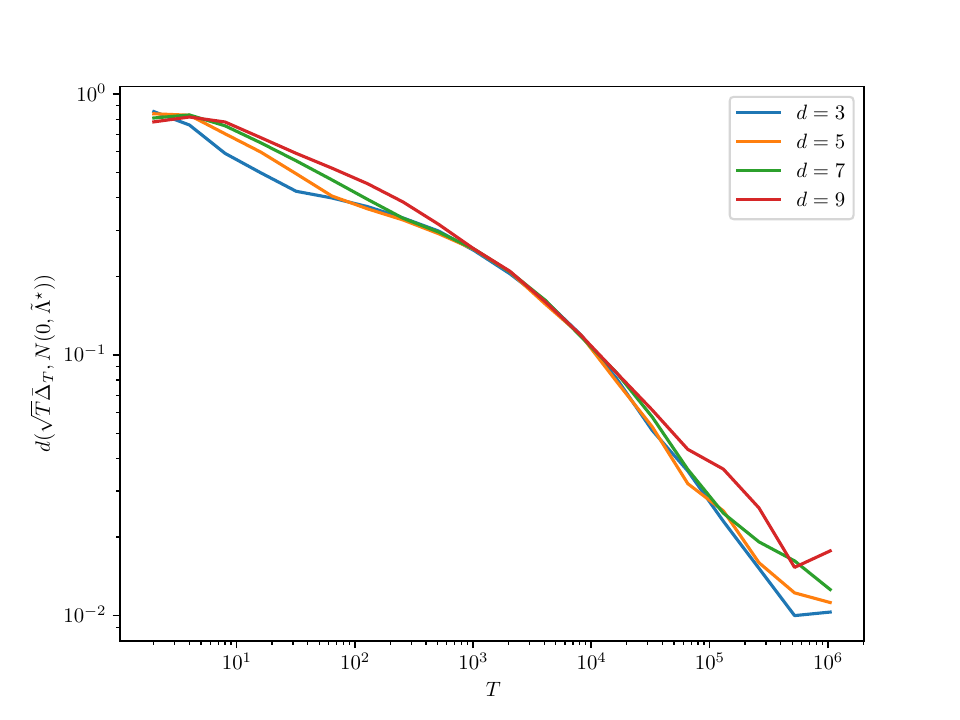}\\
\end{tabular}
\end{center}
\caption{The decay of $d(\sqrt{T}\bar{\bm{\Delta}}_T,\mathcal{N}(\bm{0},\bm{\Lambda}^\star))$, when (a) $d=3$, for different choices of $\alpha$; and (b) $\alpha = \frac{2}{3}$, for different choices of $d$.}
\label{fig:Berry-Esseen-experiment}
\end{figure*}

Figure \ref{fig:Berry-Esseen-experiment} illustrates how the finite-sample distribution of $\sqrt{T}\bar{\bm{\Delta}}_T$ converges to its asymptotic Gaussian distribution $\mathcal{N}(\bm{0},\bm{\Lambda}^\star)$. 
For ease of computation, instead of the convex distance we evaluated the distance
\begin{align}
\label{eqn:distance-measure-fig2}
\max_{1 \leq j \leq d} \sup_{x \in \mathbb{R}} \left|\hat{\mathbb{P}}(\sqrt{T}(\bar{\bm\Delta}_T)_j \leq x) - \mathbb{P}_{\bm{z}}((\bm{\Lambda}^{\star \frac{1}{2}}\bm{z})_j \leq x)\right|,
\end{align}
where $\hat{\mathbb{P}}$ refers to the empirical distribution in our $10,000$ trials, and $\bm{z}$ is the $d$-dimensional standard Gaussian.

Notably, plot (a) indicates that the convergence is slower when $\alpha = \frac{1}{2}$, compared to the choices of $\alpha =\frac{2}{3}$ and $\alpha = \frac{3}{4}$. This directly contradicts the suggestion by \cite{samsonov2024gaussian} to choose $\alpha = \frac{1}{2}$ in order to yield a faster in-distribution convergence of the TD estimation error. On the other hand, when $\alpha = 1$, the empirical distribution does \emph{not} converge to $\mathcal{N}(\bm{0},\bm{\Lambda}^\star)$; that verifies our claim in Appendix \ref{sec:tightness.thm:asymptotic} that the $O(T^{\alpha-1})$ term in the Berry-Esseen bound is unavoidable.

Meanwhile, plot (b) shows the convergence generally is slower when $d$ is larger.

\subsubsection{Estimator of the covariance matrix}

\begin{figure*}[h]
\begin{center}
\begin{tabular}{cc}
\includegraphics[width = 0.48\textwidth]{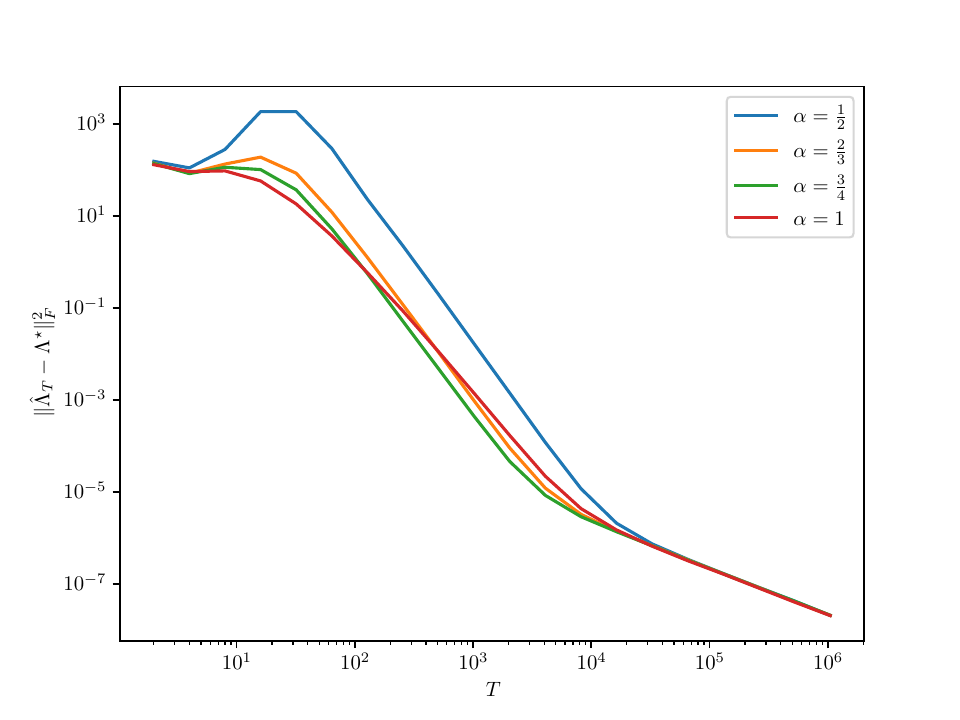} & \includegraphics[width = 0.48\textwidth]{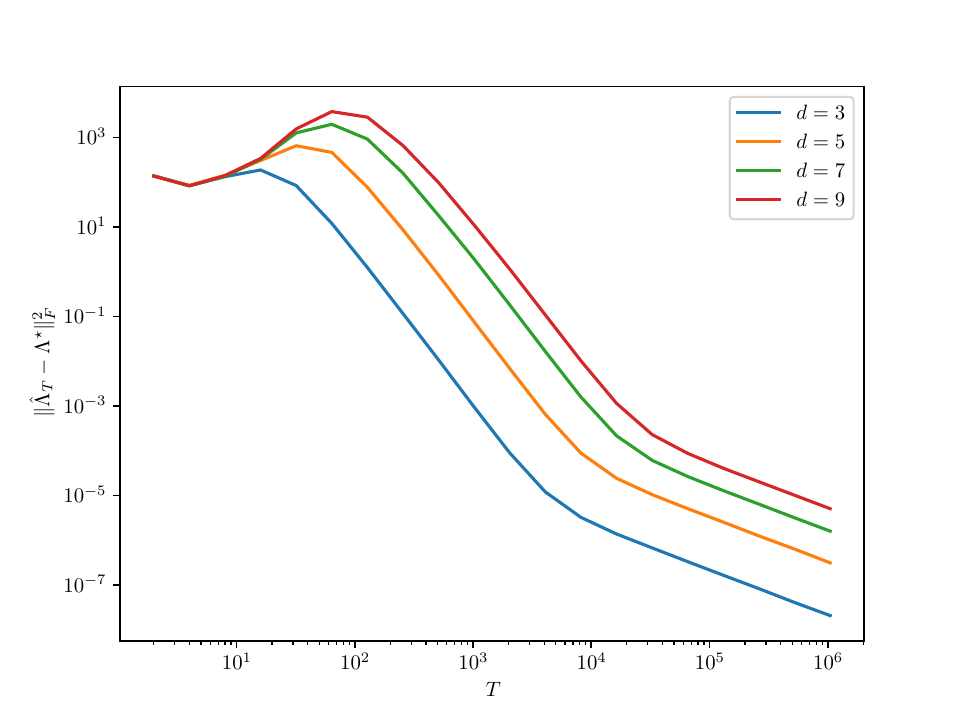}\\
\end{tabular}
\end{center}
\caption{The decay of the difference between estimated and asymptotic covariance matrices, measured in squared Frobenius norm, when (a) $d = 3$, for different choices of $\alpha$; and (b) $\alpha = \frac{2}{3}$, for different choices of $d$.}
\label{fig:hat-Lambda}
\end{figure*}

Figure \ref{fig:hat-Lambda} illustrates the rate by which our variance estimator $\hat{\bm{\Lambda}}_T$ converges to the asymptotic variance $\bm{\Lambda}^\star$. Plot (a) shows that the convergence rate is similar for different choices of $\alpha$ when the number of iterations $T$ is sufficiently large; plot(b) indicates that the convergence is slower when $d$ is larger, the nearly parallel curves in the log-log plot demonstrate that the convergence rate with respect to $T$ is similar among different choices of $d$. 

\subsubsection{Confidence intervals}\label{app:coverage}
\begin{figure*}[h]
\begin{center}
\begin{tabular}{cc}
\includegraphics[width = \textwidth]{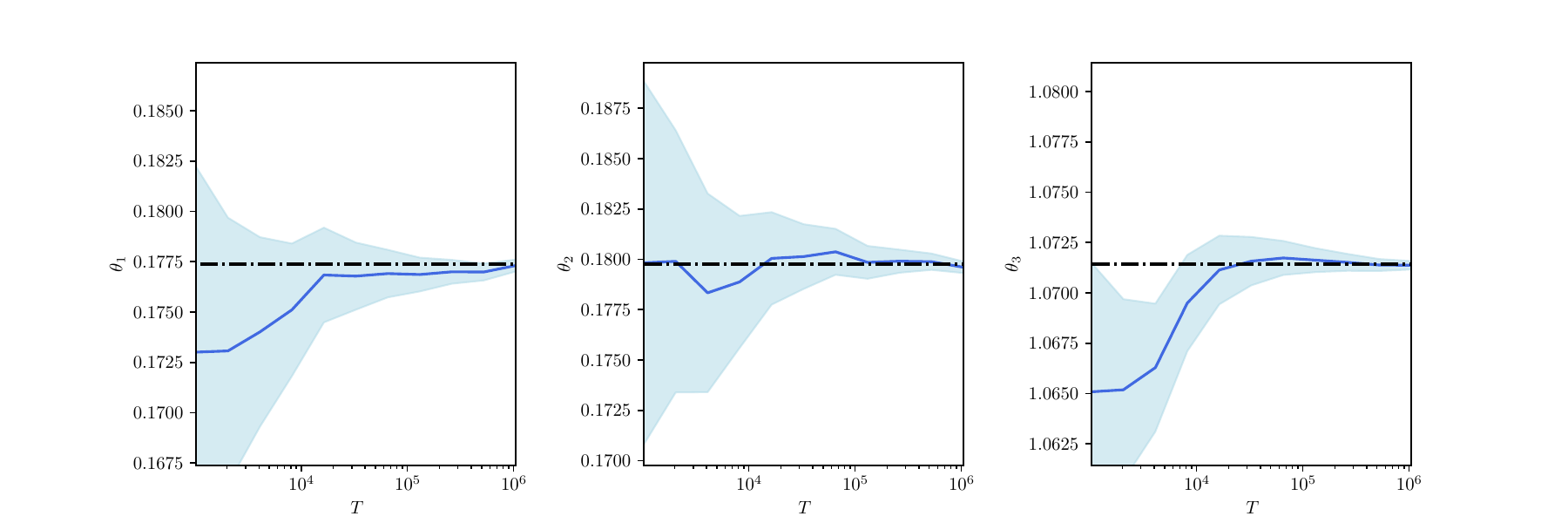} \\
\end{tabular}
\end{center}
\caption{The emergence of the point estimate $\bar{\bm{\theta}}_T$ (blue curves) and individual confidence intervals (blue shades) on the three entries of the optimal coefficients $\bm{\theta}^\star$ (black dotted lines) for the MDP with $d=3$, using TD with Polyak-Ruppert averaging, $\alpha = \frac{2}{3}$}
\label{fig:coverage-1trial}
\end{figure*}

Finally, we validate our statistical inference framework on the MDP with $d=3$ and TD learning with $\alpha = \frac{2}{3}$. We run 10,000 independent trials of Algorithm \ref{alg:online}, and at each iteration $T$, we construct both the \emph{individual} and the \emph{simultaneous} confidence intervals. Specifically, the individual confidence intervals for the three entries of $\bm{\theta}^\star$ are constructed by
\begin{align*}
\mathcal{C}_j = \left[\bar{\bm{\theta}}_T - \Phi(1-\delta/2)\cdot \sqrt{\frac{(\hat{\bm{\Lambda}}_T)_{jj}}{T}},\bar{\bm{\theta}}_T + \Phi(1-\delta/2)\cdot \sqrt{\frac{(\hat{\bm{\Lambda}}_T)_{jj}}{T}} \right], \quad \text{for} \quad j = 1,2,3.
\end{align*}
Here, $\Phi(1-\delta/2)$ is the $1-\frac{\delta}{2}$ quantile of the standard Gaussian distribution. In our experiment, we take $\delta = 0.05$ and therefore $\Phi(1-\delta/2)=1.96$. Figure \ref{fig:coverage-1trial} illustrates the emergence of $\bar{\bm{\theta}}_T$ and $\{\mathcal{C}_j\}_{j=1,2,3}$ in \emph{one single trial}. As $T$ grows larger, the three entries of the point estimate $\bar{\bm{\theta}}_T$ (blue curves) converge to the corresponding entries of the optimal coefficients $\bm{\theta}^\star$ (black dotted lines), while the individual confidence intervals $\{\mathcal{C}_j\}_{j=1,2,3}$ (blue shades) also concentrate around $\bm{\theta}^\star$. 

Meanwhile, the simultaneous confidence intervals are constructed as 
\begin{align*}
&\mathcal{C}= \{\bm{\theta}: \|\bm{\theta} - \bar{\bm{\theta}}_T\|_{\infty} \leq x^\star\}, \quad \text{where} \\ 
&x^\star:=\sup\left\{x: \mathbb{P}_{z \sim \mathcal{N}(\bm{0},\bm{I}_3)}\left(\left\|\hat{\bm{\Lambda}}_T^{\frac{1}{2}}\bm{z}\right\|_{\infty} \leq x\right) \leq 1-\delta\right\}.
\end{align*}
Essentially, $x^\star$ is the $1-\delta$ quantile of the $\ell_{\infty}$ norm of the $\mathcal{N}(\bm{0},\hat{\bm{\Lambda}}_T)$ random variable. Calculating this quantity is a non-trivial task, so we use $1000$ independent simulations from $\mathcal{N}(\bm{0},\hat{\bm{\Lambda}}_T)$ and take the $1-\delta$ quantile of the $\ell_{\infty}$ norm of the samples as an estimate. 

\begin{figure*}[h]
\begin{center}
\begin{tabular}{cc}
\includegraphics[width = 0.6\textwidth]{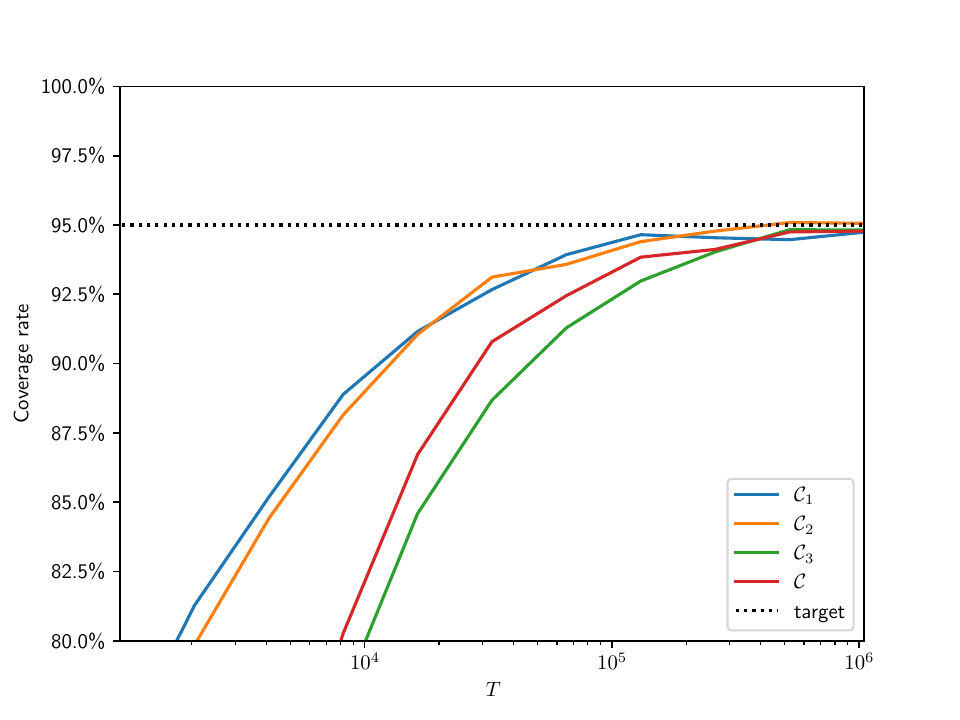} \\
\end{tabular}
\end{center}
\caption{The emergence coverage rates of independent and simultaneous confidence intervals on the three entries of the optimal coefficients $\bm{\theta}^\star$ of the MDP with $d=3$, using TD with Polyak-Ruppert averaging, $\alpha = \frac{2}{3}$. Black dotted line represents the target confidence level of $95\%$; blue, orange and green curves represent the coverage rates of the individual confidence intervals $\mathcal{C}_1$, $\mathcal{C}_2$ and $\mathcal{C}_3$ correspondingly; the red curve represents the coverage rate of the confidence region (simultaneous confidence intervals) $\mathcal{C}$.}
\label{fig:coverage-rate}
\end{figure*}

Figure \ref{fig:coverage-rate} shows the coverage rates of the individual and simultaneous confidence intervals, defined as the percentage of times when the optimal coefficient $\bm{\theta}^\star$ (or its entries) are covered by the confidence intervals in the $10,000$ independent trials. As is indicated by the plot, the coverage rates all approach the target confidence level of $1-\delta = 95\%$ when $T$ grows large.

\subsection{Decomposition of TD estimation error}\label{app:experiment-decompose}
In this section, we aim to further explore the emergence of the TD estimation error $\bar{\bm{\Delta}}_T$ in numerical simulations, in order to verify our key claims in the proofs of our theoretical results. Specifically, we aim to explore the role played by the choice of $\alpha$ through decomposing the TD estimation error. Recall that throughout our proofs, we repetitively used the decomposition
\begin{align*}
\bar{\bm{\Delta}}_T = \underset{I_1}{\underbrace{\frac{1}{T\eta_0} \bm{Q}_0 \bm{\Delta}_0}} - \underset{I_2}{\underbrace{\frac{1}{T} \sum_{i=1}^T \bm{Q}_i(\bm{A}_i \bm{\theta}^\star-\bm{b}_i)}} - \underset{I_3}{\underbrace{\frac{1}{T}\sum_{i=1}^T \bm{Q}_i (\bm{A}_i-\bm{A})\bm{\Delta}_{i-1}}},
\end{align*}
where the matrices $\{\bm{Q}_i\}_{0 \leq i \leq T}$ are defined by \eqref{eq:defn-Qt}, and therefore decided by the matrix $\bm{A}$ as well as the choice of stepsizes $\{\eta_t\}_{1 \leq t \leq T}$. Of the three terms on the right-hand-side:
\begin{enumerate}
\item The term $I_1$ is related only to the initial estimator $\bm{\theta}_0$ through $\bm{\Delta}_0$ and not to the samples $\{(\bm{A}_t,\bm{b}_t)\}_{1 \leq t \leq T}$. Therefore, we refer to this term as the \emph{initialization error}. This term is deterministic, and according to our analysis, converges by the rate of
\begin{align*}
\left\|\frac{1}{T\eta_0} \bm{Q}_0 \bm{\Delta}_0\right\|_2 \lesssim O(T^{-1}).
\end{align*}
\item The term $I_2$ is related only to the samples $\{(\bm{A}_t,\bm{b}_t)\}_{1 \leq t \leq T}$ and not to any of the iterations $\{\bm{\theta}_t\}_{0 \leq t \leq T}$. Therefore, we refer to this term as the \emph{sampling error}. According to our analysis, this term corresponds to the asymptotic error:
\begin{align*}
\frac{1}{\sqrt{T}} \sum_{i=1}^T \bm{Q}_i(\bm{A}_i \bm{\theta}^\star-\bm{b}_i) \xrightarrow{d}\mathcal{N}(\bm{0},\bm{\Lambda}^\star)
\end{align*}
when $\alpha \in (\frac{1}{2},1)$; non-asymptotically speaking, the variance of this term is $\bar{\bm{\Lambda}}_T/T$ by definition, and $\bar{\bm{\Lambda}}_T - \bm{\Lambda}^\star = \bm{O}(T^{\alpha-1})$;
\item The term $I_3$ is related to both the samples $\{(\bm{A}_t,\bm{b}_t)\}_{1 \leq t \leq T}$ and the intermediate iterations $\{\bm{\theta}_t\}_{0 \leq t \leq T}$. Hence, we refer to this term as the \emph{iterative error}. According to our analysis, this term converges by the rate of
\begin{align*}
\left\|\frac{1}{T}\sum_{i=1}^T \bm{Q}_i (\bm{A}_i-\bm{A})\bm{\Delta}_{i-1}\right\|_2 \lesssim O(T^{-\frac{\alpha}{2}}).
\end{align*}
\end{enumerate}

\begin{figure*}[h]
\begin{center}
\begin{tabular}{cc}
\includegraphics[width = 0.6\textwidth]{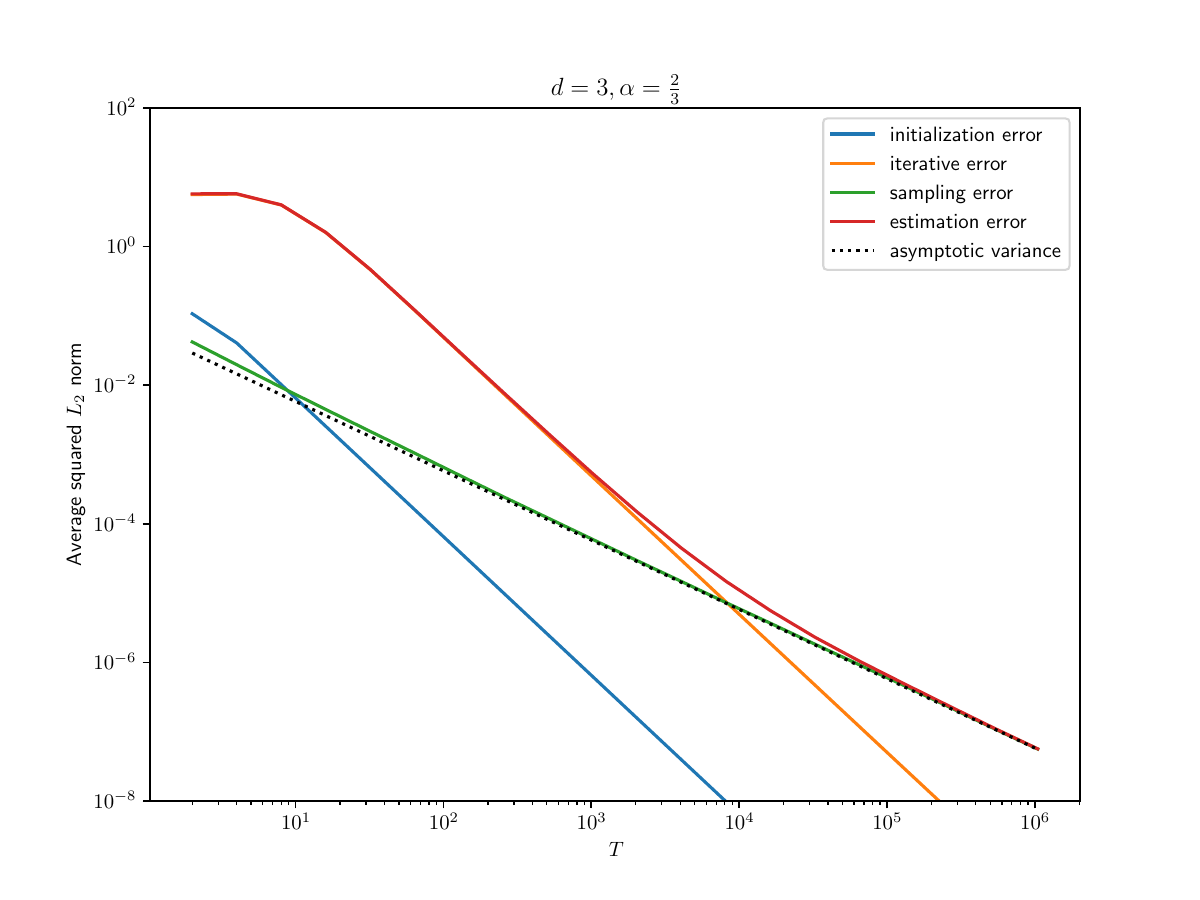} \\
\end{tabular}
\end{center}
\caption{Decomposition of the TD estimation error through $10,000$ independent trials, using MDP with $d=3$ and stepsizes $\eta_t = \eta_0 t^{-\frac{2}{3}}$}
\label{fig:decompose-errs}
\end{figure*}

\paragraph{Comparing the magnitudes of the components.} Figure \ref{fig:decompose-errs} illustrates the convergence of these terms in our $10,000$ independent trials when $d=3$ and $\alpha = \frac{2}{3}$. In the plot, the blue, orange and green lines represent the sample averages of $\|I_1\|_2^2, \|I_2\|_2^2$ and $\|I_3\|_2^2$ respectively; the red line represents the evolvement of $\|\bar{\bm{\Delta}}_T\|_2^2$, and the black dotted line represents the asymptotic variance $\mathsf{Tr}(\bm{\Lambda}^\star)/T$. As shown in the graph, the initialization error $I_1$ decreases the fastest; when $T < 10^3$, the iterative error $I_3$ is the dominant component in the estimation error, as the orange and red lines almost overlap. However, the slope of the orange line is steeper than that of the green line representing the sampling error $I_2$, an illustration that the iterative error converges faster as a function of $T$. In the mean time, the green line converges quickly to the black dotted line, verifying our claim that the sampling error corresponds to the asymptotic distribution. Finally, as $T$ becomes sufficiently large ($>10^5$ in this example), the green, red, and black dotted lines overlap with each other, an illustration that the sampling error dominates the estimation error, which converges to its asymptotic distribution.

\begin{figure*}[h]
\begin{center}
\begin{tabular}{cc}
\includegraphics[width = 0.6\textwidth]{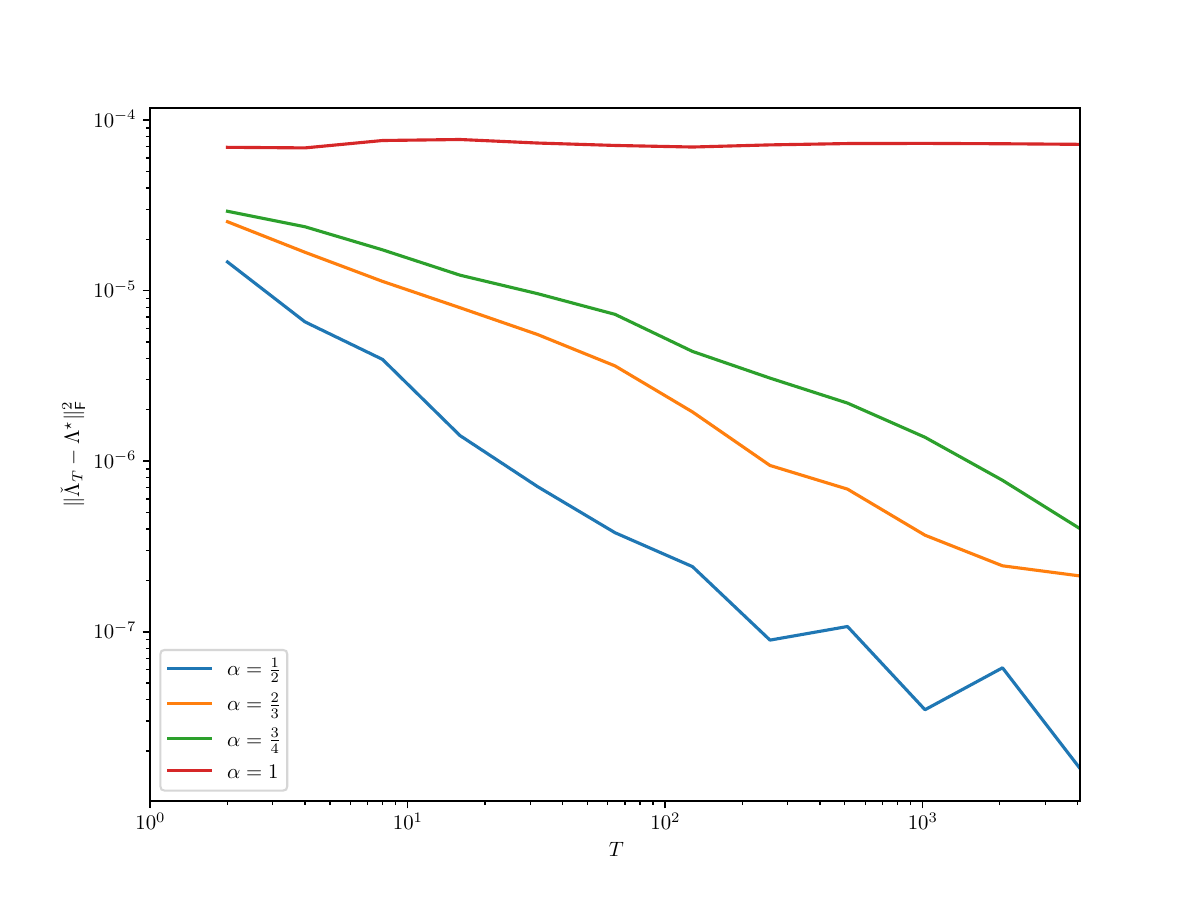} \\
\end{tabular}
\end{center}
\caption{The difference between $\check{\bm{\Lambda}}_T$ and $\bm{\Lambda}^\star$, for $d=3$ and different choices of $\alpha$}
\label{fig:compare_vars}
\end{figure*}

\paragraph{Difference between $\bar{\bm{\Lambda}}_T$ and $\bm{\Lambda}^\star$.} Theorem \ref{thm:Lambda} characterizes the difference between the "non-asymptotic variance" $\bar{\bm{\Lambda}}_T$ and the asymptotic variance $\bm{\Lambda}^\star$ as
\begin{align*}
\bar{\bm{\Lambda}}_T - \bm{\Lambda}^\star \asymp \bm{O}(T^{\alpha-1}),
\end{align*}
which results in our conclusion that the difference between the non-asymptotic and asymptotic distributions of $\bar{\bm{\Delta}}_T$ is lower bounded by $O(T^{\alpha-1})$. In order to illustrate this difference in variance through numerical simulations, we use $\check{\bm{\Lambda}}_T$ to denote the sample mean (averaged over our $10,000$ independent trials) of
\begin{align*}
\frac{1}{T}\left(\sum_{i=1}^T \bm{Q}_i(\bm{A}_i \bm{\theta}^\star-\bm{b}_i)\right)\left(\sum_{i=1}^T \bm{Q}_i(\bm{A}_i \bm{\theta}^\star-\bm{b}_i)\right)^\top.
\end{align*}
By definition, $\check{\bm{\Lambda}}_T$ is an unbiased estimator of $\bar{\bm{\Lambda}}_T$. Figure \ref{fig:compare_vars} demonstrates the difference between $\check{\bm{\Lambda}}_T$ and $\bm{\Lambda}^\star$, measured by the Frobenius norm of their difference, as $T$ increases. It is clear from the plot that the difference converges slower when $\alpha$ is larger, and especially that it does not converge to 0 when $\alpha = 1$. Notice that the plot is truncated before $T = 10^4$ as when $T$ becomes larger than the number of trials, the randomness within $\check{\bm{\Lambda}}_T$ dominates the difference, making the plot uninformative.

\section{Extension to general linear stochastic approximation}
\label{sec:sa.extension}
In this section, we explore on the conditions under which our theoretical analysis and statistical inference procedure can be extended to general Linear Stochastic Approximation (LSA) for the linear system
\begin{align*}
\bm{A\theta} = \bm{b},
\end{align*}
where a sample matrix $\bm{A}_t$ and a sample vector $\bm{b}_t$ are observed $i.i.d.$ at each time $t$, with expectations being $\bm{A}$ and $\bm{b}$ respectively. Similar to TD learning, we also run
\begin{align*}
\bm{\theta}_t = \bm{\theta}_{t-1} - \eta_t(\bm{A}_t \bm{\theta}_{t-1} - \bm{b}_t),
\end{align*}
with the same choice of stepsizes $\eta_t = \eta_0 t^{-\alpha}$, $\alpha \in (\frac{1}{2},1)$.

We first make the following assumptions on the matrix $\bm{A}$:
\begin{assumption}[Stability]\label{as:stability}
There exists a $d\times d$ symmetric matrix $\bm{U} \succeq \bm{0}$ and a constant $\beta_0 > 0$, such that $\bm{AU} + \bm{UA}^\top \succeq \beta_0 \bm{I}$;
\end{assumption}
Notice that in our setting of TD learning, this assumption is satisfied with $\bm{U} = \bm{I}$, and $\beta_0 = 2(1-\gamma)\lambda_0$; for general LSA, it is guaranteed as long as $-\bm{A}$ is \emph{Hurwitz}, i.e., all the eigenvalues of $-\bm{A}$ has strictly negative real-parts. Under that condition, $\beta_0$ can be set to $1$, and $\bm{U}$ is the unique solution to the Lyapunov equation $\bm{AU} + \bm{UA}^\top = \bm{I}$.
Meanwhile, we need the following constraints on the samples $\{(\bm{A}_t,\bm{b}_t)\}$:
\begin{assumption}[Sampling constraints]\label{as:sampling}
With $\bm{U}$ specified as in Assumption \ref{as:stability},
\begin{enumerate}
\item There exists a constant $c_U>0$, such that $\mathbb{E}[\bm{A}_t \bm{U}\bm{A}_t^\top] \preceq c_U (\bm{AU} + \bm{UA}^\top)$;
\item There exists constants $c_A>0,c_b>0$, such that $\|\bm{A}_t\|_{\bm{U}} \leq c_A$ and $\|\bm{b}_t\|_{\bm{U}}\leq c_b$ almost surely.
\end{enumerate}
\end{assumption}
Under our setting of TD learning, Assumption \ref{as:sampling} is satisfied with $c_U = 1, c_A = 2$ and $c_b = 1$ by Lemma \ref{lemma:A}. For general LSA with $-\bm{A}$ being Hurwitz, as long as (2) is satisfied, (1) can also be guaranteed by setting $c_U = c_A\|\bm{U}\|$.
As long as both Assumption \ref{as:stability} and Assumption \ref{as:sampling} are satisfied, it is immediate that $\|\bm{A}_t\bm{\theta}^\star-\bm{b}_t\|_{\bm{U}} \leq c_A \|\bm{\theta}^\star\|_{\bm{U}} + c_b$ almost surely, and that for every $\bm{x} \in \mathbb{R}^d$,
\begin{align*}
\|(\bm{I}-\eta \bm{A})\bm{x}\|_{\bm{U}}^2 &\leq \mathbb{E}\|(\bm{I}-\eta \bm{A}_t)\bm{x}\|_{\bm{U}}^2 \\ 
&= \|\bm{x}\|_{\bm{U}}^2 - \eta \bm{x}^\top (\bm{AU} + \bm{UA}^\top) \bm{x} + \eta^2 \bm{x}^\top \mathbb{E}[\bm{A}_t \bm{U}\bm{A}_t^\top]\bm{x} \\ 
&\leq \|\bm{x}\|_{\bm{U}}^2 - (\eta - c_U \eta^2)\bm{x}^\top (\bm{AU} + \bm{UA}^\top) \bm{x};
\end{align*}
as long as 
\begin{align*}
\eta \leq \frac{1}{2c_U}=:\eta_{\max},
\end{align*}
this bound can be further elaborated as
\begin{align*}
\|(\bm{I}-\eta \bm{A})\bm{x}\|_{\bm{U}}^2 &\leq \mathbb{E}\|(\bm{I}-\eta \bm{A}_t)\bm{x}\|_{\bm{U}}^2 \\ 
&\leq \|\bm{x}\|_{\bm{U}}^2-\frac{\eta}{2} \bm{x}^\top (\bm{AU} + \bm{UA}^\top) \bm{x} \\ 
&\leq \|\bm{x}\|_{\bm{U}}^2-\frac{\beta_0 \eta}{2}\|\bm{x}\|_2^2 \leq \left(1-\frac{\beta_0 \eta}{2\|\bm{U}\|}\right)\|\bm{x}\|_{\bm{U}}^2; 
\end{align*}
Consequently, as long as $\eta \leq \eta_{\max}$,
\begin{align*}
\|\bm{I}-\eta \bm{A}\|_{\bm{U}} \leq \sqrt{1-\frac{\beta_0 \eta}{2\|\bm{U}\|}} \leq 1-\frac{\beta_0 \eta}{4\|\bm{U}\|}.
\end{align*}
This is a direct generalization of Lemma \ref{lemma:A}. 

While the detailed derivations are well beyond the scope of this paper, our theoretical analysis in Theorems \ref{thm:bar-delta-t}, \ref{thm:Berry-Esseen-combined}, \ref{thm:plug-in-iid} can all be extended to general LSA with this generalized version of Lemma \ref{lemma:A}. Specifically:
\begin{enumerate}
\item For every $\delta \in (0,1)$, there exists a proper choice of $\eta_0$, such that with probability at least $1-\delta$,
\begin{align*}
\|\bar{\bm{\Delta}}_T\|_{\bm{U}} \lesssim \sqrt{\frac{\mathsf{Tr}(\bm{U\Lambda}^\star)}{T} \log \frac{1}{\delta}};
\end{align*}
\item As long as $\eta_0 \leq \eta_{\max}$, it can be guaranteed that
\begin{align*}
d_{\mathsf{C}}(\sqrt{T}\bar{\bm{\Delta}}_T,\mathcal{N}(\bm{0},\bm{\Lambda}^\star)) \lesssim C_1T^{-\frac{\alpha}{2}} + C_2T^{\alpha-1},
\end{align*}
with coefficients $C_1$ and $C_2$ related to $\bm{U},\beta_0, c_U, c_A, c_b$. Therefore, choosing $\alpha = \frac{2}{3}$ would still yield an $O(T^{-1/3})$ convergence rate;
\item With variance estimator $\hat{\bm{\Lambda}}_T$ defined in the same way as \eqref{eq:defn-hat-Lambda},
\begin{align*}
d_{\mathsf{TV}}(\mathcal{N}(\bm{0},\hat{\bm{\Lambda}}_T),\mathcal{N}(\bm{0},\bm{\Lambda}^\star)) \lesssim O(T^{-\frac{1}{3}})
\end{align*}
with probability at least $1-O(T^{-\frac{1}{3}})$;
\item Under the same statistical inference procedure as described in Section \ref{sec:inference}, the confidence region $\mathcal{C}_{\delta}$ still satisfies
\begin{align*}
\mathbb{P}(\bm{\theta}^\star \in \mathcal{C}_{\delta}) &\geq 1-\delta - O(T^{-\frac{1}{3}}).
\end{align*}
\end{enumerate}

\end{document}